\documentclass[11pt]{article}
\usepackage[T1]{fontenc}
\usepackage{mathptmx}
\usepackage[letterpaper,margin=1in]{geometry}
\usepackage{amsmath,amssymb,booktabs,graphicx,array,float,placeins,enumitem,adjustbox}
\usepackage[dvipsnames,table]{xcolor}
\usepackage[format=plain,labelformat=simple,labelsep=period,font=small]{caption}
\usepackage[numbers,sort&compress]{natbib}
\usepackage[hyphens]{url}
\usepackage{microtype}
\definecolor{resultblue}{HTML}{EEF3F8}
\definecolor{resultorange}{HTML}{FCF0E8}
\definecolor{linkblue}{rgb}{0.21,0.49,0.74}
\definecolor{impf}{HTML}{6C63D9}
\usepackage[breaklinks,colorlinks,allcolors=linkblue]{hyperref}
\hypersetup{pdftitle={Recoverable Symbols, Unstable Execution: Auditing Internal Pathways and Decision Identity in Visual State Tracking},pdfauthor={Ruiqi Liu, Jinming Li, Mo Wang, Yang Lu, Yanlan Kang}}
\newcommand{\fitsuppcolumn}[1]{#1}
\begin{document}
\begin{center}
{\LARGE\bfseries Recoverable Symbols, Unstable Execution:\\[2pt] Auditing Internal Pathways and Decision Identity\\[2pt] in Visual State Tracking\par}
\vspace{14pt}
{\large Ruiqi Liu$^{1,2}$\quad Jinming Li$^{2}$\quad Mo Wang$^{3,4,*}$\quad Yang Lu$^{2,*}$\quad Yanlan Kang$^{1,*}$\par}
\vspace{8pt}
{\normalsize $^1$\textbf{\color{impf}Institute of Medical Philosophy \& Future AI (IMPF-AI)}\par
$^2$Fudan University\qquad $^3$The University of Warwick\qquad $^4$Southern University of Science and Technology\par}
\vspace{3pt}
{\small $^*$Corresponding authors\par}
\vspace{10pt}
\includegraphics[height=1.05cm]{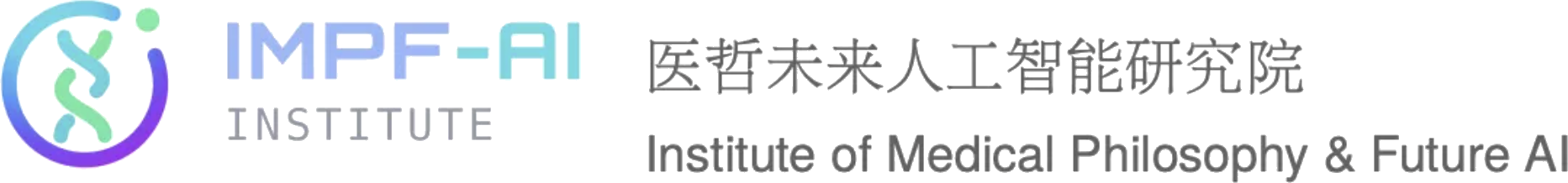}\hspace{1.2em}%
\includegraphics[height=0.95cm]{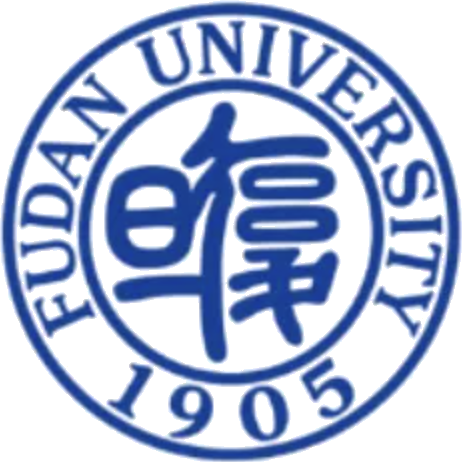}\hspace{1.0em}%
\includegraphics[height=0.95cm]{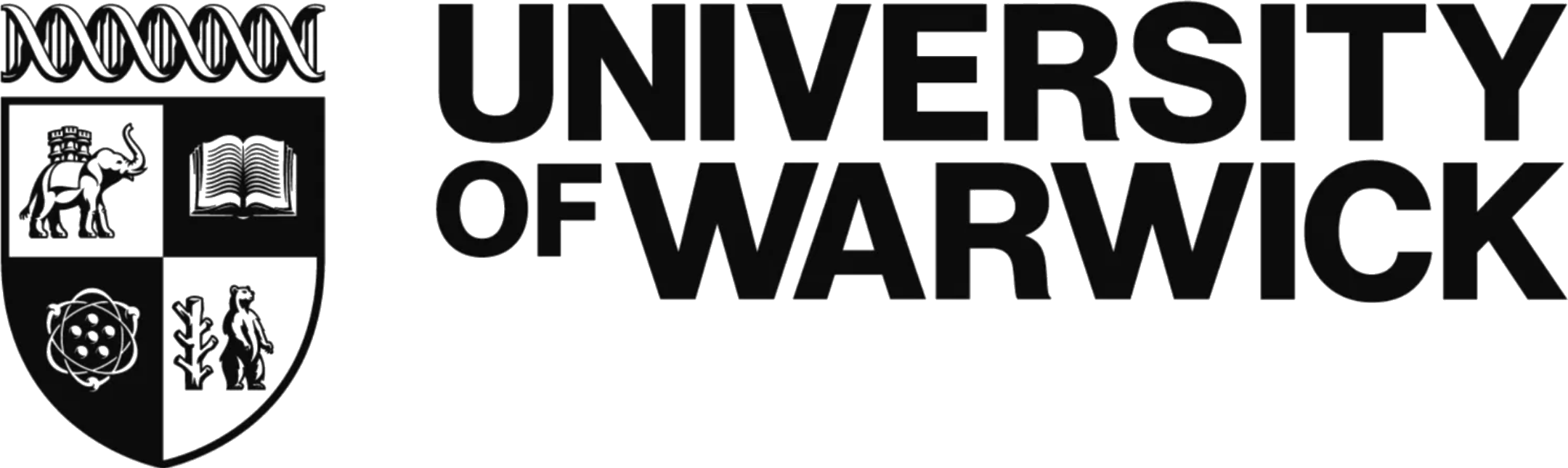}\hspace{1.0em}%
\includegraphics[height=0.95cm]{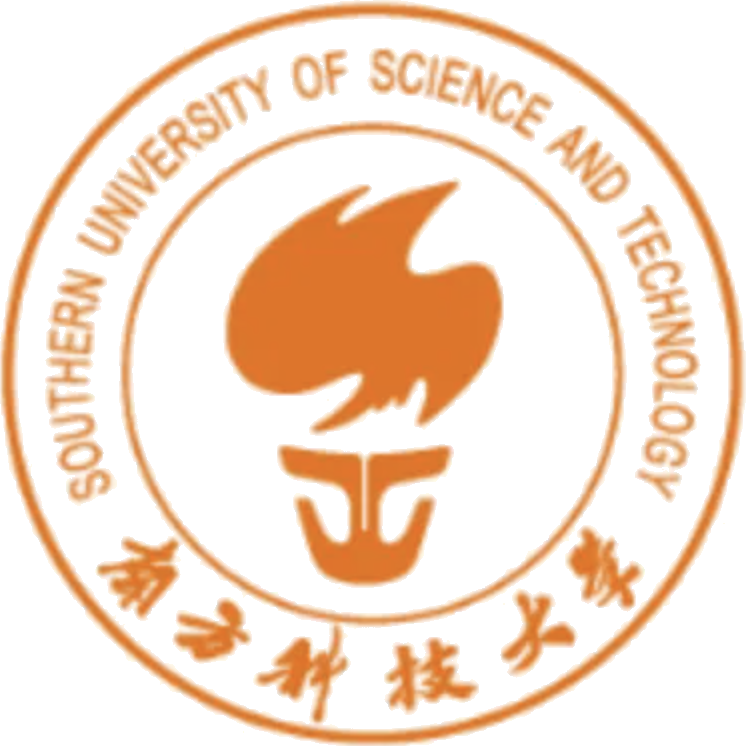}\par
\vspace{14pt}
\includegraphics[width=\linewidth]{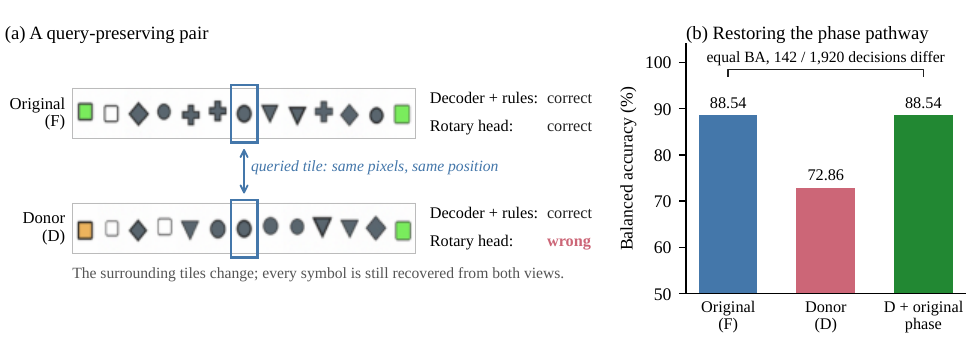}
\captionof{figure}{\textbf{Recoverable symbols, unstable decisions.} (a) For each query, the donor canvas (D) keeps the queried tile and its position from the original canvas (F) and changes the surrounding tiles; query 6 is shown. A symbol-supervised decoder reads every symbol correctly from both feature sequences, yet the Rotary head changes its answer. (b) On the 1,920-pair channel panel, putting the original rotation phase back into the donor computation restores balanced accuracy to 88.54\%, but 142 decisions still differ from the original ones.}
\label{fig:overview}
\end{center}
\begin{abstract}
\noindent Visual features can retain every symbol a task needs while a model built on them still changes its answer when irrelevant parts of the image change. We study this gap in visual state tracking, where a recurrent executor reads a sequence of symbol tiles through a frozen ViT encoder and decides whether a displayed candidate equals the final state. Our diagnostic pairs each input with a donor view that preserves every queried tile, checks with a symbol-supervised decoder that the full symbol chain is recoverable from both feature sequences, and swaps individual internal channels of the same fitted executor in both directions. On 3,526 pairs in which every symbol is recovered in both views, changing the surrounding context lowers the balanced accuracy of a rotary recurrent head by 17.13 points. The loss runs mainly through the input-dependent rotation phase: restoring the original phase repairs 351 of 391 new errors and brings accuracy back to 88.54\%, yet 142 individual predictions still differ from the original ones. A locked test on 192 unseen program families reproduces the pathway (28 of 29 new errors repaired; 382 of 384 predictions unchanged), and the effect partly carries over to single-image canvases. Consistency objectives derived from this analysis reduce the internal discrepancy they target but do not outperform ordinary data augmentation.
\end{abstract}

\section{Introduction}
Visual state tracking requires a model to read a sequence of symbols, apply their operations and compare the resulting state with a candidate. A wrong answer can arise at any of these stages. Recognition and execution are therefore different objects of evaluation, even when they share the same visual representation. A representation may retain every required symbol while the computation built on it remains sensitive to surrounding content. Fig.~\ref{fig:overview} shows an example.

Aggregate scores can conceal which decisions changed. An internal intervention may restore accuracy by correcting some examples while making others wrong. Paired inputs provide the reference needed to distinguish restored correctness from recovered original predictions.

Consider a program whose queried symbols and their positions stay fixed while other visible tiles change. A global encoder can change its features at the preserved queries because those features integrate the surrounding image. The question is what the learned executor does with this change. If every symbol can still be recovered from both feature sequences, a missing symbol cannot explain a changed answer, and the cause has to be sought in the executor.

Existing work already separates recognition from tracking, compares local and contextual inputs, and studies recurrent error dynamics~\cite{madan2026locality,cohen2026met,chung2026rethinking}. These studies measure recognition and execution on different inputs or interfaces. We measure both on the same paired instances and inside the same fitted executor.

We introduce a paired diagnostic for this purpose. It combines query-preserving interventions, complete-chain recovery and bidirectional substitutions of the executor's channels. A symbol-supervised decoder supplies an information witness, while the task head continues to operate on its original feature inputs. Injection asks whether a donor channel reproduces the execution loss. Restoration asks whether replacing that channel with its original value reverses the loss. Testing both directions separates a channel the computation depends on from one that merely changes along with it.

We report two main findings. First, complete symbol recovery coexists with a large execution loss, and a phase pathway accounts for most new errors in capable G5c Rotary heads. Second, restoring average balanced accuracy leaves individual predictions changed, so we report the errors an intervention repairs, the errors it introduces and its agreement with the original decisions. A locked test on new program families confirms the phase pathway.

The contributions are:
\begin{itemize}
\item A query-preserving diagnostic that verifies complete symbol recovery and tests the same fitted executor on paired features.
\item Bidirectional channel interventions, confirmed on new program families, that locate a phase-sensitive pathway in capable Rotary heads.
\item Decision-level recovery accounting and shared-canvas and training comparisons that test the pathway's explanatory and practical scope.
\end{itemize}

\section{Related Work}
\paragraph{Local perception and sequential tracking.}
Madan et al.~\cite{madan2026locality} study locality, shortcuts and length generalization with heads receiving common local streams. MET-Bench~\cite{cohen2026met} separates action recognition from entity tracking through a text interface. These works motivate testing the information presented to a sequential computation. We add a complete recovery witness on each paired instance and intervene within the same fitted executor. Background composites isolate recognition-context effects~\cite{xiao2021backgrounds}; the surrounding program tiles here retain semantics and are called \emph{non-query content}. Globally contextual ViT features~\cite{darcet2024registers} make preserved-pixel interventions informative about downstream execution.

\paragraph{Recoverability and internal interventions.}
Probing distinguishes information accessible to a decoder from information used by a decision rule~\cite{belinkov2022probing,elazar2021amnesic}. Visual counterfactuals examine binding and recognized geometry~\cite{cui2026binding,deng2026visualprobing}. Activation substitutions and interchange training provide tools for testing computational pathways~\cite{geiger2021causal,zhang2024patching,geiger2022inducing}. We combine these approaches with paired complete-program recovery and explicit error transitions. CVRR makes recurrent state a required image-conditioned route to the answer~\cite{park2026reason}; our analysis follows the transition, write and readout routes already present in a fitted head.

\paragraph{Recurrent dynamics and consistency learning.}
DeltaProduct motivates the one- and two-factor controls~\cite{grazzi2025negative,siems2025deltaproduct}. Rotary uses a block-rotation parameterization related to Mamba-3 within a simpler architecture~\cite{lahoti2026mamba}. Chung et al.~\cite{chung2026rethinking} analyze state-separating error, readability and finite failure horizons. We trace the additional changes induced by visual context through a fitted computation. The affine expansion organizes measurements of injection and propagation; the readout connects them to decisions. Augmentation and counterfactual invariance~\cite{hendrycks2020augmix,veitch2021counterfactual} provide training comparisons for constraints derived from this diagnosis.

\section{Method: Paired Diagnostics for Visual State Tracking}
\label{sec:design}
\paragraph{Programs and task semantics.}
A program specifies a start state $s_0$, operations $a_{1:T}$ and candidate $e$. For grammar transition $\pi_{g,a}$,
\begin{equation}
s_t=\pi_{g,a_t}(s_{t-1}),\qquad y=\mathbf1\{s_T=e\}.
\label{eq:task}
\end{equation}
G5c applies increments $\{1,-1,2,-2,0\}$ in $\mathbb Z_5$. For example, start 1 followed by $+2,-1,0$ ends at 2, making candidate 2 legal and candidate 3 illegal. Identity operations are rendered and processed like other inputs. G5r uses overlapping cycles $(j\ j+1\ j+2)$ for $j=0,1,2$, making operation order consequential. The grammars also differ in vocabulary and transition frequency.

The rendered $448\times112$ canvas contains 32-pixel tiles. A frozen DINOv2-S/14~\cite{oquab2024dinov2} or DINOv1-S/16~\cite{caron2021emerging} encoder supplies 384-dimensional features at known query centers. The supplied traversal orders the start, operations and candidate. Training contains 3--7 nonidentity operations; primary longer tests contain 8--10. Abstract-program partitioning removes identities and includes negative examples' source programs. Each full-budget block uses 16,000 training and 6,000 validation images, 5,040 updates, legality supervision and auxiliary state supervision. Blocks jointly vary data and initialization.

The operation sequence determines the final state, and the displayed candidate determines which comparison is asked, so the same program appears with both labels. Known query centers and traversal remove visual search, leaving representation and sequential updates. The candidate is still a visual input: its contextual feature affects the memory update, the query used to read memory and the classifier residual, so we also intervene on it separately.

\paragraph{The fitted executor.}
The matrix heads first standardize features using training-only statistics. A learned projection to width 128, a width-one depthwise convolution and SiLU produce $u_t$. The main Rotary head has four $22\times22$ memories. Each head projects a unit query $q_t$, unit key $k_t$, value $v_t$, decay $\rho_t=\sigma(W_\rho u_t+b_\rho)$ and eleven angles $\theta_t=\pi\tanh(W_\theta u_t+b_\theta)$. Its update is
\begin{equation}
S_t=S_{t-1}A_t+B_t,\quad
A_t=\rho_t R(\theta_t),\quad B_t=v_tk_t^\top,
\label{eq:architecture}
\end{equation}
with zero initial memory and block-diagonal planar rotations. Queries $q_t$ are continuous readout vectors; $s_t$ in Eq.~\ref{eq:task} denotes the discrete task state.

Concatenating $S_tq_t$ across heads gives $r_t$. The model forms $o_t=W_o\operatorname{LN}(r_t)+b_o+u_t$ and classifies $[o_n;u_n]$ with a two-layer GELU MLP at the candidate node $n$. A linear auxiliary readout predicts states from $o_t$. Thus the candidate updates memory before classification, and the residual enters both $o_n$ and the classifier input. The 2F control uses four $16\times16$ memories and two successive delta updates $S\leftarrow S-\beta(Sk-v)k^\top$, with unit keys and $\beta\in(0,2)$. GRU supplies a nonlinear recurrent control. Appendix~\ref{sec:complete-executors} gives complete equations, dimensions and selection protocols.

For a valid sequence, the optimized loss combines binary legality cross-entropy with auxiliary state cross-entropy of weight $1/2$. Auxiliary supervision covers the start and operation nodes, with identities inheriting the preceding state; the displayed candidate and padding are excluded. The primary recurrence processes every valid node in order, including identities. At evaluation, both views use the same fitted parameters and training-derived feature scaling. The interventions therefore change only the inputs or chosen internal channels of a fixed computation.

\paragraph{Query-preserving construction.}
For each node $t$, we copy its tile and absolute position into a matched donor canvas, encode that canvas and sample only its query feature $z_t^D$ (Fig.~\ref{fig:overview}). The assembled donor sequence $Z^D=(z_t^D)_t$ therefore combines separately encoded canvases. Its label remains that of the preserved source-symbol chain. Donors match split, node count, carrier and vocabulary. We compare the same fitted head on the original F and donor D sequences, fixing normalization, readout and decision threshold. 

Initial sampling support is audited separately. DINOv2 bilinear sampling extends 2--4 pixels beyond the tile at eight of thirteen positions, including two that reach blank space. DINOv1 sampling lies inside the preserved tile, so its initial sampled embeddings are identical across views. This second encoder tests whether sensitivity persists with unchanged initial support.

Later attention layers can still propagate distant differences. The shared-canvas protocol in Sec.~\ref{sec:canvas} instead obtains every target query from one image and one encoder pass.

\paragraph{Complete recovery witness.}
Role-specific ridge decoders learn start, operation and candidate identities from full-context training tokens; original validation tokens select regularization. An exact grammar executor then evaluates their decoded chain. A pair is \emph{jointly exact} when every symbol is correct in both views. This demands complete recovery, including intermediate operations and the candidate, so compensating token errors cannot explain the selected pair's exactness.

The witness and task head are parallel readouts of the same features. The witness receives symbol supervision and known transition rules. The learned executor receives legality and auxiliary states and must learn its own use of the representation. Complete recovery thus shows that the information needed for exact execution is present in both feature sequences; the executor's decisions on these pairs show whether it uses that information consistently.

\paragraph{Interventions and estimands.}
We exchange a named channel while keeping all other channels at their recipient values. Donor injection uses F as recipient; original restoration uses D. Both operation and candidate nodes enter full phase exchange. We report balanced accuracy (BA), the mean of classwise recalls, along with corrected errors, introduced errors and disagreement with F. Eight-block estimates weight blocks equally; paired intervals resample whole blocks unless specified otherwise. Jointly exact subsets condition on the decoding outcomes.

Corrected and introduced errors compare the intervention with its recipient using the true label. Disagreement with F tracks prediction identity independently of correctness. Repairs of newly induced donor errors test reversal of the original failure. On class-balanced panels, accuracy equals within-block BA; sensitivity exclusions require classwise recomputation.

\section{Experiments}
\label{sec:experiments}
\FloatBarrier
\subsection{Confirmation on new program families}
\label{sec:confirmation}
\begin{figure}[H]
\centering\includegraphics[width=0.62\linewidth,trim=0 0 0 13pt,clip]{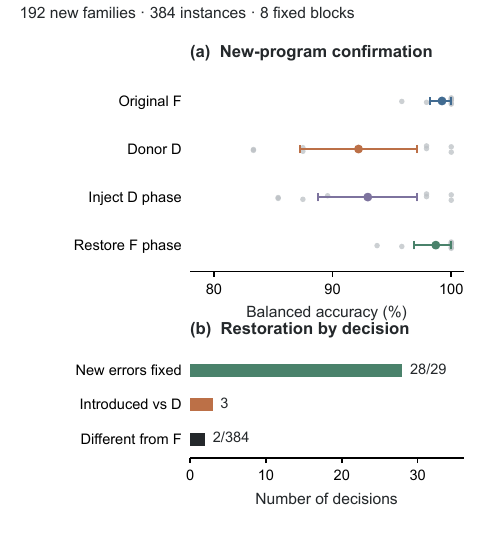}
\caption{\textbf{Phase confirmation with frozen models on new programs.} (a) BA on 384 instances from 192 new families in eight fixed blocks. Dots show every block; marks and bars give equal-block means and paired 95\% bootstrap intervals. All chains recover exactly. (b) Restoration repairs 28/29 newly induced donor errors, introduces three errors relative to D, and changes two predictions relative to F.}
\label{fig:confirmation}
\end{figure}

We start with the locked confirmation (Fig.~\ref{fig:confirmation}). Eight existing DINOv1 Rotary checkpoints and their decoders were frozen before encoding 192 new canonical families. Each family supplies legal and illegal candidates, giving 384 instances over eight blocks. All families are disjoint from retained training, validation and previous test families. Original BA ranges from 95.83\% to 100\%, and all 384 pairs have complete recovery in both views.

New-family generation is performed after the model and decoder files, exclusion rules and intervention protocol are fixed. Canonicalization removes identity operations, normalizes operation names by first occurrence, and identifies forward/reversed normalized sequences. The family exclusion union includes retained historical training, validation and test programs as well as source programs used to construct negative candidates. Uncertainty is evaluated across the eight fixed training/program blocks.

Original BA is 99.22\% and donor BA is 92.19\% ($-7.03$ pp, 95\% CI $[-11.46,-2.60]$). Injecting donor phase yields 92.97\%. Restoring original phase within D yields 98.70\%, repairs 28/29 new donor errors and introduces three errors relative to D. Two of 384 predictions still differ from F. The pathway thus carries over to unseen programs, and here too repair is not the same as exact recovery of predictions. Removing the thirteen within-family donor assignments leaves 371 pairs, a $-7.30$ pp donor effect and the same 28/29 repairs, three introduced errors and two disagreements (Appendix~\ref{sec:sibling-sensitivity}).

\FloatBarrier
\subsection{Complete recovery retains an execution gap}
\label{sec:recovery}
The larger recovery panel contains 3,840 paired inputs, 480 per block. Donor replacement changes G5c Rotary BA from 88.80\% to 71.80\% ($-17.01$ pp, $[-19.21,-13.73]$). Among 3,526 jointly exact pairs, original and donor BA are 89.18\% and 72.06\%: the effect remains $-17.13$ pp ($[-19.26,-13.60]$). The loss is unchanged when every symbol is recovered in both views. Intervals resample blocks and donor-linked components to preserve the paired dependence.

\begin{figure}[H]
\centering\includegraphics[width=0.62\linewidth]{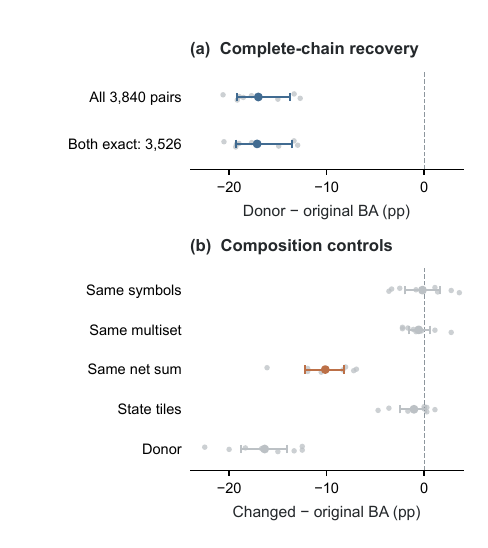}
\caption{\textbf{Execution loss survives complete recovery.} (a) All 3,840 recovery pairs and the 3,526 jointly exact pairs. (b) Controls on 960 overlapping source programs. Same net sum preserves the modular effect, identity mask and state tiles; same multiset preserves operation composition. Gray dots show eight block effects. Bars are paired 95\% bootstrap intervals, using donor-linked components in (a).}
\label{fig:recovery}
\end{figure}

A composition-control panel uses 960 source programs, 120 per block. Permuting the same operation multiset changes BA by only $-0.08$ pp on the composition-control panel. Changing surrounding operation values while preserving their modular sum, identity mask and state tiles changes BA by $-10.14$ pp ($[-12.26,-8.26]$), or $-9.64$ pp among jointly exact pairs. Preserving the sum holds the surrounding operations' net modular effect fixed, whereas preserving the multiset also holds their composition fixed. Their contrast identifies composition sensitivity beyond the net effect.

Same-symbol rerendering changes only appearance, and state-tile changes affect only the start and endpoint tiles; both have smaller effects (Fig.~\ref{fig:recovery}). The executor is therefore sensitive to the composition of the surrounding operations. Direction tests for more specific shortcuts are in Appendix~\ref{sec:supp-legacy}.

The 1,920-pair channel panel supplies the same paired inputs to the learned controls. Table~\ref{tab:paired-core} shows that 2F and GRU incur smaller accuracy losses and fewer prediction changes than Rotary. They still change 7.19\% and 3.85\% of decisions. Full capability, parameter and budget comparisons are in Appendix~\ref{sec:supp-training}. The composition and channel panels reuse sources from the recovery panel and overlap on 497 rows; they test distinct interventions on those sources.

\begin{table}[H]\centering\small
\caption{\textbf{Core paired comparison.} DINOv2 G5c channel panel: 1,920 pairs, eight blocks. F/D are original/donor. BA and flips are percentages; $\Delta$ is D minus F in pp. Every head is held fixed across views. The decoder uses symbol supervision and exact rules as a recovery witness.}
\label{tab:paired-core}
\begin{adjustbox}{max width=\linewidth}\begin{tabular}{lrrrr}\toprule
System & F BA & D BA & $\Delta$ & Flips \\
\midrule
\rowcolor{resultorange}Rotary & 88.54 & 72.86 & \textbf{$-15.68$} & \textbf{25.05} \\
2F & 96.30 & 94.53 & $-1.77$ & 7.19 \\
GRU & 97.66 & 97.24 & $-0.42$ & 3.85 \\
Decoder + rules & 96.77 & 97.50 & $+0.73$ & --- \\
\bottomrule\end{tabular}\end{adjustbox}
\end{table}

\FloatBarrier
\subsection{Bidirectional interventions identify a phase pathway}
\label{sec:mechanism}
An exact affine expansion guides the channel measurements. For original F and donor D, define
\begin{equation}
E_t=S_{t-1}^F(A_t^D-A_t^F)+(B_t^D-B_t^F).
\label{eq:innovation}
\end{equation}
The state difference satisfies
\begin{equation}
\delta_n=\sum_{t=1}^n E_t\prod_{j=t+1}^n A_j^D,
\quad \delta_t=S_t^D-S_t^F.
\label{eq:transport}
\end{equation}
It separates newly injected differences from their propagation. For two planar rotation matrices, $\|R(\theta_D)-R(\theta_F)\|_2=2|\sin((\theta_D-\theta_F)/2)|$. The effect on a decision additionally depends on the stored state, later transitions and readout.

\begin{figure}[H]
\centering\includegraphics[width=0.62\linewidth]{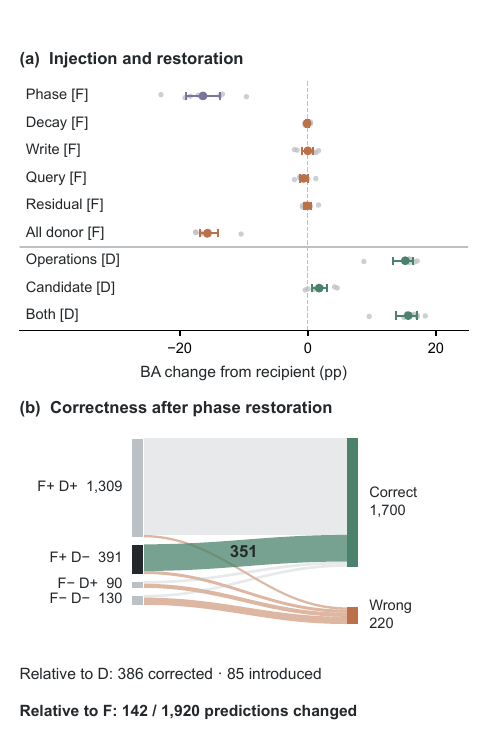}
\caption{\textbf{A pathway can restore BA while changing decisions.} DINOv2 channel panel, 1,920 pairs and eight blocks. (a) Injection rows use F as recipient; restoration rows use D. Gray dots are blocks; intervals are paired 95\% block-bootstrap intervals. (b) All original/donor/restored correctness transitions are retained; $+$/$-$ mean correct/incorrect and ribbon widths encode counts. Restoration repairs 351 new errors, corrects 386 donor errors overall, introduces 85, and leaves 142 predictions different from F.}
\label{fig:mechanism}
\end{figure}

On the DINOv2 channel panel, injecting donor phase lowers BA from 88.54\% to 72.19\%, close to the full donor value of 72.86\%. Individual decay, write, query and residual substitutions have smaller effects. In the reverse direction, restoring original phase raises donor BA to 88.54\% and fixes 351/391 newly wrong decisions (Fig.~\ref{fig:mechanism}). The four-cell comparison fixes non-phase channels to F or D while choosing phase from either view. It links both the induced loss and its reversal to the phase pathway under the respective recipient computations.

Restoration corrects 386 donor errors and introduces 85 errors. Relative to F, 71 originally correct answers become wrong and 71 originally wrong answers become correct. Equal BA therefore conceals 142/1,920 changed predictions (7.40\%). Recovering the score is therefore not the same as recovering the original outputs. Appendix~\ref{sec:supp-mechanism} gives the full four-cell results and restoration costs.

Role-specific swaps further locate the response. Operation-only restoration repairs 343/391 new errors, compared with 61/391 for the single candidate node. These compare node sets of different sizes. For one head, the candidate update before reading gives
\begin{equation}
r_{n,h}=S_{n-1,h}(A_{n,h}q_{n,h})+B_{n,h}q_{n,h}.
\label{eq:effective-query}
\end{equation}
Candidate phase thus changes the effective query applied to accumulated memory. The normalized readout, residual and classifier convert the resulting change into a final margin. DINOv1 tests on the same channel-panel RGB inputs reach 99.58\% original and 92.86\% donor BA with complete recovery; full phase restoration repairs all 134 new errors and reproduces the original predictions.

\FloatBarrier
\subsection{Shared canvases and partial phase transfer}
\label{sec:canvas}
The query-specific protocol assembles features from multiple canvases. We therefore train new heads in a fixed two-row layout with a target program and a distractor. Each view samples a distractor from the same balanced law, while keeping all target pixels and query positions fixed. All target features in a view come from one encoder pass. D0 and D1 are symmetric samples, making prediction disagreement the natural sensitivity measure. 

\begin{figure}[H]
\centering\includegraphics[width=0.62\linewidth]{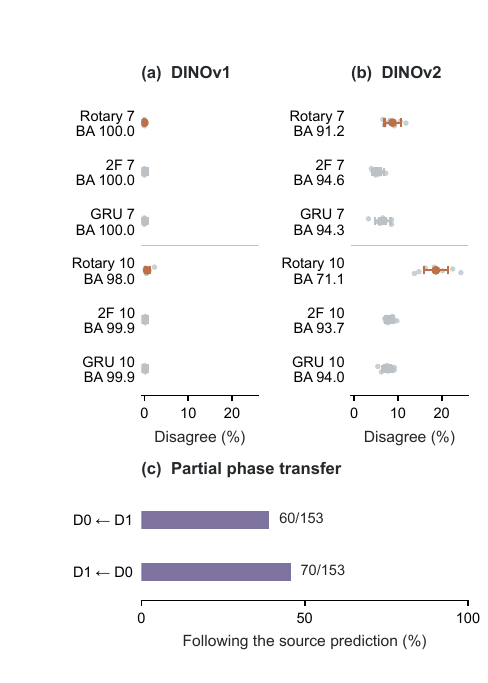}
\caption{\textbf{Coherent images retain sensitivity with partial phase transfer.} (a,b) All twelve conditions, separated by encoder; rows identify head and length, with mean-view BA. Dots show all eight blocks; bars are archived 95\% block/family-structured intervals for disagreement. L7/L10 contain 2,432/3,840 pairs. (c) Of 153 jointly exact DINOv2 L7 disagreements, phase exchange follows the source prediction on 60 and 70 pairs in the two directions.}
\label{fig:canvas}
\end{figure}

On 2,432 length-seven pairs, DINOv2 Rotary scores 91.04\% and 91.37\% BA and changes 214 decisions (8.80\%, $[6.95,10.69]$). Among 2,053 jointly exact pairs, 153 still disagree (7.45\%, $[5.65,9.38]$). DINOv1 answers every length-seven pair correctly. At length ten, DINOv1 reaches 98.05\% and 97.99\% BA with 24/3,840 disagreements and complete recovery. Figure~\ref{fig:canvas} shows all twelve encoder/head/length conditions.

The mechanism transfers partially. Among the 153 jointly exact DINOv2 length-seven disagreements, exchanging all phase makes only 60 predictions follow D1 when D0 is the recipient, and 70 follow D0 in the reverse direction. These counts measure how often the prediction follows the source, not corrected errors. In coherent images the gap between recoverability and stability remains, but phase explains only part of it.

\FloatBarrier
\subsection{From internal restoration to training objectives}
\label{sec:adaptation}
The original four adaptation arms share checkpoints, 2,000 training rows, 480 validation rows and 256 updates. Clean ERM, donor augmentation, output consistency and counterfactual transport consistency (CTC) receive matched task-loss evaluations. CTC penalizes normalized innovation in Eq.~\ref{eq:innovation} using a stop-gradient original trajectory. A later read/residual transport objective (RTC) covers the interface feeding the classifier. Table~\ref{tab:training-objectives} summarizes the original and later training signals; Appendix~\ref{sec:transport-followup} gives formulas, validation selection and costs.

\begin{table}[H]\centering\small
\caption{\textbf{Training signals compared.} Except for clean ERM, each arm uses the paired task loss. Additional penalties target different interfaces. Weight and checkpoint selection use validation data; Appendix~\ref{sec:transport-followup} gives exact formulas and budgets.}
\label{tab:training-objectives}
\begin{adjustbox}{max width=\linewidth}\begin{tabular}{lp{1.93in}}\toprule
Arm & Training signal \\
\midrule
Clean ERM & Original-view task loss \\
Augmentation & Original and donor task losses \\
Probability & Paired loss + output probability agreement \\
Margin & Paired loss + binary-margin agreement \\
CTC & Paired loss + normalized update innovation \\
RTC & Paired loss + normalized read/residual agreement \\
\bottomrule\end{tabular}\end{adjustbox}
\end{table}

Augmentation gains 5.16 pp donor BA and loses 3.44 pp original BA relative to clean ERM. CTC reduces held-out innovation discrepancy in every block but adds only $+0.16$ pp donor BA over augmentation ($[-0.73,1.09]$). In a full-budget rerun with validation-selected weights and 1,920 new rows from 960 families, augmentation yields 74.95\%/67.76\% original/donor BA, CTC 74.84\%/67.40\%, and RTC 74.17\%/66.61\%. Both objectives reduce the discrepancy they target but do not improve accuracy over augmentation.

A two-block DINOv1 validation pilot tests a more direct constraint: predicting operation phase from local pixels makes the angles exactly invariant across views. Replacing those angles by a pooled constant reduces original BA to 48.1\% and 52.1\%. Execution therefore depends on input-varying operation phase in these fitted heads. Nevertheless, restoring original writes raises donor BA from 65.21\% to 83.54\% in one block and from 98.13\% to 100\% in the other. Making phase invariant thus leaves a write pathway open. The pilot stopped at validation (Appendix~\ref{sec:supp-local}).

\FloatBarrier
\section{Discussion}
Complete recovery changes what a contextual failure means. If a decoder reads every symbol from both feature sequences, the encoder has not lost the information; the failure lies in how the learned executor uses it. Probing work separates decodability from use~\cite{belinkov2022probing,elazar2021amnesic}; the paired construction measures both on the same instances.

In capable G5c Rotary heads, most of the contextual loss passes through the rotation phase: injecting donor phase reproduces it and restoring the original phase removes it, on the channel panel and on new program families. Phase acts on stored memory together with decay, writes and readout, so its effect depends on the recipient computation. In shared canvases phase explains only part of the disagreements, and the other channels need the same analysis.

\paragraph{Implications for training.}
Fixing one channel does not make the executor robust. When operation phase is computed from local pixels, it agrees across views, yet the refitted head still depends on how phase varies with the input and becomes sensitive through its writes instead. Penalizing local state innovation reduces its average size but protects neither a small classification margin nor the query and residual paths. The readout bound in Appendix~\ref{sec:readout-transport-bound} makes these terms explicit. Margin-aware weighting and joint constraints on update, state and readout are natural next objectives, with ordinary augmentation as the baseline.

\paragraph{Reporting decisions, not only scores.}
A restored score can hide repaired and newly introduced errors that cancel. 2F and GRU lose less accuracy but still change 7.19\% and 3.85\% of decisions. We therefore report error transitions and paired agreement next to BA. Keeping every original answer is not always desirable, since some original answers were wrong; the counts make this trade-off visible.

\paragraph{Limitations.}
All experiments use generated symbols, a supplied traversal and frozen encoders. The query-specific and shared-canvas protocols differ in both training and context distribution, and the two encoders differ in pretraining and patch geometry. The local-phase study covers two validation blocks. Natural scenes, unknown transition rules and capable VLMs are left for future work.

\section{Conclusion}
A visual executor can fail on inputs from which every needed symbol is recoverable. In capable rotary heads this failure runs mainly through the input-dependent phase, as shown by injection and restoration on both analyzed and unseen programs. Restoring accuracy does not restore the original predictions, so decision-level counts should accompany aggregate scores in this kind of analysis.

\section*{LLM Usage Statement}
GPT through Codex assisted with language polishing and routine code completion. The authors verified the manuscript and code against the recorded methods and results and take responsibility for all content.
\FloatBarrier

\clearpage
\appendix
\setcounter{figure}{0}\setcounter{table}{0}\setcounter{equation}{0}
\renewcommand{\thefigure}{A\arabic{figure}}\renewcommand{\thetable}{A\arabic{table}}\renewcommand{\theequation}{A\arabic{equation}}
\section*{Appendix Overview}
\addcontentsline{toc}{section}{Appendix Overview}
This appendix gives the protocols, full numerical results and additional controls behind the main text. All trained blocks and comparison arms are reported, including low-capability conditions. Unless stated otherwise, the eight paired training/program blocks are the independent units, intervals are paired 95\% whole-block bootstrap intervals, accuracy is balanced accuracy (BA) in percent, and differences are in percentage points (pp). Table~\ref{tab:supp-guide} maps each result in the main text to its supporting material.

\paragraph{Panels and notation.}
We use the panel names of the main text. The \emph{recovery panel} contains 3,840 archived G5c pairs (480 per block). The \emph{composition-control panel} contains 960 of its source programs (120 per block). The \emph{channel panel} contains 240 pairs per block and grammar (1,920 per grammar); it supports the channel substitutions and the complete context matrix, and its G5c rows share 497 source programs with the composition-control panel (Fig.~\ref{fig:panel-lineage}). The \emph{new-family confirmation} contains 384 pairs from 192 unseen canonical families (Sec.~\ref{sec:review-independent}). F denotes the original view and D the query-preserving donor view; V, S, P and P$^+$ are the further context interventions of Sec.~\ref{sec:supp-matrix}. G5c and G5r are the two five-state grammars, and Rotary, 2F, 1F and GRU are the learned heads (Sec.~\ref{sec:complete-executors}). \emph{Decoder + rules} denotes the symbol-supervised role decoder followed by exact transitions. It serves as the recovery witness and is labeled ``fixed rules'', ``probe + rules'' or ``decoder--executor'' in some displays. Flips are the percentage of paired decisions that differ between two views. CTC and RTC are the counterfactual transport-consistency and readout transport-consistency objectives (Secs.~\ref{sec:supp-adaptation} and~\ref{sec:transport-followup}).

\begin{table}[!htbp]\centering\small
\caption{\textbf{Guide to the appendix.}}
\label{tab:supp-guide}
\begin{adjustbox}{max width=\linewidth}\begin{tabular}{@{}p{.40\linewidth}p{.56\linewidth}@{}}
\toprule
Result in the main text & Supporting material \\
\midrule
Task, heads and training (Sec.~\ref{sec:design}) & Secs.~\ref{sec:supp-data}--\ref{sec:supp-training} and~\ref{sec:supp-identities}; Tabs.~\ref{tab:original} and~\ref{tab:full-capability-matrix} \\
New program families (Sec.~\ref{sec:confirmation}, Fig.~\ref{fig:confirmation}) & Sec.~\ref{sec:review-independent}; Tabs.~\ref{tab:review-independent}, \ref{tab:final-recovery-cost} and~\ref{tab:final-sibling} \\
Complete recovery and composition (Sec.~\ref{sec:recovery}, Fig.~\ref{fig:recovery}) & Secs.~\ref{sec:supp-matrix}--\ref{sec:supp-legacy}; Tab.~\ref{tab:newmatrix}; Fig.~\ref{fig:panel-lineage} \\
Phase pathway and decision accounting (Sec.~\ref{sec:mechanism}, Tab.~\ref{tab:paired-core}, Fig.~\ref{fig:mechanism}) & Secs.~\ref{sec:supp-mechanism} and~\ref{sec:supp-role-phase}; Tabs.~\ref{tab:final-four-cell} and~\ref{tab:role-phase-supp}; Fig.~\ref{fig:historical-channels} \\
Shared canvases (Sec.~\ref{sec:canvas}, Fig.~\ref{fig:canvas}) & Secs.~\ref{supp:patch} and~\ref{sec:supp-canvas}; Tabs.~\ref{tab:coherent-canvas-supp} and~\ref{tab:coherent-phase-supp} \\
Training objectives (Sec.~\ref{sec:adaptation}, Tab.~\ref{tab:training-objectives}) & Secs.~\ref{sec:supp-adaptation}, \ref{sec:transport-followup} and~\ref{sec:full-budget-followup}; Tabs.~\ref{tab:adaptation} and~\ref{tab:full-transport} \\
Second encoder (DINOv1) & Sec.~\ref{sec:supp-second-encoder}; Tab.~\ref{tab:second-encoder}; Fig.~\ref{fig:second-encoder} \\
Capability gates and scope & Secs.~\ref{sec:supp-vocab}, \ref{sec:supp-independent} and~\ref{sec:supp-local}; Tab.~\ref{tab:review-scope} \\
Reproducibility & Sec.~\ref{sec:supp-repro} \\
\bottomrule
\end{tabular}\end{adjustbox}
\end{table}

\section{Experimental Matrix and Reproducibility}
\subsection{Data, rendering, and partitions}
\label{sec:supp-data}
The original canvas is $448\times112$ pixels. Tile $t$ is centered at $(32+32t,56)$ and has width 32. Each tile is independently supersampled, jittered, and perturbed with pixel noise. Randomness depends on the rendering seed and node index. A queried tile is copied exactly when its surrounding context changes. The original and isolated renderings use the same operation vocabulary and state carrier. This prevents future semantic states from controlling local rendering noise.

There are 16,000 training, 6,000 validation, 6,000 training-range test, and 6,000 long-chain test images per grammar and block. Training and validation contain 3--7 nonidentity operations; the long-chain split contains 8--10. Abstract operation strings with identities removed are partitioned, including displayed strings and the sources used to generate negative examples. The mean number of distinct displayed training programs is approximately 2,146 for G5r and 6,476 for G5c; long-chain counts are approximately 5,597 and 5,957. Image counts therefore include repeated abstract programs. Blocks are the independent unit in all interval estimates.

DINOv2-S/14 supplies a 384-dimensional feature at each known center. Raw features and earlier standardized caches use half precision. The recurrence uses single precision. Training-node means and standard deviations exclude padding and remain fixed for all validation and intervention inputs. Isolated encoding required approximately 8.8--8.9 times as many canvases and 3.1--3.3 times the encoder time. Cost therefore differs even when the downstream sequence is identical.

\subsection{Complete executor definitions}
\label{sec:complete-executors}
The primary inputs are ordered 384-dimensional frozen-encoder features. Training-only feature means and standard deviations are fixed at evaluation. Matrix heads share a width-128 affine projection, a depthwise convolution with kernel one, and SiLU. Thus $u_t=\operatorname{SiLU}(C(W_z\widetilde z_t+b_z))$, with a pointwise learned $C$. The primary projection has no additional neighboring sequence inputs. Global visual encoding can already transmit contextual information to $\widetilde z_t$.

For each matrix head $h$, queries and keys are linear projections of $u_t$ normalized to unit Euclidean length; values are unconstrained linear projections. Rotary uses four heads with key and value dimension 22, giving four $22\times22$ zero-initialized memories and an 88-dimensional concatenated read. Each head has eleven input-dependent planar angles and a scalar decay:
\begin{align}
 \theta_{t,h}&=\pi\tanh(W_{\theta,h}u_t+b_{\theta,h}),\\
 \rho_{t,h}&=\sigma(W_{\rho,h}u_t+b_{\rho,h}),\\
 R_h(\theta)&=\operatorname{blockdiag}_{j=1}^{11}
 \begin{pmatrix}\cos\theta_j&\sin\theta_j\\-\sin\theta_j&\cos\theta_j\end{pmatrix},\\
 S_{t,h}&=\rho_{t,h}S_{t-1,h}R_h(\theta_{t,h})+v_{t,h}k_{t,h}^{\top}.
\end{align}
The displayed rotation follows the implementation's row-vector right-multiplication convention. The spectral norm of the difference of two corresponding planar matrices is $2|\sin((\theta_D-\theta_F)/2)|$. Angles and matrix differences are distinct quantities.

The 2F head uses four $16\times16$ memories and two factors at each input. With independently projected $k_{t,h,j},v_{t,h,j}$ and $\beta_{t,h,j}=2\sigma(W_{\beta,h,j}u_t+b_{\beta,h,j})$, define $A_j=I-\beta_jk_jk_j^\top$ and $B_j=\beta_jv_jk_j^\top$. The sequential update is $S_t=(S_{t-1}A_1+B_1)A_2+B_2$. Equivalently, its effective affine transition and write are $A_1A_2$ and $B_1A_2+B_2$. The one-factor negative-eigenvalue control has four heads of dimension 24. All valid nodes, including identities, update the memory; padding leaves it unchanged.

The shared matrix-head readout is
\begin{align}
 r_t&=\operatorname{concat}_h(S_{t,h}q_{t,h}),\\
 o_t&=W_o\operatorname{LN}(r_t)+b_o+u_t,\\
 \ell&=W_2\operatorname{GELU}(W_1[o_n;u_n]+b_1)+b_2.
\end{align}
Here $W_o$ maps the concatenated read to 128 dimensions, $W_1$ maps 256 to 64 dimensions, and $W_2$ maps 64 to two legality logits. A linear 40-way auxiliary head reads $o_t$ using the dataset's shared state-label encoding; the G5 tasks occupy five task states. Start and operation positions contribute auxiliary supervision; the displayed candidate and padding are excluded. The candidate is processed before legality classification. The same $u_n$ contributes through the residual and the separate classifier input.

The primary GRU projects 384 inputs to width 128, applies one unidirectional GRU layer of width 128, and uses linear legality and auxiliary readouts of the last valid and per-node hidden vectors. The parameter-matched GRU uses width 112 for these visual inputs. The full-budget parameter-matched Transformer uses a 384-to-64 input projection, learned positions, two causal pre-normalized layers, four attention heads, width-256 feedforward sublayers, dropout 0.1, and linear task/state readouts. Padding is masked and legality is read at the candidate position. The complete Mamba-1 head and its numerical checks are specified in the training matrix section below. Its reference scan and the simpler rotary head are separate architectures.

\subsection{Original and new training matrices}
\label{sec:supp-training}
Original blocks pair data seeds 101--108 with initialization seeds 0--7. The full budget is 80 epochs, 5,040 updates, batch 256, AdamW learning rate $3\times10^{-3}$, weight decay $10^{-2}$ on eligible matrices, gradient clipping at one, and cosine decay following 100 warm-up updates. Legality cross-entropy is augmented by state cross-entropy of weight $1/2$. Padding and the displayed candidate endpoint are excluded from state supervision. Identity positions inherit the preceding state.

Original model selection uses the latest epoch within $\max(0.005,\sqrt{p(1-p)/n_{\rm val}})$ of the best validation BA. The original matrix comprises two grammars, five heads, two contexts, and eight blocks: 160 fits. Its saved predictions were independently recomputed. Table~\ref{tab:original} gives the complete matrix, retaining the widened one-factor control.

\begin{table}[!htbp]\centering\small
\caption{\textbf{Original full-budget long-chain matrix.} Mean and sample SD over eight blocks; all cells use 6,000 long-chain examples per block. Both contexts are separately trained and evaluated.}
\label{tab:original}\fitsuppcolumn{\begin{adjustbox}{max width=\linewidth}\begin{tabular}{lrrrrr}
\toprule
& & \multicolumn{2}{c}{G5r} & \multicolumn{2}{c}{G5c} \\ \cmidrule(lr){3-4} \cmidrule(lr){5-6}
Head & Parameters & Full & Isolated & Full & Isolated \\
\midrule
1F & 105,006 & $54.96\pm0.36$ & $56.45\pm1.00$ & $63.19\pm2.21$ & $68.18\pm13.17$ \\
1F-wide & 121,550 & $54.94\pm0.63$ & $55.54\pm1.05$ & $65.19\pm2.08$ & $66.48\pm7.51$ \\
2F & 122,034 & $97.90\pm0.22$ & $99.68\pm0.13$ & $96.86\pm0.54$ & $99.60\pm0.09$ \\
Rotary & 123,090 & $52.44\pm0.92$ & $54.00\pm1.02$ & $88.57\pm3.40$ & $99.56\pm0.25$ \\
GRU & 153,770 & $98.29\pm0.14$ & $99.66\pm0.08$ & $97.84\pm0.22$ & $99.58\pm0.16$ \\
\bottomrule
\end{tabular}\end{adjustbox}}

\end{table}

The short-budget and width controls comprise 896 fits: 384 short-budget architecture cells, 160 additional context-dose cells, 320 additional-backbone cells, and 32 full-budget GRU width controls. Short-budget cells use 2,000 training and 480 validation examples and 512 updates. These runs use different budgets from the full-budget Transformer/Mamba controls.

\begin{table}[!htbp]\centering\small
\caption{\textbf{Short-budget architecture controls.} Each architecture receives the same 512-update regime. Mamba-2-style decay tests the update alone; the full-budget Mamba-1 comparison uses the complete architecture.}
\label{tab:short}\fitsuppcolumn{{\begin{adjustbox}{max width=\linewidth}\begin{tabular}{lrrrr}
\toprule
& \multicolumn{2}{c}{G5r} & \multicolumn{2}{c}{G5c} \\ \cmidrule(lr){2-3} \cmidrule(lr){4-5}
Head & Full & Isolated & Full & Isolated \\
\midrule
1F & 49.32 & 50.34 & 50.21 & 50.00 \\
1F-wide & 49.82 & 50.60 & 49.01 & 49.66 \\
2F & 51.90 & 95.26 & 50.57 & 87.53 \\
Rotary & 49.56 & 50.94 & 49.64 & 75.55 \\
GRU & 53.62 & 85.57 & 49.84 & 53.88 \\
GRU (matched) & 50.34 & 78.23 & 50.81 & 50.49 \\
LSTM & 50.26 & 60.65 & 50.18 & 49.17 \\
Causal Transformer & 49.30 & 48.83 & 49.97 & 50.26 \\
Mamba-2-style decay & 49.97 & 50.55 & 51.82 & 50.91 \\
Positive-step DeltaNet & 49.87 & 50.03 & 50.21 & 50.31 \\
Positive-step Gated DeltaNet & 49.40 & 49.58 & 49.51 & 49.22 \\
3F & 52.94 & 94.82 & 54.95 & 88.49 \\
\bottomrule
\end{tabular}\end{adjustbox}}}

\end{table}

The full-budget comparisons include 32 Transformer fits, 16 full-budget Mamba-1 fits, 128 frozen-encoder head adaptations, 16 matched encoder-adaptation runs, and eight learning-rate runs. These 200 fits reuse the existing block structure. Jobs ran on two RTX A5000 GPUs; raw probabilities, configurations, validation histories and checkpoints are saved. Frozen replay also records the encoder and checkpoint hashes.

The Mamba head contains two residual pre-LayerNorm blocks of width 80 and state size 16, followed by normalization and task/state readouts. The official Mamba-1 class and official differentiable reference scan are used with source provenance and license preserved. Causal convolution, input-dependent $\Delta$, $B$, and $C$, skip $D$, and input gate are all present. A numerical gate compares the sequence forward with repeated official step calls; gradients through the input must be finite and nonzero. Runtime describes the differentiable reference implementation; the optimized fused CUDA kernel has a different cost profile.

\begin{table}[!htbp]
\centering\small
\caption{\textbf{Capability and context sensitivity at the full head-training budget.} F/F and I/I train and test on full and isolated features. The channel-panel columns compare original (F) and donor (D) views of the same programs within block. BA and flips are percentages; $\Delta$ is in percentage points. $B$ gives the block count. Bold values identify the strong learned controls and Rotary contrasts discussed in the text. The fixed decoder--executor uses symbol supervision and the exact grammar as a recovery witness. $\dagger$ marks the four interfaces in the forty-contrast frozen family. Parameters exclude the encoder; Table~\ref{tab:archsensitivity} reports common-four-block comparisons.}
\label{tab:full-capability-matrix}
\begin{adjustbox}{max width=\linewidth}\begin{tabular}{llrrrrrrrr}
\toprule
& & \multicolumn{2}{c}{Long split BA} & \multicolumn{4}{c}{Channel panel (paired)} & & \\
\cmidrule(lr){3-4}\cmidrule(lr){5-8}
Grammar & System & F/F & I/I & F BA & D BA & $\Delta$ & Flips & $B$ & Params (k) \\
\midrule
G5c & 2F$^\dagger$ & 96.86 & 99.60 & 96.30 & \textbf{94.53} & $-1.77$ & 7.19 & 8 & 122.0 \\
 \rowcolor{resultorange} & Rotary$^\dagger$ & 88.57 & 99.56 & 88.54 & 72.86 & \textbf{$-15.68$} & 25.05 & 8 & 123.1 \\
 \rowcolor{resultblue} & GRU$^\dagger$ & 97.84 & 99.58 & 97.66 & \textbf{97.24} & $-0.42$ & 3.85 & 8 & 153.8 \\
 & GRU matched & -- & -- & 97.40 & 97.19 & $-0.21$ & 4.58 & 4 & 123.8 \\
 & Transformer & 66.95 & 75.15 & 66.51 & 49.17 & $-17.34$ & 48.49 & 8 & 128.2 \\
 & Mamba-1 & 61.49 & 83.84 & 61.56 & 55.21 & $-6.35$ & 32.60 & 4 & 132.3 \\
 & Fixed decoder + rules$^\dagger$ & -- & -- & 96.77 & 97.50 & $+0.73$ & -- & 8 & -- \\
\midrule
G5r & 2F$^\dagger$ & 97.90 & 99.68 & 98.18 & \textbf{98.02} & $-0.16$ & 3.18 & 8 & 122.0 \\
 \rowcolor{resultorange} & Rotary$^\dagger$ & 52.44 & 54.00 & \textbf{53.59} & 52.81 & $-0.78$ & 42.03 & 8 & 123.1 \\
 \rowcolor{resultblue} & GRU$^\dagger$ & 98.29 & 99.66 & 99.06 & \textbf{98.70} & $-0.36$ & 2.03 & 8 & 153.8 \\
 & GRU matched & -- & -- & 98.23 & 98.12 & $-0.10$ & 3.02 & 4 & 123.8 \\
 & Transformer & 52.77 & 54.09 & 51.30 & 53.12 & $+1.82$ & 38.07 & 8 & 128.2 \\
 & Mamba-1 & 66.13 & 71.82 & 65.83 & 64.27 & $-1.56$ & 21.77 & 4 & 132.3 \\
 & Fixed decoder + rules$^\dagger$ & -- & -- & 97.40 & 97.97 & $+0.57$ & -- & 8 & -- \\
\bottomrule
\end{tabular}\end{adjustbox}
\end{table}

\subsection{Complete context matrix on the channel panel}
\label{sec:supp-matrix}
\begin{figure}[!htbp]
\centering\includegraphics[width=\linewidth]{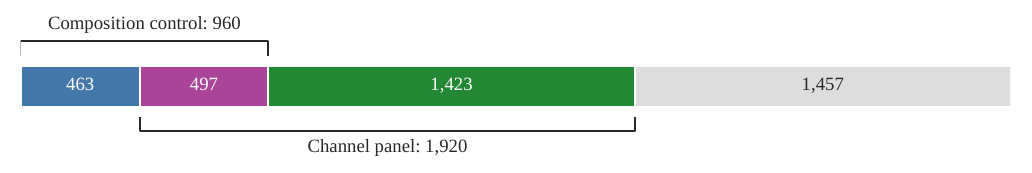}
\caption{\textbf{Overlap between the G5c source panels.} The bar partitions the 3,840 rows of the recovery panel. The brackets mark the composition-control (960 rows) and channel (1,920 rows) subsets. Their 497 shared source rows enter distinct interventions in the two panels.}
\label{fig:panel-lineage}
\end{figure}

The channel panel takes the first 40 examples per length/label stratum from each archived deterministic 480-row panel, retaining all eight blocks. The G5c rows contain 8--10 nonidentity operations, 0--3 identities and 10--13 total nodes. The separately generated panels with exactly ten nonidentity operations and one identity use a fixed 13-node composition. Six contexts, four learned heads, and the fixed rule interface are measured on every row. Table~\ref{tab:newmatrix} reports all five systems, including 1F. The four interfaces in the corrected frozen-comparison family are 2F, Rotary, GRU, and the symbol-supervised fixed decoder--executor. 1F is shown for reference and is not part of that family. Fig.~\ref{fig:panel-lineage} shows how the source panels overlap.

\begin{table}[!htbp]\centering\small\setlength{\tabcolsep}{3pt}
\caption{\textbf{Complete frozen-context matrix on the channel panel.} BA (mean $\pm$ block SD), eight blocks, 240 examples per block and grammar. F: original; V: operation values; S: state tiles; P: nonidentity permutation; P$^+$: all-operation permutation; D: matched donor.}
\label{tab:newmatrix}\fitsuppcolumn{\begin{adjustbox}{max width=\linewidth}\begin{tabular}{llrrrrrr}
\toprule
Task & System & F & V & S & P & P$^+$ & D \\
\midrule
G5c & 1F & 62.19 $\pm$ 3.83 & 49.38 $\pm$ 3.36 & 61.67 $\pm$ 5.90 & 60.94 $\pm$ 4.32 & 60.52 $\pm$ 3.29 & 49.27 $\pm$ 3.49 \\
G5c & 2F & 96.30 $\pm$ 1.33 & 94.74 $\pm$ 4.99 & 96.09 $\pm$ 1.41 & 96.88 $\pm$ 1.02 & 96.51 $\pm$ 0.80 & 94.53 $\pm$ 5.27 \\
G5c & Rotary & 88.54 $\pm$ 4.00 & 71.35 $\pm$ 5.04 & 87.86 $\pm$ 4.46 & 88.59 $\pm$ 3.91 & 88.80 $\pm$ 4.77 & 72.86 $\pm$ 5.02 \\
G5c & GRU & 97.66 $\pm$ 1.07 & 97.66 $\pm$ 1.41 & 97.71 $\pm$ 1.22 & 97.66 $\pm$ 0.97 & 97.97 $\pm$ 0.65 & 97.24 $\pm$ 0.99 \\
G5c & Fixed rules & 96.77 $\pm$ 1.39 & 97.14 $\pm$ 0.79 & 96.88 $\pm$ 0.92 & 97.29 $\pm$ 1.20 & 97.34 $\pm$ 1.16 & 97.50 $\pm$ 1.51 \\
G5r & 1F & 53.07 $\pm$ 2.54 & 52.08 $\pm$ 2.35 & 51.09 $\pm$ 2.57 & 53.80 $\pm$ 2.51 & 54.01 $\pm$ 3.04 & 51.46 $\pm$ 3.77 \\
G5r & 2F & 98.18 $\pm$ 1.59 & 96.20 $\pm$ 2.02 & 97.97 $\pm$ 1.50 & 98.07 $\pm$ 1.39 & 98.23 $\pm$ 0.94 & 98.02 $\pm$ 0.43 \\
G5r & Rotary & 53.59 $\pm$ 2.29 & 51.56 $\pm$ 3.60 & 52.08 $\pm$ 2.12 & 52.19 $\pm$ 3.23 & 50.42 $\pm$ 1.89 & 52.81 $\pm$ 4.09 \\
G5r & GRU & 99.06 $\pm$ 0.86 & 96.51 $\pm$ 1.43 & 99.11 $\pm$ 0.75 & 98.33 $\pm$ 0.89 & 98.49 $\pm$ 0.86 & 98.70 $\pm$ 0.98 \\
G5r & Fixed rules & 97.40 $\pm$ 0.91 & 95.73 $\pm$ 0.99 & 97.19 $\pm$ 0.73 & 97.34 $\pm$ 1.30 & 97.81 $\pm$ 1.04 & 97.97 $\pm$ 0.57 \\
\bottomrule
\end{tabular}\end{adjustbox}}

\end{table}

In V, every surrounding nonidentity operation is changed to a different operation using the original rendering seed; the queried tile is restored. The identity mask and operation-region length remain fixed. In S, start and endpoint states are shifted by fixed nonzero offsets before restoring the queried tile. In P, only surrounding nonidentity tiles move. In P$^+$, all surrounding operation-region tiles may move, including identities; nonidentity order can also change. P and P$^+$ preserve the full pixel multiset but can alter spatial statistics. D uses a cyclic donor assignment within split, node count, carrier, and vocabulary. Singleton strata yield seven self-donor programs in G5c and five in G5r. Dropping them changes the G5c Rotary mean donor effect by less than 0.1 pp. This donor seed differs from that of the earlier four-block estimate. Latent feature statistics can covary across these interventions.

\subsection{Statistics and multiplicity}
BA equals the average of sensitivity and specificity. Each new frozen panel is exactly balanced within every block, so overall accuracy equals BA. We recompute scores using the classwise definition. On the channel panel, paired changes use 20,000 whole-block bootstrap resamples. Exact two-sided sign-flip tests enumerate $2^B$ assignments and assume independent blocks with sign-symmetric contrasts under the null. The frozen family comprises 40 contrasts: five perturbations against F, the four interfaces 2F/Rotary/GRU/fixed decoder--executor, and two grammars. 1F and the later Transformer, Mamba-1 and width-matched GRU comparisons are not in this family and are reported separately. The adaptation family comprises 12 contrasts across ERM, output consistency, and CTC against augmentation for both heads and grammars. The direct frozen family compares the Rotary and 2F donor changes and the Rotary V and S effects for each grammar. The phase-restoration family contains two comparisons with full D. The equal-radius family has six V-versus-P contrasts over three heads and two grammars. Holm adjustment is applied within each family after the relevant block sets are complete. Complete unrounded effects, intervals, standardized paired effects $d_z$, and raw/adjusted $p$ values accompany the analysis record. The 95\% bootstrap intervals apply to individual contrasts. Holm correction controls the comparison family, so its rejection threshold can differ.

With four blocks, even unanimous nonzero differences cannot give a two-sided exact $p<0.125$. With eight, the minimum is 0.0078125. Cohen's paired $d_z$ is reported when the block-difference SD is positive, alongside uncertainty in percentage points.

\subsection{Recovery-panel replay and composition controls}
\label{sec:supp-legacy}
The recovery-panel replay contains 480 images per block per grammar. The first four blocks use cached earlier predictions and the other four are reconstructed with the same protocol. We hold the encoder, feature scaling, readouts, and decision threshold fixed. Paired effects weight blocks equally. The replay intervals resample blocks and, within block, connected components linking source and donor programs. Selection of the jointly exact subset is specific to each pair of views. It conditions on decoding outcomes. G5r donor effects are small where Rotary is close to chance. The 2F and GRU donor changes on G5c are $-1.82$ and $-0.55$ pp, respectively.

The G5c composition-control panel contains 960 archived source images, 120 per block. It preserves each queried tile and its position. The composition intervention changes surrounding nonidentity operation values while preserving their total increment modulo five and their identity positions. The permutation control preserves the operation multiset. Same-symbol rerendering changes local rendering draws without changing the surrounding symbols. Three draws are averaged per source for random controls and a single donor is used for the donor condition. All 316,212 query audits pass. The maximum anchor probability discrepancy is 0.00170, with no decision changes. Interval estimates resample the eight block means using 10,000 draws. The channel-panel matrix (Table~\ref{tab:newmatrix}) uses a different donor assignment and different perturbation definitions.

Two- and six-edit direction tests attain 50.87\% (95\% CI [48.78, 52.99]) and 51.98\% ([49.38, 54.58]) ordering accuracy, below the 65\% target. A separate five-candidate test uses 60 archived source images per block. Original-view endpoint accuracy is 79.38\% ([75.63, 82.71]), below the 80\% capability criterion. The fixed-sum result shows sensitivity to composition, although the internal rule is not identified. The composition-control and channel panels are stratified subsets of the recovery panel.

\subsection{Training budgets and full-encoder controls}
Table~\ref{tab:budgets} lists the training and adaptation regimes.
\begin{table}[!htbp]
\centering\small
\caption{\textbf{New training and adaptation regimes.} Counts refer to completed fits on the existing blocks. The earlier 160 full-budget fits and 896 follow-up fits are separate. Joint encoder and Mamba controls use four blocks.}
\label{tab:budgets}
\setlength{\tabcolsep}{4pt}
\fitsuppcolumn{\begin{adjustbox}{max width=\linewidth}\begin{tabular}{lrrrrl}
\toprule
Regime & Blocks & Train / validation & Updates & Fits & Selection and initialization\\
\midrule
Transformer, two grammars/two contexts & 8 & 16k / 6k & 5,040 & 32 & Original validation, new head\\
Mamba-1, two grammars/two contexts & 4 & 16k / 6k & 5,040 & 16 & Original validation, new head\\
Two-head adaptation, two grammars/four arms & 8 & 2k / 480 & 256 & 128 & Mean F/D validation, original head\\
DINOv2 joint adaptation, G5c/four arms & 4 & 2k / 480 & 256 & 16 & F validation, original encoder and head\\
Encoder LR sensitivity, G5c/two rates & 4 & 2k / 480 & 256 & 8 & F validation, original encoder and head\\
\bottomrule
\end{tabular}\end{adjustbox}}
\end{table}
\begin{table}[!htbp]
\centering\small
\caption{\textbf{Full DINOv2 adaptation on G5c.} BA (mean $\pm$ block SD), four blocks, 240 matched evaluation programs per block. Frozen and Joint use clean training; Joint+Aug.\ and Joint+CTC perturb one randomly chosen queried feature per training example. Evaluation perturbs every queried view. All arms share 256 updates and clean-validation selection.}
\label{tab:e2e}
\fitsuppcolumn{\begin{adjustbox}{max width=\linewidth}\begin{tabular}{lrr}
\toprule
Adaptation & F & D \\
\midrule
Frozen & 88.54 $\pm$ 0.80 & 71.04 $\pm$ 6.71 \\
Joint clean & 50.83 $\pm$ 1.02 & 51.35 $\pm$ 1.97 \\
Joint + Aug. & 51.88 $\pm$ 1.54 & 50.73 $\pm$ 1.15 \\
Joint + CTC & 51.77 $\pm$ 2.53 & 52.08 $\pm$ 2.57 \\
\bottomrule
\end{tabular}\end{adjustbox}}

\end{table}
At encoder LR $10^{-5}$, the joint clean arm reaches 50.83\% on F and 51.35\% on D. Its mean selected validation BA is 51.61\%, compared with 96.41\% for the frozen-encoder arm. All 21.63M encoder parameters are trainable, with nonzero gradient checks at the patch projection and first and last attention blocks. At this validation accuracy, the small F-to-D gaps of the joint arms are not informative.

Before adaptation, the differentiable encoder and archived head achieve 96.25\% on validation. Its validation BA matches the cached-feature score in all four blocks. The rate control keeps the head schedule fixed. Encoder LR $10^{-7}$ gives 96.98\% validation BA, 87.92\% on F, and 72.19\% on D. Encoder LR $10^{-6}$ gives 98.12\% validation BA, 88.23\% on F, and 73.75\% on D. Both rates preserve validation performance and retain a donor penalty in this adaptation regime.

Joint adaptation updates all DINOv2 parameters and the rotary head. Gradients are checked in the patch embedding and first and last attention blocks. The encoder learning rate is $10^{-5}$, the head rate is $3\times10^{-4}$, and batch size is 16. For paired training, one random queried feature is replaced by the feature from a donor canvas that preserves that tile. This keeps training tractable while evaluation uses a donor view for every queried node. All arms start from the existing checkpoints; Table~\ref{tab:e2e} reports four blocks.

\begin{table}[!htbp]
\centering\small
\caption{\textbf{Full-budget architecture controls.} Long-chain BA (mean $\pm$ block SD). F/F and I/I train and test in the same context. Transformer has eight blocks. Mamba-1 has four. Both use 5,040 updates, unlike the earlier 512-update controls.}
\label{tab:architectures}
\fitsuppcolumn{\begin{adjustbox}{max width=\linewidth}\begin{tabular}{llrr}
\toprule
Task & Head & F/F & I/I \\
\midrule
G5c & Transformer & 66.95 $\pm$ 2.58 & 75.15 $\pm$ 9.60 \\
G5c & Mamba-1 & 61.49 $\pm$ 1.35 & 83.84 $\pm$ 3.37 \\
G5r & Transformer & 52.77 $\pm$ 0.64 & 54.09 $\pm$ 0.85 \\
G5r & Mamba-1 & 66.13 $\pm$ 1.34 & 71.82 $\pm$ 0.49 \\
\bottomrule
\end{tabular}\end{adjustbox}}

\end{table}
Under the original 5,040-update schedule, the G5c isolated Transformer and Mamba-1 heads reach 75.15\% and 83.84\%. They have 128,170 and 132,282 trainable parameters, respectively. Table~\ref{tab:architectures} separates this architecture comparison from the earlier 512-update controls.

\begin{table}[!htbp]\centering\small
\caption{\textbf{Architecture sensitivity on a shared panel.} Four common blocks per task and 240 examples per block. F BA and D BA are percentages, and the donor change is in percentage points with 95\% block-bootstrap intervals. The decision flip rate is the percentage of paired answers that differ.}
\label{tab:archsensitivity}\fitsuppcolumn{\begin{adjustbox}{max width=\linewidth}\begin{tabular}{llrrrr}
\toprule
Task & Head & F BA & D BA & $\Delta$ BA (95\% CI) & Decision flips \\
\midrule
G5c & 1F & 61.04 & 49.06 & $-$11.98 [$-$15.62, $-$8.54] & 41.77 \\
G5c & Rotary & 88.96 & 72.19 & $-$16.77 [$-$17.50, $-$15.94] & 25.10 \\
G5c & 2F & 96.56 & 93.65 & $-$2.92 [$-$9.06, +0.94] & 7.29 \\
G5c & GRU & 97.40 & 97.40 & +0.00 [$-$1.25, +1.46] & 3.96 \\
G5c & GRU (width matched) & 97.40 & 97.19 & $-$0.21 [$-$1.67, +1.25] & 4.58 \\
G5c & Transformer & 65.00 & 49.38 & $-$15.62 [$-$20.62, $-$11.04] & 48.54 \\
G5c & Mamba-1 & 61.56 & 55.21 & $-$6.35 [$-$10.21, $-$3.13] & 32.60 \\
G5r & 1F & 52.29 & 52.81 & +0.52 [$-$2.29, +2.71] & 39.69 \\
G5r & Rotary & 52.81 & 53.02 & +0.21 [$-$2.29, +2.92] & 41.25 \\
G5r & 2F & 97.19 & 97.81 & +0.62 [$-$0.42, +1.77] & 4.17 \\
G5r & GRU & 98.65 & 98.23 & $-$0.42 [$-$1.88, +1.04] & 2.71 \\
G5r & GRU (width matched) & 98.23 & 98.12 & $-$0.10 [$-$0.83, +1.35] & 3.02 \\
G5r & Transformer & 50.62 & 54.48 & +3.85 [$-$0.94, +10.00] & 36.56 \\
G5r & Mamba-1 & 65.83 & 64.27 & $-$1.56 [$-$3.44, +0.00] & 21.77 \\
\bottomrule
\end{tabular}\end{adjustbox}}

\end{table}
On the shared four-block context panel, G5c Transformer goes from 65.00\% to 49.38\%, and Mamba-1 goes from 61.56\% to 55.21\%. Table~\ref{tab:archsensitivity} reports paired effects and decision flips for the same 240 programs per block. G5r Transformer changes from 50.62\% to 54.48\% and flips 36.56\% of decisions. The width-matched GRU retains high accuracy in both contexts.

The complete Mamba-1 comparison uses the validated implementation described above. The comparison measures capability at the recorded schedule and model sizes.

\subsection{Training context versus test context}
In the earlier G5c panel, neighboring operations are more decodable from full-context features than from isolated features (58.00\% versus 23.58\%). Metadata alone yields 23.49\%. Yet training and testing Rotary with isolated features improves length extrapolation. An input-to-phase map can respond strongly to information from surrounding nodes that is unnecessary once the ordered local operations are supplied. The operation-value and phase interventions test this account within the fitted systems.

Removing context at inference produces a different result. The full-trained Rotary head drops from 89.27\% to 50.57\%, and fixed decoding with execution drops to 67.71\% on those isolated views. The frozen models now receive a feature distribution they did not train on. The same transformation affects symbol recovery as well as learned computation. This distinction motivates testing adaptation with paired views while measuring performance on the original input.

The endpoint controls show why net accuracy and per-example sensitivity can differ. A previous endpoint probe ranks changes at 99.04\%, while prefix exchange changes Rotary BA from 87.34\% to 87.40\%. There are 82 corrections and 80 new errors. The near-zero net change and the small state-tile effect argue against an explanation based only on endpoint leakage.

\section{Mechanism, Theory, and Same-Example Evidence}
\label{sec:supp-mechanism}
\subsection{Affine identities}
\label{sec:supp-identities}
The identities below describe the implemented recurrences exactly and show how paired inputs enter the state and readout.
For one delta factor $S\mapsto SA_j+B_j$, $A_j=I-\beta_j k_jk_j^\top$ and $B_j=\beta_jv_jk_j^\top$. Two sequential factors give
\begin{equation}
S\mapsto S A_1A_2+B_1A_2+B_2.
\end{equation}
This order follows the implementation's right-multiplication convention. Rotary updates use planar blocks
\begin{equation}
R(\theta)=\begin{pmatrix}\cos\theta&\sin\theta\\-\sin\theta&\cos\theta\end{pmatrix}
\end{equation}
and a positive sigmoid decay. Keys and queries are unit-normalized. The four matrix memories start at zero, and inactive padded steps leave them unchanged.

Index the $n$ encoded nodes from start to candidate endpoint. For any two input sequences, subtracting their affine recurrences yields
\begin{align}
\delta_t&=S_{t-1}^DA_t^D+B_t^D-S_{t-1}^FA_t^F-B_t^F\\
&=\delta_{t-1}A_t^D+
 S_{t-1}^F(A_t^D-A_t^F)+(B_t^D-B_t^F).
\end{align}
Writing the final two terms as $E_t$ and substituting recursively proves the main text's finite transport identity. No differentiability assumption is needed. Its norm bound follows from the triangle inequality and $\|XA\|_F\leq\|X\|_F\|A\|_2$. For a product of delta factors, each unit-key factor has eigenvalues $1$ and $1-\beta$ with $\beta\in[0,2]$, so its spectral norm is at most one. Rotary has norm $\rho$. Both 2F and Rotary are therefore nonexpansive, so the bound alone does not rank their robustness. The bound discards innovation directions and their cancellation. A large state distance can leave the decision unchanged, whereas a small distance can cross a boundary when the classification margin is small. The query, residual and nonlinear readout must be considered to relate a state perturbation to a decision.

For a planar rotation, direct multiplication gives
\begin{equation}
(R_D-R_F)^\top(R_D-R_F)
=4\sin^2(\Delta\theta/2)I.
\end{equation}
Across planar blocks, the spectral norm is the largest block norm. If $\rho_D\geq\rho_F$, write the transition difference as $(\rho_D-\rho_F)R_D+\rho_F(R_D-R_F)$; otherwise use the symmetric decomposition. Thus
\[
\|A_D-A_F\|_2\leq|\rho_D-\rho_F|+
2\min(\rho_D,\rho_F)\max_j|\sin(\Delta\theta_j/2)|.
\]
Angles are wrapped with $\operatorname{atan2}(\sin\Delta,\cos\Delta)$ for measurement.

If $a_j=\|A_j^D\|_2$ and $w_t=\prod_{j>t}a_j$, Cauchy--Schwarz further gives
\begin{equation}
\|\delta_n\|_F\leq\left(\sum_t w_t^2\right)^{1/2}
\left(\sum_t\|E_t\|_F^2\right)^{1/2}.
\end{equation}
This motivates a squared innovation penalty. Our normalized penalty differs from this unnormalized bound, and a changing query/readout requires additional terms to obtain a logit bound. The readout-aware extension below supplies explicit constants and measures bound tightness. The effect of minimizing these innovations is evaluated through task accuracy and paired decision stability.

\paragraph{Signed-margin decomposition.}
For Rotary, use the hybrid transition $A_t^*=\rho_t^D R(\theta_t^F)$. The three local terms are
\begin{align}
E_t^\theta&=S_{t-1}^F(A_t^D-A_t^*), &
E_t^\rho&=S_{t-1}^F(A_t^*-A_t^F),\\
E_t^B&=B_t^D-B_t^F .
\end{align}
Propagating each term through the remaining donor transitions gives three contributions whose sum equals $S_n^D-S_n^F$. The norm of their sum can be smaller than the sum of their norms. An angle difference contributes in proportion to the stored state in its rotation plane. The contributions have directions, and later transitions can rotate and attenuate them. The accumulated difference therefore depends on direction and propagation as well as chain length. To decompose the final logit margin, evaluate the actual nonlinear readout after adding the three state contributions in a fixed order, then substitute the query and residual. Consecutive logit differences telescope. Individual terms depend on the substitution order, and a decision changes only when the signed margin crosses zero.

\subsection{From update transport to a decision condition}
\label{sec:readout-transport-bound}
The affine identity concerns memory, whereas the implemented head reads that
memory through an input-dependent query and a residual path. We make these
interfaces explicit. For the final valid node (including the candidate), let
\begin{align}
 r^v &= \mathop{\mathrm{concat}}_h(S_{n,h}^v q_{n,h}^v),\\
 o^v &= W_o\operatorname{LN}(r^v)+b_o+u_n^v,\\
 z^v &= [o^v;u_n^v],\qquad
 m^v = f_1(z^v)-f_0(z^v),
 \label{eq:actual-readout}
\end{align}
where $v\in\{F,D\}$ and $f$ is the shared two-layer GELU classifier.
This is the checkpoint's actual readout: the residual enters both the
normalized-memory branch and the final concatenation. All comparisons use
the same parameters in evaluation mode.

For multiple heads, use the joint Frobenius norm
$\|S\|_F^2=\sum_h\|S_h\|_F^2$ and
$a_t=\max_h\|A_{t,h}^D\|_2$. Set
\begin{align}
 U &= \sum_{t=1}^n \|E_t\|_F\prod_{j=t+1}^n a_j,\\
 Q_D &= \max_h\|q_{n,h}^D\|_2,\\
 J &= \left[\sum_h \|S_{n,h}^F\|_2^2
          \|q_{n,h}^D-q_{n,h}^F\|_2^2\right]^{1/2},\\
 V &= Q_D U+J,\qquad d_u=\|u_n^D-u_n^F\|_2.
\end{align}
Normalized queries satisfy $Q_D\leq1$. Let $L_N$ be a Lipschitz constant
for LayerNorm on the line segment joining the two actual read vectors, and
let $L_m$ be a Lipschitz constant for the scalar margin $f_1-f_0$.

\paragraph{Proposition (readout-aware decision condition).}
For the deterministic shared head in Eq.~\eqref{eq:actual-readout},
\begin{align}
 \|r^D-r^F\|_2 &\leq V,\\
 |m^D-m^F| &\leq C,\\
 C &= L_m\sqrt{\big(\|W_o\|_2L_N V+d_u\big)^2+d_u^2}.
 \label{eq:readout-certificate}
\end{align}
If $|m^F|>C$, the two binary decisions agree. If the original decision is
also correct, the donor decision is correct. This is a sufficient condition;
its failure does not imply a changed or incorrect decision.

\paragraph{Proof.}
For each head, the exact read difference is
\begin{equation}
 S_{n,h}^Dq_{n,h}^D-S_{n,h}^Fq_{n,h}^F
 =\delta_{n,h}q_{n,h}^D
  +S_{n,h}^F(q_{n,h}^D-q_{n,h}^F).
\end{equation}
The triangle inequality, the memory transport bound, and the spectral norm
bound for the second term give $\|r^D-r^F\|_2\leq Q_DU+J$.
Thus $\|o^D-o^F\|_2\leq\|W_o\|_2L_NV+d_u$ and
$\|z^D-z^F\|_2^2=\|o^D-o^F\|_2^2+d_u^2$.
The margin Lipschitz inequality proves Eq.~\eqref{eq:readout-certificate}.
A change of sign requires $|m^D-m^F|\geq|m^F|$, which proves the
strict decision condition. No assumption about labels, correct decoding,
training convergence, or the donor distribution is used.

\paragraph{Constants for the implemented head.}
Write the classifier as $f(z)=W_2\operatorname{GELU}(W_1z+b_1)+b_2$.
For the exact GELU used here, $\sup_x|\operatorname{GELU}'(x)|<1.13$, so
\begin{equation}
 L_m=1.13\,\|(W_2)_{1,:}-(W_2)_{0,:}\|_2\|W_1\|_2
\end{equation}
is valid. LayerNorm with learned scale $\gamma$ and numerical constant
$\epsilon_N$ has the global bound
$\|\gamma\|_\infty/\sqrt{\epsilon_N}$. A tighter segment bound is
\begin{equation}
 L_N=\frac{\|\gamma\|_\infty}
 {\sqrt{\epsilon_N+\min_{s\in[0,1]}\|c_F+s\Delta c\|_2^2/d}},
\end{equation}
where $c_F$ and $\Delta c$ are the mean-centered original read and read
difference, and $d$ is the read dimension. The minimum is a scalar quadratic
minimum, attained at
$s=\operatorname{clip}_{[0,1]}(-c_F^\top\Delta c/\|\Delta c\|_2^2)$
when $\Delta c\ne0$; when $\Delta c=0$ the denominator is constant.
The LayerNorm Jacobian norm is at most the reciprocal normalization
denominator times $\|\gamma\|_\infty$, which proves this bound along
the segment. These constants are evaluated on observed pairs and are not a
certificate against every possible image edit.

\paragraph{Numerical tightness of the bound.}
On the retained 48-pair length-10 G5c fixture from block 1, the maximum
elementwise transport reconstruction error is $1.2\times10^{-6}$.
The median ratio of observed to bounded displacement is $0.240$ for memory,
$0.0484$ for the final read, and $0.000510$ for the margin bound in
Eq.~\eqref{eq:readout-certificate}. Even the segment-specific LayerNorm
constant certifies none of the 48 pairs; using the exact observed read
distance instead of $V$ also certifies none. The median bound is
$1.54\times10^4$ logit units, versus an original absolute margin of $13.87$.
The certificate is valid but vacuous on this fixture. It does, however, show which terms the decision depends on.

\paragraph{State and decision invariance under CTC.}
If every $E_t=0$ and the initial memories agree, induction gives
$S_t^D=S_t^F$ at every step. Transition--write cancellation can therefore
preserve the state exactly; it is not intrinsically a failure. If the final queries and
residuals also agree, the full decisions agree. In contrast, zero CTC alone
does not imply equal decisions. At the affine readout interface, the scalar
example $S_n^F=S_n^D=1$, $q_n^F=1$, $q_n^D=-1$, and margin $m=S_nq_n$
has $E_t=0$ but opposite decisions. The implemented head additionally
contains an unconstrained residual-to-classifier path. The scalar example
illustrates the missing assumption, not a measurement on a fitted model.
An example within the implemented architecture is also possible: set the
value projection to zero so that both memories stay zero, and let the
classifier use one coordinate of $u_n$. Two input projections giving that
coordinate opposite signs change the decision while every $E_t=0$.
A two-unit GELU pair can implement this linear margin exactly because
$\operatorname{GELU}(x)-\operatorname{GELU}(-x)=x$. This construction establishes a possible failure mode within the
architecture; the fitted checkpoint's errors require empirical diagnosis.

Normalization introduces a second qualification. For one chain, write
$\eta_t=\|S_{t-1}^F\|_F^2+\|B_t^F\|_F^2+\epsilon$ and
$\ell_{\rm CTC}=n^{-1}\sum_t\|E_t\|_F^2/\eta_t$.
If $\eta_t\leq M$ for every node, then
\begin{equation}
 U\leq \left(\sum_t\left[\prod_{j>t}a_j\right]^2\right)^{1/2}
        \sqrt{Mn\ell_{\rm CTC}}.
\end{equation}
For nonexpansive transitions this is at most
$n\sqrt{M\ell_{\rm CTC}}$. Thus a small mean normalized penalty,
without scale control, query/residual control, and an adequate decision
margin, is not a robustness guarantee. This bound permits length-dependent
worst cases; it does not predict monotonic empirical degradation.

\paragraph{A targeted training implication.}
The readout condition motivates testing consistency of the actual final
memory read $r=S_nq_n$ together with $u_n$, in addition to the paired task
loss. This covers the query and residual terms omitted by CTC while
allowing transition--write cancellation that preserves the computed state.
The resulting readout consistency objective is a falsifiable training
hypothesis. Its effect on accuracy requires comparison with ordinary
augmentation and output consistency under matched tuning and compute,
including the clean--donor trade-off. Because the proof is pair-specific, transfer
to new donor mechanisms and longer chains remains an empirical question.

\subsection{State-direction controls in 2F and GRU}
\label{sec:review-direction}
We reuse the previously evaluated 1,920-row DINOv2 full-budget test, with fixed 2F and GRU checkpoints and no fitting. For 2F, hold the original final query and residual fixed and set $\Delta r=(S^D-S^F)q^F$. The local projection $p=\nabla_r m^F\cdot\Delta r$ approximates the state-only margin change. The flip-ranking score is $-\operatorname{sign}(m^F)(m^F+p)$: it combines the original margin with the projected change. Norm ranking uses relative state displacement within that head; original uncertainty uses $-|m^F|$.

Table~\ref{tab:review-direction} compares decision-flip rankings with projection-error measurements.

\begin{table}[!htbp]
\centering\small
\caption{\textbf{Flip rankings and projection errors.} Rankings of observed paired decision flips and mean absolute projection errors in logit units. Entries average eight block metrics on the same 1,920 evaluated rows. The projection score uses the original margin and both paired views. The state-only 2F intervention retains the original query/residual; full change also includes their actual changes. State magnitudes are interpreted within each architecture.}
\label{tab:review-direction}
\begin{adjustbox}{max width=\linewidth}\begin{tabular}{lrrrrr}
\toprule
Head & State norm AUROC & Original uncertainty AUROC & Projected margin AUROC & State-only error & Full-change error \\
\midrule
2F & .702 & .770 & .916 & 4.83 & 5.59 \\
GRU & .792 & .766 & 1.000 & $1.28\times10^{-6}$ & $1.28\times10^{-6}$ \\
\bottomrule
\end{tabular}\end{adjustbox}

\end{table}

For 2F, the projection has mean block Spearman correlation .765 with the exact state-only margin change, but its 4.83-logit average absolute error remains substantial. Curvature and the omitted query/residual changes make the combined score a paired association rather than a predictor of errors or repairs. GRU's linear final classifier makes its projection an exact reconstruction of margin change up to numerical error, so its perfect flip ranking checks the algebra.

\subsection{Recoverability and the learned decision}
Let $d$ be a fixed role decoder and $\mathcal E$ the exact executor. If every token is recovered, then $\mathcal E(d(Z))$ is correct. Its error event is therefore contained in the union of token errors, giving $\Pr(\text{rule error})\leq\sum_t\Pr(\text{token}_t\text{ error})$. This implication is one-way: compensating start and endpoint mistakes can preserve legality. The decoder-executor accuracy is therefore a lower bound on what this interface can achieve.

A simple counterexample shows that recovery alone does not constrain the learned head. Let $Z$ be the complete one-hot token sequence, so decoding is perfect. A learned function that always outputs the opposite of the correct answer has error one. The empirical value of the comparison is consequently the coexistence of a fixed successful decoding procedure and errors of a particular learned system under matched inputs. The channel interventions then ask how that system's computation changes.

\subsection{Numerical and channel checks}
The diagnostic reconstructs each tested matrix head from its actual projection, convolution, normalized keys/queries, transition factors, writes, normalization, residual, and classifier. Its factual probability must agree with the archived head forward within $2\times10^{-5}$. For every paired sequence, propagated innovations must reconstruct the state difference within $2\times10^{-4}$ in maximum absolute element error; the norm bound must hold to that tolerance. The saved audit reports actual residuals, per-sequence bounds and the pass threshold.

Replacing $A$, write $B$, query $q$, or residual $u$ alone defines different synthetic networks at inference time. Rotary phase and decay swaps separate the two factors in $A$. Other channels stay original in those substitutions. In the reverse experiment only the original phase is restored within D. Both CPU replays are checked against the archived GPU predictions before restoring phase. For each of F and D, all decisions match at the threshold across 240 images per grammar and block, with two grammars and eight blocks. The cross-device tolerance is $10^{-4}$ on probability, and the within-device transport check retains $2\times10^{-5}$. The swaps measure conditional responses; Table~\ref{tab:final-four-cell} gives the complete background--phase comparison. Interactions can prevent their BA changes from adding to the full donor effect. We also record feature distance, phase RMS, mean decay displacement, final-output CKA, state distance, local innovation norms and propagated upper bounds.

\begin{table}[!htbp]\centering\small
\caption{\textbf{Background--phase four-cell diagnostic.} The 1,920-pair DINOv2 G5c channel panel, eight fixed blocks. Each row fixes all non-phase channels to its recipient view. Original and donor phase replace both operation and candidate phase. BA is a block-equal mean; Diff.\ F counts prediction disagreements with original F. The design measures conditional effects of phase with other channels fixed.}
\label{tab:final-four-cell}
\begin{adjustbox}{max width=\linewidth}\begin{tabular}{llrr}\toprule
Non-phase channels & Phase source & BA (\%) & Diff.\ F \\\midrule
Original features & Original phase & 88.54 & 0 \\
Original features & Donor phase & 72.19 & 470 \\
\rowcolor{resultblue}Donor features & Original phase & \textbf{88.54} & \cellcolor{resultorange}\textbf{142} \\
Donor features & Donor phase & 72.86 & 481 \\
\bottomrule\end{tabular}\end{adjustbox}\end{table}

\paragraph{Phase replay and angular magnitude.}
On the retained 48-program fixture, the candidate swap has the largest median single-node margin displacement, whereas restoration over all operation nodes repairs more new errors. On the two earlier fresh 384-pair panels, donor-phase replay ranks decision flips with mean block AUROCs .88 and .92, compared with .56 and .51 for angular RMS. These are paired, model-dependent scores that use both views. Seven of eight blocks in the 11-operation panel are below the 80\% clean capability gate, so these rankings mainly describe low-capability models.

\subsection{Same-example separation on the four-block panel}
\begin{figure}[!htbp]\centering
\includegraphics[width=\linewidth]{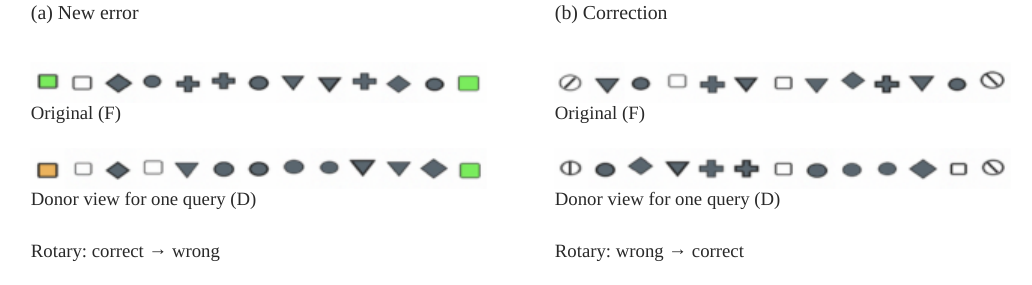}
\caption{\textbf{Paired examples from actual predictions.} The archived selection takes the first qualifying new error and correction in the first block, with correct rule-interface answers in both contexts. A displayed donor view illustrates one query; the model answer uses the full sequence of queried views. The examples illustrate the construction; paired panels quantify effect prevalence.}
\label{fig:examples}
\end{figure}

The earlier four-block 480-row G5c panel adds joint outcome counts. Its Rotary BA changes from 89.27\% to 71.61\% under D, while the rule interface changes from 96.61\% to 96.88\%. Of 417 newly incorrect Rotary answers, 388 have correct rule-interface answers in both contexts. The rule-stable set covers 1,817/1,920 examples and also contains 72 of 78 Rotary corrections. Table~\ref{tab:paired} gives the full accounting. These examples overlap with the eight-block panel and are reported separately; Fig.~\ref{fig:examples} illustrates actual paired decisions. The channel panel additionally counts exact recovery of every displayed token in both contexts by rerunning the fixed ridge decoder. We check its legality outputs against the previously archived answers.

\begin{table}[!htbp]\centering\small
\caption{\textbf{Joint outcomes on the four-block panel.} Four blocks, 480 programs per block. Losses are correct-to-incorrect transitions and corrections are the reverse. Rule-stable programs are answered correctly in both contexts.}
\label{tab:paired}\fitsuppcolumn{{\begin{adjustbox}{max width=\linewidth}\begin{tabular}{lrrrrr}
\toprule
 & \multicolumn{2}{c}{Rotary} & & \multicolumn{2}{c}{Among rule-stable examples} \\
\cmidrule(lr){2-3}\cmidrule(lr){5-6}
Block & Losses & Corrections & Rule-stable & Losses & Corrections \\
\midrule
1 & 83 & 19 & 455 & 79 & 17 \\
2 & 127 & 28 & 458 & 120 & 27 \\
3 & 103 & 18 & 456 & 95 & 17 \\
4 & 104 & 13 & 448 & 94 & 11 \\
\midrule
Total & 417 & 78 & 1,817 & 388 & 72 \\
\bottomrule
\end{tabular}\end{adjustbox}}}

\end{table}

\section{Training Controls and Generalization Boundaries}
\subsection{Matched head adaptation}
\label{sec:supp-adaptation}
The four adaptation arms start from the same original full-context checkpoint and fixed normalization, using the same 2,000 training and 480 validation rows. Donors are selected within each split to avoid leakage. ERM receives two original views, while the other arms receive original and donor views. Task and state losses are averaged across views. Output consistency measures squared Euclidean distance between the two class-probability vectors, and transport consistency uses the normalized local affine innovation. Both weights are fixed at 0.1 without a test-driven sweep.

All arms use batch 128, 256 updates, AdamW learning rate $3\times10^{-4}$ with cosine decay, weight decay 0.01, and gradient clipping at one. The checkpoint maximizing the mean of clean and donor validation BA at 32-update intervals is retained, using the earliest maximum on ties. The ERM arm uses the same criterion. Table~\ref{tab:adaptation} reports all original and donor scores. The other pixel interventions are included in the source predictions. Runtime includes the additional penalty computation and is reported by arm.

\begin{table}[!htbp]\centering\small
\caption{\textbf{Complete adaptation comparison.} BA (mean $\pm$ block SD), eight blocks. Paired contrasts hold head, grammar, original checkpoint, and test images fixed.}
\label{tab:adaptation}\fitsuppcolumn{{\begin{adjustbox}{max width=\linewidth}\begin{tabular}{lllrr}
\toprule
Task & Head & Adaptation & F & D \\
\midrule
G5c & Rotary & Clean ERM & 88.28 $\pm$ 4.05 & 72.76 $\pm$ 4.95 \\
G5c & Rotary & Augmentation & 84.84 $\pm$ 4.53 & 77.92 $\pm$ 4.09 \\
G5c & Rotary & Output consistency & 84.90 $\pm$ 4.81 & 77.92 $\pm$ 4.44 \\
G5c & Rotary & CTC & 84.17 $\pm$ 5.22 & 78.07 $\pm$ 4.07 \\
G5c & 2F & Clean ERM & 96.09 $\pm$ 1.46 & 93.85 $\pm$ 5.47 \\
G5c & 2F & Augmentation & 95.05 $\pm$ 1.73 & 94.06 $\pm$ 4.39 \\
G5c & 2F & Output consistency & 95.42 $\pm$ 1.73 & 93.75 $\pm$ 4.32 \\
G5c & 2F & CTC & 95.16 $\pm$ 1.82 & 94.11 $\pm$ 4.34 \\
G5r & Rotary & Clean ERM & 53.54 $\pm$ 2.81 & 52.50 $\pm$ 4.36 \\
G5r & Rotary & Augmentation & 53.07 $\pm$ 4.45 & 53.59 $\pm$ 3.64 \\
G5r & Rotary & Output consistency & 53.18 $\pm$ 3.69 & 53.49 $\pm$ 3.81 \\
G5r & Rotary & CTC & 53.44 $\pm$ 4.18 & 53.91 $\pm$ 3.56 \\
G5r & 2F & Clean ERM & 98.33 $\pm$ 1.34 & 98.18 $\pm$ 0.31 \\
G5r & 2F & Augmentation & 97.92 $\pm$ 1.34 & 98.02 $\pm$ 0.83 \\
G5r & 2F & Output consistency & 97.76 $\pm$ 1.34 & 97.92 $\pm$ 1.04 \\
G5r & 2F & CTC & 98.07 $\pm$ 1.20 & 98.02 $\pm$ 0.70 \\
\bottomrule
\end{tabular}\end{adjustbox}}}

\end{table}

\begin{figure}[!htbp]
\centering\includegraphics[width=\linewidth]{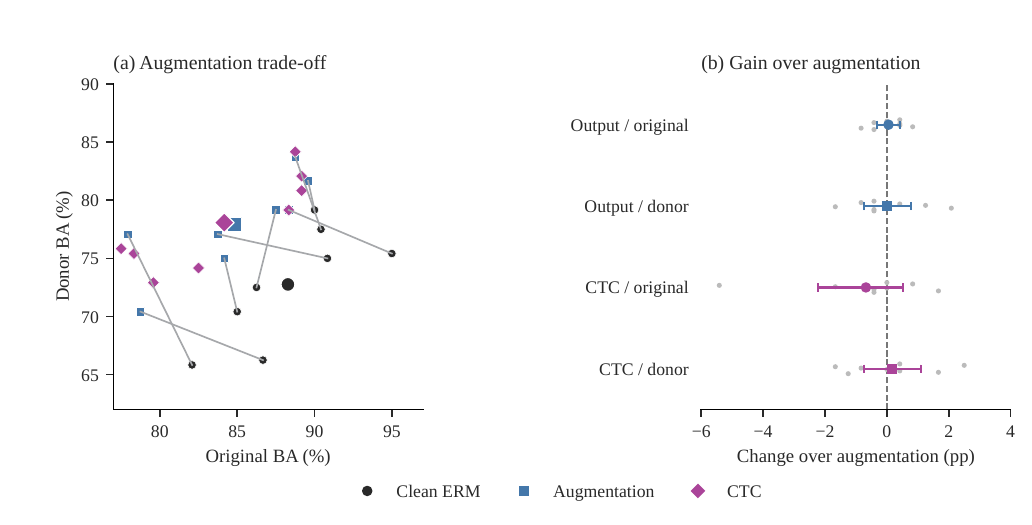}
\caption{\textbf{Original matched adaptation comparison.} Revised G5c Rotary test: 1,920 pairs, eight blocks. (a) Absolute original and donor BA; lines connect clean ERM to augmentation within each block, small marks show blocks and large marks show means. (b) Paired changes relative to ordinary augmentation; gray dots are block differences and bars are paired 95\% block-bootstrap intervals. CTC donor change is $+0.16$ pp, CI $[-0.73,+1.09]$. Arms share training samples and update counts; penalty computation differs.}
\label{fig:adaptation}
\end{figure}

\paragraph{Paired accuracy contrasts.}
For G5c Rotary, the clean and donor changes of ordinary augmentation relative to ERM have 95\% block-bootstrap intervals of [$-5.68$, $-1.20$] and [$3.44$, $7.19$] pp, respectively. The clean change of CTC relative to augmentation is $-0.68$ pp ([$-2.24$, $0.52$]). These are per-contrast intervals. For the donor gain of ordinary augmentation, the two-sided sign-flip $p$ is 0.0078 before adjustment and 0.0938 after Holm adjustment over the twelve donor contrasts. Figure~\ref{fig:adaptation} shows the observed trade-off and the paired effects relative to augmentation.

\subsection{Angle consistency as a training objective}
At a fixed step and memory, let $S=S^F_{t-1}$. Expanding the update difference gives
\begin{equation}
E=S\big[\rho_D(R_D-R_F)+(\rho_D-\rho_F)R_F\big]
  +B_D-B_F.
\label{eq:s-phase-innovation}
\end{equation}
The first term inside the brackets is exactly the transition difference between D and D with the original rotation restored. This decomposition uses the original trajectory state and attributes local differences at that state. It allows cancellation. For example, take $S=e_1e_1^\top$ in two dimensions, $R_F=I$, $R_D=R(\pi/2)$, and $\rho_F=\rho_D=0.8$. Set $B_F=0$ and $B_D=-0.8S(R_D-R_F)$. The donor write has rank one and admits the rotary write parameterization. Then $E=0$ despite a quarter-turn phase difference. This allowed update preserves the next state when the incoming states agree. If every innovation is zero from the common zero initialization, the entire states agree. Cancellation therefore does not explain a failure of state preservation; the example only shows that angle agreement is unnecessary. A small local innovation at one step does not erase differences accumulated earlier. Query and residual changes can still alter the final decision, as shown by the readout analysis.

For a planar rotation, $\operatorname{tr}(R_F^\top R_D)=2\cos(\theta_D-\theta_F)$ and $\|R_F\|_F^2=\|R_D\|_F^2=2$. Expanding $\|R_D-R_F\|_F^2$ gives $4[1-\cos(\theta_D-\theta_F)]$. The term vanishes for angles a full turn apart. It controls the rotation independently of memory energy, decay, or writes. Supervised losses still determine which rotations are useful. A constant angle predictor would satisfy the angular penalty, so accuracy and sensitivity to genuine operation edits are necessary checks.

\paragraph{Prospective comparison.}
Use paired augmentation as the main comparator, with output consistency and CTC as controls. Match the initial heads, training views, and updates. Choose regularizer weights on validation data under comparable tuning budgets. Lock independent test programs and their donor assignments before analysis. Report paired angular discrepancy and both original and donor BA, including whether clean long-chain performance meets a prespecified tolerance. Include a genuine edit to the queried operation that changes the correct answer as a control for task sensitivity. The operation-only mask applies to the training penalty. Frozen substitutions act on all valid nodes. The separate development screen below reports matched tests of this objective.

\subsection{Full-encoder gradient and selection checks}
The full-encoder control starts from the same pretrained DINOv2-S/14 weights and archived full-budget rotary checkpoint in every arm. Frozen and Joint use original views, while Joint+Augmentation and Joint+CTC create one random node-preserving donor canvas per training example. Only the corresponding queried feature is substituted into the original sequence. All auxiliary labels remain those of the original program. At evaluation, each queried feature is obtained from its own donor canvas, matching the node-wise diagnostic logic. Training and test donor assignments use a separate deterministic seed from the frozen factorial experiment.

The implementation obtains differentiable spatial patch features directly from the encoder with gradients enabled. It checks nonzero gradients in the patch projection and attention projections of the first and last transformer blocks. Every encoder parameter is trainable in the joint arms. AdamW uses encoder rate $10^{-5}$ and head rate $3\times10^{-4}$, batch 16, 256 updates, cosine decay, and joint gradient clipping at one. Checkpoints are selected on clean validation BA every 64 updates, identically across arms.

\subsection{Architecture and learning-rate sensitivity}
The replay comparison uses the 240-row shared panel within blocks 1--4 for each grammar. For Transformer, Mamba-1, and the parameter-matched GRU, the selected checkpoint first reproduces archived probabilities on the original feature cache. It then receives the rerendered factorial features, using its fixed training normalization. Table~\ref{tab:archsensitivity} reports both absolute BA and within-block donor change for these same programs. The flip rate counts any original/donor decision disagreement, including both new errors and corrections. This matters when a low-accuracy model has a small mean BA change.

Low early validation scores under encoder LR $10^{-5}$ motivated the smaller-rate control. The eight additional runs keep the same initial weights, clean batches, head schedule, and eligible checkpoint steps. We trace validation BA and feature drift at steps 0, 8, 32, 64, 128, 192, and 256. The early observations describe the trajectory, while only multiples of 64 are eligible for checkpoint selection. Table~\ref{tab:encoderrate} lists all rates.
At step 8, the mean raw feature RMSE relative to initialization is 0.070 at $10^{-7}$ and 0.508 at $10^{-6}$. This measures early representation drift in the two lower-rate arms. The original $10^{-5}$ run records validation from step 64.

\begin{table}[!htbp]\centering\small
\caption{\textbf{Encoder-rate sensitivity for joint clean adaptation.} Four G5c blocks. Validation is the block mean, while test BA is mean $\pm$ block SD. The frozen arm uses zero encoder updates.}
\label{tab:encoderrate}\fitsuppcolumn{\begin{adjustbox}{max width=\linewidth}\begin{tabular}{lrrr}
\toprule
Encoder LR & Validation & F & D \\
\midrule
Frozen & 96.41\% & 88.54 $\pm$ 0.80 & 71.04 $\pm$ 6.71 \\
$10^{-7}$ & 96.98\% & 87.92 $\pm$ 2.01 & 72.19 $\pm$ 7.82 \\
$10^{-6}$ & 98.12\% & 88.23 $\pm$ 3.03 & 73.75 $\pm$ 7.75 \\
$10^{-5}$ & 51.61\% & 50.83 $\pm$ 1.02 & 51.35 $\pm$ 1.97 \\
\bottomrule
\end{tabular}\end{adjustbox}}

\end{table}

\paragraph{Equal-radius feature directions.}
Let $v=Z^V-Z^F$ and $p=Z^P-Z^F$ after the head's fixed training normalization. For each program we take $r=\min(\|v\|_F,\|p\|_F)$ over all valid nodes and replay $Z^F+rv/\|v\|_F$ and $Z^F+rp/\|p\|_F$. A zero displacement produces the original features. Both interpolations stay between the original and perturbed vectors. They equalize aggregate displacement and can differ in the distribution of changes over nodes or coordinates. We check their achieved radii and original-context decision agreement. Because the interpolation takes place in feature space, this comparison concerns synthetic feature sequences.

\subsection{Decoder choice and regularization}
The additional G5c decoder audit uses blocks 1--4, a common cap of 12,000 training and 6,000 validation tokens per role, and original-context training-role normalization. Ridge regularization spans $10^{-5}$, $10^{-3}$, 0.1, 10 and 1000. Logistic regression and linear SVM choose $C\in\{0.01,1,100\}$ on original validation accuracy. All ridge strengths are reported; selection uses original validation accuracy. The rules and role information are held fixed across decoder families.
Table~\ref{tab:decoderablation} shows accuracy for all decoder variants. The numerical retry increases the iteration cap from 1,000 to 10,000 and keeps the original samples and validation selection. All selected logistic and SVM fits converged in the numerical retry.

\begin{table}[!htbp]\centering\small
\caption{\textbf{Fixed rule execution by decoder family and regularization.} Mean BA $\pm$ block SD over four G5c blocks. Logistic and linear SVM hyperparameters use original validation data. All five ridge strengths are reported.}
\label{tab:decoderablation}\fitsuppcolumn{\begin{adjustbox}{max width=\linewidth}\begin{tabular}{lrrrr}
\toprule
Decoder / ridge $\alpha$ & F & D & V & S \\
\midrule
$10^{-5}$ & 95.73 $\pm$ 0.92 & 96.77 $\pm$ 2.08 & 97.19 $\pm$ 1.24 & 96.56 $\pm$ 0.98 \\
$10^{-3}$ & 96.04 $\pm$ 0.87 & 97.08 $\pm$ 1.70 & 97.19 $\pm$ 1.24 & 96.56 $\pm$ 0.92 \\
$10^{-1}$ & 94.58 $\pm$ 1.23 & 95.73 $\pm$ 2.05 & 95.52 $\pm$ 0.62 & 95.10 $\pm$ 0.98 \\
$10$ & 70.00 $\pm$ 2.95 & 72.71 $\pm$ 1.05 & 70.62 $\pm$ 3.31 & 73.02 $\pm$ 1.99 \\
$10^3$ & 49.27 $\pm$ 2.67 & 49.27 $\pm$ 1.91 & 49.48 $\pm$ 1.81 & 49.90 $\pm$ 2.21 \\
Logistic & 96.67 $\pm$ 1.23 & 97.71 $\pm$ 0.87 & 97.92 $\pm$ 0.68 & 97.40 $\pm$ 0.71 \\
Linear SVM & 96.98 $\pm$ 0.52 & 97.71 $\pm$ 1.42 & 98.02 $\pm$ 0.63 & 96.87 $\pm$ 1.10 \\
\bottomrule
\end{tabular}\end{adjustbox}}

\end{table}

\subsection{Longer computational streams}
We additionally sample 240 fresh programs at each of 10, 20, 30, and 50 nonidentity operations, for both grammars and each of the eight trained blocks. This yields 15,360 streams. Each start, operation, and endpoint feature is sampled by its true role and symbol from the archived held-out isolated-token bank, with the exact source row and node retained. Correct endpoints are computed by the grammar. Balanced negative endpoints differ from that endpoint. The original isolated-trained heads and fixed isolated-trained decoders are reused without adaptation.

\begin{figure}[!htbp]\centering\includegraphics[width=\linewidth]{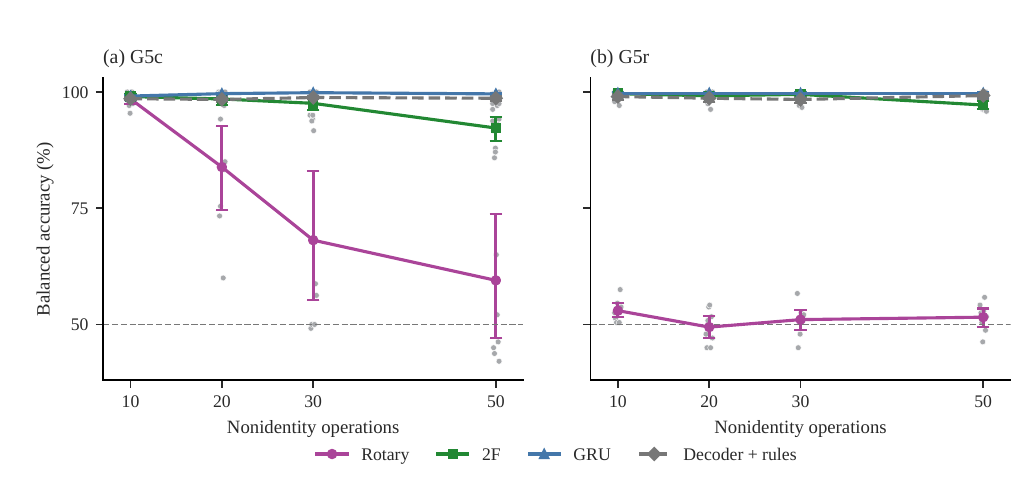}
\caption{\textbf{Length stress on recomposed held-out features.} Each point summarizes 240 programs in each of eight trained blocks. Bars show 95\% block-bootstrap intervals. Panels show G5c and G5r. Gray dots show all eight block values; colored marks show their means. Small offsets separate block observations. The dashed line marks binary chance at 50\%; Probe + rules denotes the measured decoder--executor.}
\label{fig:longstream}
\end{figure}

The test uses held-out DINOv2 features encoded in isolation. Their retained absolute pixel positions can vary nonmonotonically along the new sequence. The length-10 row controls for this changed construction before the longer streams test sequence computation.
On G5c, Rotary scores 98.49\% at 10 operations and 59.48\% at 50. At 50 operations, the 2F head, GRU, and fixed rules reach 92.24\%, 99.58\%, and 98.65\%. Fig.~\ref{fig:longstream} includes the intermediate lengths and the order-sensitive grammar. The contrasting curves are evidence about computation on recomposed features.

\subsection{Context dose, probes, and calibration}
The earlier dose study retains the queried node plus radius $k=0,1,2,4$, full context, or donor neighbors within radius two. Each cell uses the same short budget. On G5c, isolated and radius-two donor features yield nearly identical neighbor decoding (23.62\% and 23.63\%), but Rotary BA is 76.77\% and 48.39\%. Matched probe scores therefore coexist with sharply different task performance.

Neighbor probes include current-symbol, position, and node-count metadata. Full-context G5c features support mean BA 58.00\% for offsets $\pm1$, versus 23.58\% for isolated features and 23.49\% for metadata alone. Endpoint change ranking uses 1,500 matched pairs per block. Prefix exchange uses a separate 240-pair subset. Its 82 corrections and 80 losses explain how a negligible net accuracy change can coexist with individual decision flips.

Validation-selected scalar thresholds change original full-context Rotary long-chain BA from 88.57\% to 88.50\%. A test-optimal scalar threshold, selected retrospectively as an oracle diagnostic, reaches only 89.07\% on average. Full-budget original-width and parameter-matched isolated GRUs remain approximately 99.6\% on both tasks. Table~\ref{tab:backbones} lists the additional frozen-backbone controls; these use shared programs and a fixed projection to 384 input dimensions when required.

\begin{table}[!htbp]\centering\small
\caption{\textbf{Frozen-backbone controls at the short budget.} The backbone comparisons jointly vary architecture, pretraining and spatial grids.}
\label{tab:backbones}\fitsuppcolumn{\begin{adjustbox}{max width=\linewidth}\begin{tabular}{lrrrrrrrr}
\toprule
& \multicolumn{4}{c}{G5r} & \multicolumn{4}{c}{G5c} \\ \cmidrule(lr){2-5} \cmidrule(lr){6-9}
& \multicolumn{2}{c}{Neighbor} & \multicolumn{2}{c}{2F} & \multicolumn{2}{c}{Neighbor} & \multicolumn{2}{c}{Rotary} \\
Encoder & Full & Isol. & Full & Isol. & Full & Isol. & Full & Isol. \\
\midrule
DINOv2-S/14 & 61.93 & 29.51 & 53.02 & 94.53 & 56.72 & 23.62 & 48.70 & 76.77 \\
DINOv2-B/14 & 61.55 & 29.54 & 79.90 & 92.14 & 55.50 & 23.70 & 58.59 & 78.80 \\
CLIP ViT-B/16 & 56.57 & 29.62 & 56.88 & 99.79 & 49.33 & 23.63 & 52.60 & 91.41 \\
SigLIP 2 ViT-B/16 & 47.68 & 29.62 & 69.48 & 99.38 & 41.04 & 23.55 & 57.40 & 86.98 \\
ResNet-50 & 95.26 & 29.86 & 49.17 & 88.65 & 92.95 & 23.62 & 50.31 & 72.34 \\
\bottomrule
\end{tabular}\end{adjustbox}}

\end{table}

\section{Visual Vocabulary, Legends, and VLM Controls}
\label{sec:supp-vocab}
\subsection{Capability gates and scope}
Table~\ref{tab:review-scope} lists the capability and transfer limits that bound the mechanism claims.

\begin{table}[!htbp]
\centering\small
\caption{\textbf{Capability and scope audit.} Every observation appears in the corresponding numerical tables.}
\label{tab:review-scope}
\begin{adjustbox}{max width=\linewidth}\begin{tabular}{@{}p{.21\linewidth}p{.43\linewidth}p{.28\linewidth}@{}}
\toprule
Panel & Recorded capability or result & Interpretation boundary \\
\midrule
Independent 11-operation test & Original Rotary BA 65.10\%; seven of eight blocks fail the 80\% gate. Phase restoration is $+4.95$ pp, with a zero-crossing interval. & Falls below the prespecified capable-model mechanism criterion. \\
10-operation plus identity follow-up & Original BA 77.60\%; the follow-up was added after the 11-operation screen. It changes operation count and identity presence together. & Post-result, length-matched diagnostic with operation count and identity presence jointly changed. \\
New operation glyphs and refits & Frozen and refitted heads are near chance. Refit training BA is 86.25--97.50\%, but validation BA is 45.00--55.00\%. & Recognition and task learning remain obstacles. Ranges describe the two observed blocks. \\
Endpoint and operation prototypes & Symbolic repairs can execute correctly, while learned prototype-input paths remain near chance and fail the fresh-panel gates. & True-ID class means are privileged inputs and change the head's feature distribution; richer readouts could still recover task information. \\
Consistency training & Earlier incremental CTC donor effect over augmentation is $+0.16$ pp, with 95\% interval $[-0.73,+1.09]$. Later locked comparisons retain their own samples and budgets. & The smaller discrepancy accompanies an uncertain incremental BA effect. \\
Local phase and all-node CNN & The two-block phase pilot misses its later expansion criterion. The all-node local head is exactly invariant but has validation BA 52.53\% and 64.03\%. & Invariance without task capability is insufficient. These validation and engineering screens fall below the capability criteria. \\
Small open VLMs & Symbolic-input forced-choice accuracy is already near binary chance. The separate structured Sonnet panel has stronger but uneven capability. & Near-chance symbolic execution limits visual-context attribution. The separate API follow-up measures behavior at the input/output interface. \\
\bottomrule
\end{tabular}\end{adjustbox}

\end{table}

\subsection{Vocabulary changes and legends}
An unfamiliar glyph vocabulary changes the mapping from pixels to operation identities. A decoder trained on the original mapping has no way to recover a new one without extra information. The prior legend controls therefore distinguish an uninformative renaming from a labeled legend that specifies the intended mapping. The rule executor knows the grammar but cannot infer an unknown glyph mapping from the task label alone.

The legend audit includes independent 11-step programs, matched incorrect bindings, fixed feature/decoder/head checkpoints, and saved per-query predictions. These controls therefore test recovery under a supplied correspondence and supplied decoder role labels.

\subsection{Bounded open-model comparisons}
The prior comparison uses Qwen3-VL-2B~\cite{bai2025qwen3vl} and SmolVLM2-500M~\cite{marafioti2025smol}, evaluated locally on the same 320 programs. Each model receives full images, ordered node tiles, or ground-truth symbolic text, with a common legend and transition table: 960 outputs per model, 1,920 total. The official scored protocol constrains a Yes/No single-token choice. Table~\ref{tab:vlm} gives the results.

\begin{table}[!htbp]\centering\small
\setlength{\tabcolsep}{1.7pt}
\caption{\textbf{Constrained-output VLM comparison.} Both models use the same balanced programs. The Sym. column uses symbolic text, where accuracy is already near chance.}
\label{tab:vlm}\fitsuppcolumn{\begin{adjustbox}{max width=\linewidth}\begin{tabular}{llrrrrrr}
\toprule
& & \multicolumn{3}{c}{Accuracy (\%)} & \multicolumn{3}{c}{Yes responses (\%)} \\ \cmidrule(lr){3-5} \cmidrule(lr){6-8}
Model & Task & Full & Tiles & Sym. & Full & Tiles & Sym. \\
\midrule
SmolVLM2-500M & G5r & 50.00 & 49.38 & 50.00 & 0.00 & 4.38 & 0.00 \\
SmolVLM2-500M & G5c & 50.00 & 52.50 & 50.00 & 0.00 & 6.25 & 0.00 \\
Qwen3-VL-2B & G5r & 50.00 & 49.38 & 52.50 & 0.00 & 8.12 & 31.25 \\
Qwen3-VL-2B & G5c & 50.00 & 49.38 & 51.25 & 0.00 & 8.12 & 12.50 \\
\bottomrule
\end{tabular}\end{adjustbox}}

\end{table}

An early free-response trial allowed only eight generated tokens and often ended before producing a scorable answer. SmolVLM2's default image splitting also produced 89 requests beyond its configured 8,192-token limit. The scored rerun disabled splitting. The near-chance results concern that constrained interface.

\paragraph{VLM capability.}
These small models are near chance even on symbolic input, so their scores say little about visual context. A useful VLM test would first establish execution on symbolic input and on ordered tiles, and only then apply a label-preserving context intervention to full canvases. The structured follow-up in Sec.~\ref{sec:structured-vlm} uses a separate model, parser and program panel; it yields six wrong decisions among 58 exactly transcribed full-canvas chains (Table~\ref{tab:structured-vlm}).

\subsection{Structured multimodal follow-up}
\label{sec:structured-vlm}
We evaluate the model identified as Claude Sonnet 4.6 by the AICodeMirror API on 64 archived hue-carrier programs. The sample contains two programs per grammar, length (3 or 10 nonidentity operations), block (1--4), and legality label. Each program is presented as its full canvas, as an ordered strip of indexed node crops, and as ground-truth symbols. The two visual conditions include the same labeled legend. All three conditions supply the transition table, order and candidate convention. Input layouts and token counts differ. We use temperature zero and a 1,536-token output cap, and request the start, all operation IDs including identities, candidate, complete state trajectory, and binary decision. A separate four-program pilot fixed the JSON parser and output cap before testing. We accept one complete JSON object, including a final fenced object after prose. One symbolic request failed and counts as incorrect. No returned response hit the token cap.

\begin{table}[!htbp]
\centering\small
\caption{\textbf{Multimodal comparison with a common program panel.} Entries are counts out of $n$, except BA. Full and Tiles include a visual legend. Symbols supplies the true IDs. Exact chain means the reported start, every operation and candidate agree with ground truth. Exact trace compares every reported state. Rule applies the known transitions to the model's reported IDs.}
\label{tab:structured-vlm}
\setlength{\tabcolsep}{5pt}
\fitsuppcolumn{{\begin{adjustbox}{max width=\linewidth}\begin{tabular}{llrrrrrrr}
\toprule
Input & Grammar & Length & $n$ & Parsed & Chain & Trace & Rule & BA (\%) \\
\midrule
Full & G5c & 3 & 16 & 16 & 13 & 10 & 15 & 100.00 \\
Full & G5c & 10 & 16 & 16 & 13 & 4 & 15 & 56.25 \\
Full & G5r & 3 & 16 & 16 & 16 & 15 & 16 & 93.75 \\
Full & G5r & 10 & 16 & 16 & 16 & 9 & 16 & 100.00 \\
\addlinespace
Tiles & G5c & 3 & 16 & 16 & 13 & 12 & 14 & 81.25 \\
Tiles & G5c & 10 & 16 & 16 & 5 & 4 & 11 & 50.00 \\
Tiles & G5r & 3 & 16 & 16 & 16 & 14 & 16 & 100.00 \\
Tiles & G5r & 10 & 16 & 16 & 4 & 3 & 12 & 75.00 \\
\addlinespace
Symbols & G5c & 3 & 16 & 16 & 16 & 8 & 16 & 75.00 \\
Symbols & G5c & 10 & 16 & 16 & 16 & 11 & 16 & 87.50 \\
Symbols & G5r & 3 & 16 & 16 & 16 & 9 & 16 & 68.75 \\
Symbols & G5r & 10 & 16 & 15 & 15 & 6 & 15 & 68.75 \\
\midrule
Full & All & Both & 64 & 64 & 58 & 38 & 62 & 87.50 \\
Tiles & All & Both & 64 & 64 & 38 & 33 & 53 & 76.56 \\
Symbols & All & Both & 64 & 63 & 63 & 34 & 63 & 75.00 \\
\bottomrule
\end{tabular}\end{adjustbox}}}
\end{table}

On full canvases, 6 of 58 exactly reported chains receive an incorrect model decision, and 20 of those 58 have an incorrect reported trajectory. On true-symbol input the corresponding counts are 15 and 29 of 63 parsed cases. All parsed cases with both an exact chain and an exact trajectory yield a correct decision. Model decisions can be right despite an earlier state error, as in the long G5r full-canvas cell. The exact executor changes the evaluation question: for example, it scores 62/64 on the full-canvas transcriptions while the model itself scores 56/64. It evaluates transcribed symbols using the supplied exact execution algorithm. Moving from the canvas to tiles fixes five model decisions and breaks twelve; moving from the canvas to symbols fixes six and breaks fourteen. These results come from one model on a small fixed panel. Raw requests, responses and scoring code are saved with the release.

VSTAT~\cite{yu2026vstat} emphasizes continuous event perception and visual integration in videos, while MET-Bench~\cite{cohen2026met} compares visual and textual entity-state reasoning. Both benchmarks offer settings for testing the recoverability/execution distinction beyond the current generated-symbol tasks. Realistic traversal, ambiguous symbols, natural scene clutter, and unknown transition rules are left for future work.

\section{Independent Programs and Development Screens}
\label{sec:supp-independent}
\subsection{Separate recognition and execution tests}
Given an ordered list of queries, pair the original and changed views before scoring. The following evaluation uses a role-specific classifier and exact transition rules.
\begin{enumerate}\setlength{\itemsep}{1pt}
\item Replace the surroundings for each query and keep that tile's pixels and position. Encode and concatenate the queried features in the supplied order.
\item Decode the start, each operation, and the proposed endpoint. Execute the decoded program. Run the same executor with true tokens as a separate oracle.
\item Score the fitted head in both contexts. Count paired decision changes over all programs, then describe changes among pairs with both chains exact.
\item Recompute the head twice with exchanged phases. Use donor angles with original decay, writes, query and residual, then original angles with the corresponding donor channels.
\end{enumerate}

\subsection{Frozen mechanism on disjoint families}
The new G5c panels each contain 48 programs per block, one legal and one illegal endpoint for each of 24 canonical families, across eight blocks. We exclude all previously used families of the same length, canonicalized under operation renaming and reversal. A cyclic donor assignment deranges entire families within each block, matching the source label and keeping the query tile fixed. The 11-operation panel was specified first. Its capability gate was 80\% original BA per block. We then tested a separately generated 10-operation panel with one identity to match the original long-chain regime and keep the node count fixed. This follow-up changes both the number of operations and the presence of an identity.

\begin{figure}[!htbp]
\centering\includegraphics[width=\linewidth]{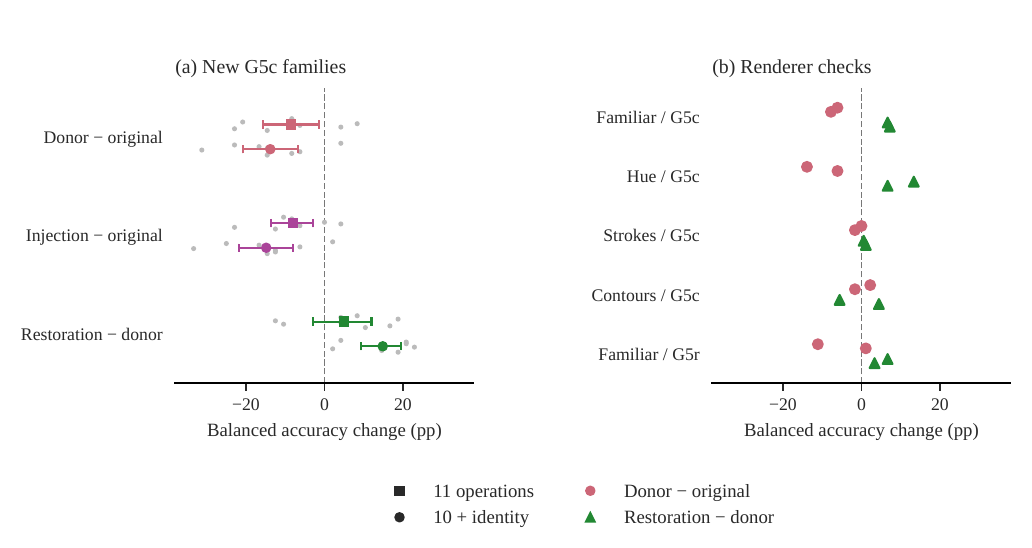}
\caption{\textbf{Paired mechanism across independent families and renderers.} (a) The 11-operation and 10-operation G5c panels each use eight blocks of 48 distinct programs. Bars show 95\% whole-block bootstrap intervals. The full-donor and donor-phase comparisons use original BA as baseline. Restoration uses full donor BA. (b) Distinct renderings of the two held-out program blocks use the same programs within each grammar. Each point is one block (180 G5c or 90 G5r programs). The two renderers with new symbols are strokes and contours. Absolute scores are reported in the adjacent table.}
\label{fig:independent-transfer}
\end{figure}

Table~\ref{tab:independent-mechanism} and Fig.~\ref{fig:independent-transfer} compare frozen interventions on the disjoint-family panels.

\begin{table}[!htbp]
\centering\small
\caption{\textbf{Frozen Rotary and non-Rotary controls on independent G5c families.} Accuracy is BA (\%). Differences are in percentage points, with 95\% whole-block bootstrap intervals in brackets. Each row uses 384 programs across eight blocks.}
\label{tab:independent-mechanism}
\setlength{\tabcolsep}{3pt}
\fitsuppcolumn{\begin{adjustbox}{max width=\linewidth}\begin{tabular}{lrrrrrrr}
\toprule
Operations & Original & Donor & \shortstack{Donor\\$-$ original} & \shortstack{Injection\\$-$ original} & \shortstack{Restoration\\$-$ donor} & \shortstack{2F donor\\effect} & \shortstack{GRU donor\\effect}\\
\midrule
11 & 65.10 & 56.51 & $-8.59$ [$-15.62$, $-1.30$] & $-8.07$ [$-13.54$, $-2.86$] & $4.95$ [$-2.86$, $11.98$] & -5.47 & -0.78 \\
10 + identity & 77.60 & 63.80 & $-13.80$ [$-20.83$, $-6.77$] & $-14.84$ [$-21.88$, $-8.07$] & $14.84$ [$9.38$, $19.53$] & -1.82 & -1.82 \\
\bottomrule
\end{tabular}\end{adjustbox}}
\end{table}

The original BA in the 11-operation panel ranges from 47.92\% to 81.25\%. Seven of the eight blocks fall below the capability gate. The panel therefore misses the prespecified mechanism criterion. Holm-adjusted two-sided sign-flip probabilities for the three effects in table order are 0.156, 0.094 and 0.281. Of 83 new donor errors, 66 occur with both decoded chains exact. The 10-operation follow-up has original BA 68.75--93.75\% across blocks. The maximum elementwise state reconstruction residual is $1.7\times10^{-6}$. Across these blocks, the median ratio of the net state displacement to the sum of the three component norms ranges from 0.63 to 0.69. The within-block correlation between angular RMS and the change in signed true-class margin ranges from $-0.25$ to $0.14$. The signs and magnitudes depend on the original state and on subsequent transitions. Both panels have fixed lengths within block.

\subsection{Transfer across state carriers and operation glyphs}
Table~\ref{tab:renderer-range} gives both blocks for each renderer and grammar.
The external two-block program sets use 11 nonidentity operations. They reuse each underlying program across the familiar rendering, a hue carrier, and two new operation glyph inventories. Each new glyph renderer changes the symbol inventory while the paired original-versus-donor test keeps the current queried glyph identical. The visual changes across renderers can lower original performance, and the two blocks provide a range of observations.

\begin{table}[!htbp]
\centering\small
\caption{\textbf{Renderer and grammar transfer.} Original (F) and donor (D) BA (\%) across renderer and grammar conditions. Each entry is the range across two blocks of 180 G5c or 90 G5r programs.}
\fitsuppcolumn{\begin{adjustbox}{max width=\linewidth}\begin{tabular}{llrrrrrr}
\toprule
Grammar & View style & Rotary F & Rotary D & 2F F & 2F D & GRU F & GRU D\\
\midrule
G5c & Familiar & 59.44--71.67 & 51.67--65.56 & 87.22--89.44 & 86.11--88.89 & 91.11--92.22 & 91.11--95.00 \\
G5c & Hue & 61.67--78.33 & 55.56--64.44 & 97.22--98.89 & 98.89--99.44 & 98.33--100.00 & 98.33--99.44 \\
G5c & Strokes & 47.22--52.78 & 45.56--52.78 & 48.89--51.11 & 48.33--53.33 & 46.11--48.89 & 47.22--50.00 \\
G5c & Contours & 43.89--50.00 & 46.11--48.33 & 46.67--47.22 & 50.00--52.78 & 44.44--51.11 & 44.44--53.89 \\
G5r & Familiar & 48.89--51.11 & 40.00--50.00 & 94.44--94.44 & 88.89--92.22 & 93.33--93.33 & 94.44--95.56 \\
G5r & Hue & 44.44--57.78 & 42.22--44.44 & 98.89--98.89 & 98.89--100.00 & 98.89--98.89 & 98.89--98.89 \\
G5r & Strokes & 44.44--52.22 & 50.00--58.89 & 41.11--48.89 & 47.78--58.89 & 45.56--51.11 & 48.89--52.22 \\
G5r & Contours & 52.22--54.44 & 44.44--50.00 & 47.78--50.00 & 50.00--54.44 & 46.67--54.44 & 47.78--51.11 \\
\bottomrule
\end{tabular}\end{adjustbox}}
\label{tab:renderer-range}
\end{table}

\subsection{Matched-budget training on disjoint programs}
\begin{figure}[!htbp]
\centering\includegraphics[width=\linewidth]{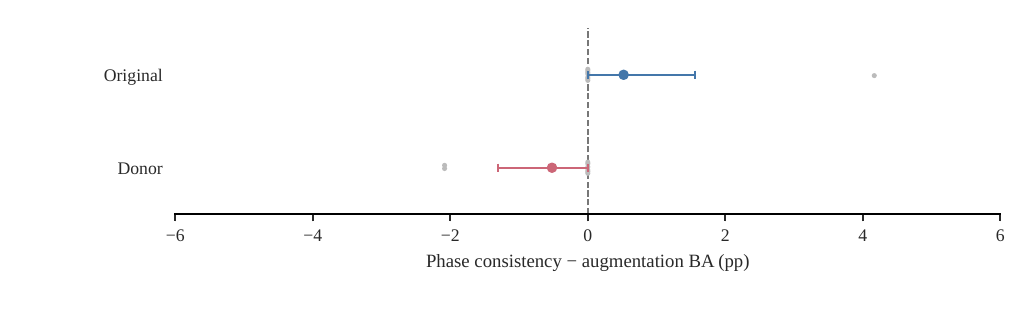}
\caption{\textbf{Independent accuracy after matched adaptation.} The baseline is ordinary augmentation. Gray dots denote individual block contrasts. Colored points and bars denote the eight-block mean and 95\% percentile block-bootstrap interval. Each block contributes 48 paired programs.}
\label{fig:training-confirmation}
\end{figure}

Figure~\ref{fig:training-confirmation} compares every trained objective with matched augmentation on the independent test.
We use the same initialization and sample order for each arm, with 80 training and 80 validation rows per block, two supervised forwards per update, batch size 32, and 64 AdamW updates. The validation criterion weights original and donor BA equally. Clean ERM, paired augmentation, output consistency, CTC and operation-node phase consistency are compared on the first two development blocks. In the phase penalty, the original-view angle supplies a stop-gradient target. The independent test compares phase consistency with augmentation on the 10-operation families in all eight blocks. The prespecified gates require a donor benefit of at least 2 points and an original penalty smaller than 1 point, using the respective lower confidence bounds.

Relative to augmentation, phase consistency changes original BA by $0.52$ [$0.00$, $1.56$] pp and donor BA by $-0.52$ [$-1.30$, $0.00$] pp. The joint criterion fails because the donor-gain lower bound does not reach $+2$ pp; the clean-capability requirement is met. The mean paired operation-angle RMS change is $-0.0057$ radians.

\subsection{Frozen locality and symbol interfaces}
A budget-matched visibility screen keeps the queried tile and either the two nearest or the two farthest tiles in their original positions. Both interventions use the same canvas, encoder, frozen heads, and number of visual forwards. The held-out clean-source Rotary BA in the near condition is 60.42\% and 54.17\% in the two development blocks, versus 89.58\% in the full original condition in each block. The far comparison scores 52.08\% and 54.17\%. The substantial absolute loss disqualifies this frozen masking method under the clean-accuracy gate.

Role-specific ridge decoders learned from the small development split use the same frozen features, with training on originals or matched original/donor pairs. The decoder-executor and learned recurrence can fail on different examples. The executor given true symbols obtains 100\% in the closed grammar. In a separate glyph transfer pilot, supervised decoding on one glyph inventory yields incomplete chains on unseen glyph assignments, even when the original grammar and executor are unchanged. All fits and per-condition recognition and decision scores are recorded with the experiment data.

In a secondary transfer screen, the two development checkpoint pairs yield donor gains from $-0.56$ to $2.08$ pp across the 11-operation and alternate-renderer conditions.

\paragraph{Operation-node phase consistency.}
For paired views, average over valid operation nodes $t$ (including identities) and rotation planes $j$:
\[
\mathcal L_{\theta}=\mathbb E_{i,t,j}
\left[1-\cos(\theta^F_{i,t,j}-\theta^D_{i,t,j})\right].
\]
The original angle is a stop-gradient target. In one plane the squared Frobenius distance between the rotations is four times the corresponding term. In two development blocks, phase consistency and matched augmentation yield identical original and donor BA, while operation-node phase RMS is lower by 0.012 and 0.001 radians. The independent eight-block test uses the selected training objective and the same matched augmentation baseline.

\paragraph{Both chains exact in the eight-block tests.}
In the 11-operation panel, 283 of 384 pairs meet joint exactness. Donor minus original BA is $-11.41$ pp [$-19.63$, $-3.01$], donor-phase injection is $-9.85$ pp [$-17.47$, $-3.12$], and restoration is $+6.48$ pp [$-1.29$, $14.08$]. Restoration corrects 45 of 66 new errors in that subset, and 48 of 283 decisions still differ from the original. In the 10-operation-plus-identity follow-up, 305 of 384 pairs meet joint exactness. The corresponding contrasts are $-16.14$ pp [$-24.54$, $-8.06$], $-15.85$ pp [$-24.35$, $-8.30$], and $+18.38$ pp [$10.78$, $25.04$]. Restoration corrects 63 of 77 new errors, and 46 of 305 decisions differ from the original. These intervals resample whole blocks.

\subsection{Target-renderer refit}
We train a separate Rotary head in each target operation-glyph renderer, initialized from its archived familiar-renderer checkpoint. DINOv2 and feature standardization are fixed. Each fit uses 80 clean training rows and 80 clean validation rows, sampled by label from the formal program split, with canonical families excluded between fitting, validation, and testing. AdamW runs for 64 updates at batch size 32, with initial learning rate $3\times10^{-4}$, cosine decay, and gradient clipping at 1. We select the earliest checkpoint with maximal validation BA among updates 16, 32, 48 and 64. The role decoder is independently fitted on the same 80 clean rows. The 11-operation test uses the independent glyph validation manifest; the 10-operation-plus-identity test uses blocks 1 and 2 of the length-matched follow-up. Both renderers share programs and donor identities within a block and length. States and identity tiles use the shared renderer.

Every zero-shot and refit block falls below the 80\% original-BA gate (Tables~\ref{tab:refit-all} and~\ref{tab:refit-exact}). The selected refits reach 86.25--97.50\% training BA but 45.00--55.00\% validation BA. Original BA is 48.33--50.56\% on the 11-operation tests and 41.67--52.08\% on the length-matched tests. These observations localize a failure of generalization under this refit budget even within the target renderer. The fixed zero-shot probe recovers no complete chains jointly. The refitted probe identifies 99.3--100\% of operations while start and endpoint recognition span 13.33--64.58\% and 27.78--56.25\%, respectively. It recovers both complete chains for only 1--23 programs per block. High operation recognition therefore coexists with poor full-chain recovery. The exact-subset estimates have small and sometimes single-label denominators. Phase substitutions change decisions, but with clean scores this low and few complete chains we draw no mechanism conclusion. Decay, write, query and residual controls in both directions and their per-block scores are in \texttt{renderer\_refit\_by\_block.csv}.

\begin{table}[!htbp]
\centering\small
\caption{\textbf{Operation-glyph shift, all held-out pairs.} F and D are original and donor BA (\%). $\Delta$F inserts donor phase into original channels; $\Delta$D restores original phase with other donor channels (pp). Flips counts original/donor decision changes. Rescue is the count corrected by restoration over original-correct/donor-wrong pairs. Diff counts restored decisions different from original. Each row is a block observation. I denotes identity.}
\label{tab:refit-all}
\fitsuppcolumn{\begin{adjustbox}{max width=\linewidth}\begin{tabular}{llrlrrrrrrrr}
\toprule
Renderer & Block & Ops & Head & $n$ & F & D & $\Delta$F & $\Delta$D & Flips & Rescue & Diff\\
\midrule
Contours & 1 & 11 & Zero-shot & 180 & 43.9 & 46.1 & 3.9 & $-$5.6 & 80 & 20/38 & 42 \\
Contours & 1 & 11 & Refit & 180 & 49.4 & 50.0 & $-$1.1 & 3.3 & 95 & 27/47 & 69 \\
Contours & 2 & 11 & Zero-shot & 180 & 50.0 & 48.3 & $-$4.4 & 4.4 & 79 & 28/41 & 45 \\
Contours & 2 & 11 & Refit & 180 & 48.3 & 46.1 & $-$3.3 & 3.3 & 80 & 30/42 & 42 \\
Strokes & 1 & 11 & Zero-shot & 180 & 52.8 & 52.8 & $-$3.9 & 0.6 & 60 & 20/30 & 41 \\
Strokes & 1 & 11 & Refit & 180 & 50.6 & 43.3 & $-$12.8 & 3.3 & 77 & 27/45 & 55 \\
Strokes & 2 & 11 & Zero-shot & 180 & 47.2 & 45.6 & 5.6 & 1.1 & 71 & 20/37 & 59 \\
Strokes & 2 & 11 & Refit & 180 & 50.0 & 53.9 & 4.4 & 1.1 & 65 & 19/29 & 49 \\
Contours & 1 & 10 + I & Zero-shot & 48 & 47.9 & 50.0 & $-$6.2 & $-$2.1 & 21 & 6/10 & 10 \\
Contours & 1 & 10 + I & Refit & 48 & 41.7 & 45.8 & $-$6.2 & 6.3 & 28 & 11/13 & 13 \\
Contours & 2 & 10 + I & Zero-shot & 48 & 47.9 & 56.2 & 2.1 & 2.1 & 14 & 3/5 & 17 \\
Contours & 2 & 10 + I & Refit & 48 & 45.8 & 54.2 & 2.1 & 2.1 & 22 & 7/9 & 15 \\
Strokes & 1 & 10 + I & Zero-shot & 48 & 50.0 & 50.0 & 0.0 & 2.1 & 18 & 7/9 & 7 \\
Strokes & 1 & 10 + I & Refit & 48 & 52.1 & 54.2 & 2.1 & $-$2.1 & 21 & 8/10 & 12 \\
Strokes & 2 & 10 + I & Zero-shot & 48 & 58.3 & 45.8 & $-$10.4 & 2.1 & 18 & 7/12 & 17 \\
Strokes & 2 & 10 + I & Refit & 48 & 47.9 & 58.3 & 6.2 & $-$6.2 & 15 & 2/5 & 10 \\
\bottomrule

\end{tabular}\end{adjustbox}}
\end{table}

\begin{table}[!htbp]
\centering\small
\caption{\textbf{Same evaluation restricted to pairs with both chains exact.} Each checkpoint regime uses its own fixed role probe. Dashes indicate undefined BA when the selected set lacks either label.}
\label{tab:refit-exact}
\fitsuppcolumn{\begin{adjustbox}{max width=\linewidth}\begin{tabular}{llrlrrrrrrrr}
\toprule
Renderer & Block & Ops & Head & $n$ & F & D & $\Delta$F & $\Delta$D & Flips & Rescue & Diff\\
\midrule
Contours & 1 & 11 & Zero-shot & 0 & -- & -- & -- & -- & 0 & 0/0 & 0 \\
Contours & 1 & 11 & Refit & 4 & 50.0 & 50.0 & $-$16.7 & 16.7 & 4 & 1/1 & 1 \\
Contours & 2 & 11 & Zero-shot & 0 & -- & -- & -- & -- & 0 & 0/0 & 0 \\
Contours & 2 & 11 & Refit & 15 & 26.1 & 30.7 & 12.5 & $-$13.6 & 7 & 1/3 & 2 \\
Strokes & 1 & 11 & Zero-shot & 0 & -- & -- & -- & -- & 0 & 0/0 & 0 \\
Strokes & 1 & 11 & Refit & 5 & 33.3 & 66.7 & $-$16.7 & $-$41.7 & 5 & 0/2 & 3 \\
Strokes & 2 & 11 & Zero-shot & 0 & -- & -- & -- & -- & 0 & 0/0 & 0 \\
Strokes & 2 & 11 & Refit & 23 & 41.1 & 51.1 & $-$5.6 & $-$2.8 & 7 & 2/3 & 4 \\
Contours & 1 & 10 + I & Zero-shot & 0 & -- & -- & -- & -- & 0 & 0/0 & 0 \\
Contours & 1 & 10 + I & Refit & 1 & -- & -- & -- & -- & 0 & 0/0 & 0 \\
Contours & 2 & 10 + I & Zero-shot & 0 & -- & -- & -- & -- & 0 & 0/0 & 0 \\
Contours & 2 & 10 + I & Refit & 4 & 33.3 & 33.3 & $-$33.3 & 66.7 & 2 & 1/1 & 2 \\
Strokes & 1 & 10 + I & Zero-shot & 0 & -- & -- & -- & -- & 0 & 0/0 & 0 \\
Strokes & 1 & 10 + I & Refit & 3 & 75.0 & 75.0 & 0.0 & 0.0 & 0 & 0/0 & 0 \\
Strokes & 2 & 10 + I & Zero-shot & 0 & -- & -- & -- & -- & 0 & 0/0 & 0 \\
Strokes & 2 & 10 + I & Refit & 7 & 30.0 & 50.0 & 0.0 & $-$20.0 & 2 & 0/0 & 2 \\
\bottomrule

\end{tabular}\end{adjustbox}}
\end{table}

\subsection{Endpoint substitutions and frozen repair on new families}
\label{sec:supp-endpoint}
The start-only, candidate-only and joint truth replacements operate at the symbol interface. The original mixed-carrier role probes supply unchanged operations. We retain the exact grammar and labels for every arm. The all-true arm verifies agreement with the manifest labels. For the learned head, endpoint classes index the mean standardized visual feature of that class and role in orientation-carrier training rows. This is a feature intervention with a class-conditioned proxy. The last-operation auxiliary state classifier supplies a separate diagnostic. We compare its predicted state with the true candidate and verify that replacing the later candidate feature leaves the earlier prediction invariant.

We screen two deployable symbolic repairs on the eight development cells: an orientation-only ridge fit and a template reader for the two endpoint tiles with the operation decoder unchanged. Their original BA means are 52.34\% and 99.31\%, respectively. The template reader uses five existing legend exemplars per state and only the queried crop. For learned Rotary, we compare the frozen and target-renderer-refitted heads under the same local reader and class-prototype input. Their original BA means are 47.14\% and 49.31\%. We lock the highest-scoring arm in each path before constructing the new holdout. No holdout score changes the frozen selection.

The prospective panels use eight new program/render-seed blocks per renderer and length, with 48 pairs in each block. Every block has 24 new canonical families and two endpoint variants per family. Original and donor queries share pixels, position, and traversal. Glyph renderers share the underlying programs and donor assignments. Two calibration dictionaries and checkpoints alternate across blocks. The intervals describe variation across the independent program blocks conditional on these two calibrations. The learned-head mechanism gate requires at least 80\% original BA in every block of a renderer-length panel. The symbolic executor is scored separately. In the repaired joint-exact subset, the rule path yields 100\% BA in both views by construction. The Rotary path still averages 47.80--49.41\% original BA across the four panels. Counts range from 356 to 383 of 384 pairs. Per-example predictions, flips and block intervals accompany the package.

Table~\ref{tab:endpoint-holdout} separates endpoint substitutions from complete-program execution on the same fresh pairs.

\begin{table}[!htbp]
\centering\small
\setlength{\tabcolsep}{3.2pt}
\caption{\textbf{Frozen endpoint repair on independent programs.} Each row has 384 pairs. F and D denote original and donor. Rule baseline uses the unmodified role probes and exact execution. Rule repair uses local endpoints with the same operation probe and executor. Gain is repair minus baseline in percentage points, with 95\% whole-block intervals. Joint counts pairs with both repaired chains exact. Rotary uses the frozen selection of the refit with endpoint feature prototypes. Gate gives the number of Rotary blocks with original BA at least 80\%.}
\label{tab:endpoint-holdout}
\fitsuppcolumn{\begin{adjustbox}{max width=\linewidth}\begin{tabular}{lrrrrrrrrrrr}
\toprule
& & & \multicolumn{2}{c}{Rule baseline} & \multicolumn{2}{c}{Rule repair} &
\multicolumn{2}{c}{Paired gain (pp)} & \multicolumn{2}{c}{Rotary} & \\
Renderer & Ops & Joint & F & D & F & D & F & D & F & D & Gate \\
\midrule
Strokes & 11 & 356 & 53.12 & 52.86 & 98.44 & 97.14 & 45.31 [40.62, 48.70] & 44.27 [40.62, 47.66] & 49.48 & 47.14 & 0/8 \\
Contours & 11 & 383 & 52.34 & 51.56 & 100.00 & 99.74 & 47.66 [44.79, 50.26] & 48.18 [45.05, 51.04] & 47.92 & 48.18 & 0/8 \\
Strokes & 10 + I & 370 & 54.69 & 58.07 & 98.44 & 99.22 & 43.75 [41.41, 46.61] & 41.15 [36.72, 45.31] & 48.70 & 48.44 & 0/8 \\
Contours & 10 + I & 382 & 54.69 & 53.91 & 100.00 & 99.74 & 45.31 [39.32, 51.04] & 45.83 [44.27, 47.40] & 49.22 & 49.48 & 0/8 \\
\bottomrule

\end{tabular}\end{adjustbox}}
\end{table}

\subsection{Operation-prefix interventions}
\label{sec:supp-operation}
The operation follow-up keeps the endpoint repair fixed. In each original and donor view, the local pixel reader supplies endpoint IDs and the fitted role probe supplies operation IDs. For a prefix of length $k$, we replace the first $k$ operation features with standardized class centroids from calibration training. The matched pair of arms indexes the same centroids by either true or decoded IDs. The feature intervention changes the inputs at the same positions in both arms. In the separate symbolic path we use the same ID substitutions with the exact G5c executor. The identity counts as a displayed operation in both analyses. We check that an additional edit leaves the preceding auxiliary state logits invariant up to floating-point tolerance.

A true-ID class mean replaces an individual symbol observation with a calibration prototype. Replacing individual features by calibration means changes feature variance, cross-node correlations and potentially the distribution expected by the frozen recurrence. The true-versus-decoded comparison controls the ID used to select the same means, while retaining this distribution shift. Prototype-fed heads therefore test the substituted interface rather than the original-input mechanism.

We record the five-state prediction after the start and after every operation. At the last operation, we compare this prediction with the locally read candidate to form a separate legality estimate. Rotary's native legality uses its candidate-position output and residual after the candidate update. The state-based decision therefore bypasses the candidate update and native classifier together. We fit calibration-only ridge readouts of the projected per-step output, its recurrent read component, the input residual, and the flattened post-update memory matrix. These probes test whether the correct state is accessible to the chosen readouts. Poor probe scores leave open whether a richer readout could recover the state.

Development compares the existing endpoint-only head, the same head given decoded operation prototypes, and a head refit on decoded prototypes for 64 additional AdamW updates on the 80 calibration training examples. The equal-cell original BA values are 49.31\%, 53.30\%, and 51.79\%, respectively. We select the second path and freeze the selection, training inputs, and weights. The operation counterfactuals on previously evaluated endpoint pairs are kept separate from the prospective test. On the latter, each of eight new blocks per length uses eight new canonical families and two endpoint variants per family. We evaluate 11 operations before the length-matched 10 operations plus identity. The two renderers share program and donor assignments. Block intervals are conditional on the two alternating calibration dictionaries and checkpoint sets. Phase swaps are eligible only when the selected learned path reaches 80\% original BA in all eight fresh blocks of its renderer-length panel.

Table~\ref{tab:operation-gate} gives the operation-prefix interventions for every renderer--length panel.

\begin{figure}[!htbp]
\centering\includegraphics[width=\linewidth]{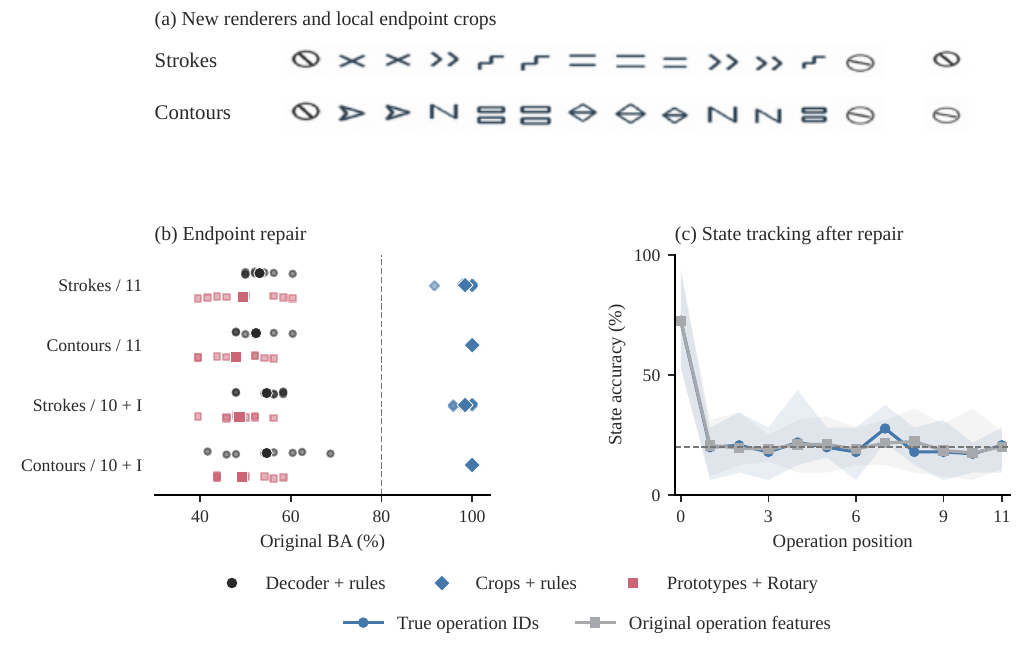}
\caption{\textbf{Endpoint reading and execution under renderer transfer.} (a) Actual new glyph renderings, with a start crop and candidate crop at right. (b) Original BA for each renderer--length cell (384 paired rows, eight blocks). Small points are blocks and large marks means; 80\% is the capability gate. (c) Auxiliary state accuracy using true-ID prototypes or no operation replacement. Each renderer--length cell has 128 paired programs; curves first average four cells within block. Both curves retain the same repaired endpoint interface. Shading shows the range across blocks; the dashed line is five-state chance (20\%).}
\label{fig:renderer-gate}
\end{figure}

Figure~\ref{fig:renderer-gate} compares the renderer capability gate and the state-tracking responses after endpoint and operation repair.

\begin{table}[!htbp]
\centering\small
\setlength{\tabcolsep}{3pt}
\caption{\textbf{Frozen operation repair on fresh pairs.} Each renderer-length panel uses 128 pairs across eight blocks. F and D denote original and donor BA. The F interval resamples whole program blocks. Gate counts blocks at or above 80\% in F.}
\label{tab:operation-gate}
\fitsuppcolumn{\begin{adjustbox}{max width=\linewidth}\begin{tabular}{lrrrr}
\toprule
Renderer & Ops & F [95\% interval] & D & Gate\\
\midrule
Strokes & 11 & 52.34 [44.53, 59.38] & 55.47 & 0/8\\
Contours & 11 & 50.78 [41.41, 59.38] & 50.78 & 0/8\\
Strokes & 10 + I & 52.34 [39.06, 67.19] & 52.34 & 1/8\\
Contours & 10 + I & 52.34 [47.66, 57.03] & 52.34 & 0/8\\
\bottomrule
\end{tabular}\end{adjustbox}}
\end{table}

Across the 512 fresh pairs, the frozen operation probe reads 5,627/5,632 operation IDs in original views and 5,628/5,632 in donor views. Full true-ID replacement changes the learned decision in one original and three donor views relative to matched decoded-ID prototypes. The exact executor with true operations and local endpoints reaches 100\% BA. The frozen learned arm scores 51.95\% original and 52.73\% donor BA, averaged across the four panels. The all-true prototype arm scores 52.15\% in both views. For the selected non-oracle arm, the class prototype sequences are identical across views in 503 of 512 pairs, including all 16 pairs in the single high-scoring strokes block. This bottleneck collapses most of the original/donor feature variation.

For the non-oracle selected arm, the original-view last-state accuracies of the calibration-only probes of the projected output, recurrent read component, residual, and matrix memory are 23.24\%, 24.41\%, 14.45\%, and 19.53\%. Their state-equality BA values are 48.63\%, 50.00\%, 44.73\%, and 52.34\%. These are short-training to long-test readout results. In the true-operation arm, directly comparing the auxiliary last state to the local candidate yields 50.98\% BA. The candidate-update and legality-readout stages can disagree with this comparison even when the auxiliary state is correct.

For the original-view all-true operation-prototype arm, auxiliary state accuracy is 72.66\% at the start and 19.92\% after the first operation. The matched decoded-ID arm has the same two means. With no operation replacement, first-operation accuracy is also low (20.70\%). These values average the four renderer--length cells within each of eight program blocks; the four cells share the same blocks. First-operation accuracy is similarly low with and without replacement.

\section{Additional Replays and Audits}
\subsection{Exact patch support and one-canvas controls}
\label{supp:patch}
DINOv2 receives the original $112\times448$ array without a canvas resize. With 14-pixel patches and bilinear sampling, the horizontal patch coordinate at $x=32+32t$ is $x/14-1/2$. The two adjacent columns and rows have positive interpolation weights. The exact initial pixel footprint is the corresponding $28\times28$ rectangle. Its vertical range is $[42,70)$ and its horizontal range is
$[14\lfloor x/14-1/2\rfloor,14(\lfloor x/14-1/2\rfloor+2))$.
A nominal tile can miss up to four columns of this rectangle. We reconstruct the archived renders and paste back every pixel in the rectangle after the original tile paste. This preserves the query's initial patch support and removes dependence on extrapolating the geometric result to encoder outputs.

\paragraph{RGB and donor audits.}
The full revised panel has 24,333 G5c and 24,301 G5r query footprints. The ordinary donor changes at least one RGB pixel in 11,326 (46.55\%) and 11,329 (46.62\%) footprints, respectively. There are seven G5c and five G5r self-donors in 1,920 source programs per grammar. We keep them in the audits. Donor legality agrees with source legality in 950/1,920 G5c and 974/1,920 G5r programs. The G5c composite seen at an individual query is legal for 2,506/12,159 queries with legal source and 2,495/12,174 with illegal source. The respective G5r counts are 2,719/12,148 and 2,502/12,153. These counts are clustered within programs. They distinguish the label of the visible composite from the source label used by the evaluation.

\begin{figure}[!htbp]
\centering\includegraphics[width=\linewidth]{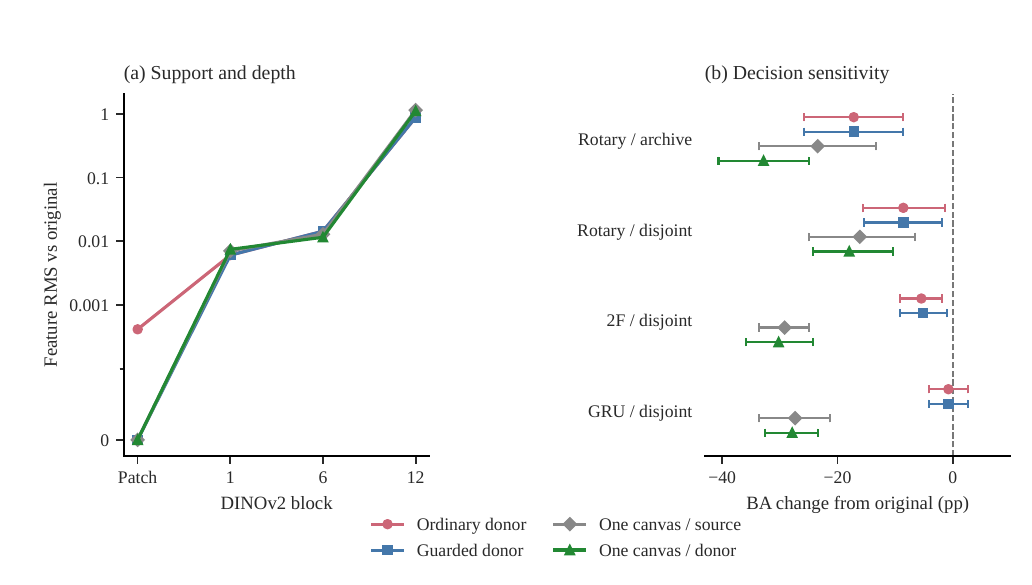}
\caption{\textbf{Guarded replay and the common-canvas control.} (a) Raw feature RMS relative to the original for archived G5c, pooled across valid queries. Patch embeddings are measured before attention and the final block output includes the model's output normalization. Layer scales can differ. (b) Paired changes from the original image with 95\% block-bootstrap intervals. Archive and disjoint results use 128 and 384 programs across eight blocks, respectively.}
\label{fig:guarded-support}
\end{figure}

\paragraph{Guarded replay.}
Figure~\ref{fig:guarded-support} compares initial support, later feature changes and paired task effects.
We draw eight legal and eight illegal programs per grammar and block with a fixed seed and keep the same draws across views. DINOv2 runs on MPS with the archived weights and frozen heads. All 1,536 clean and ordinary-donor decision anchors match the archive across both grammars and three heads. The maximum absolute probability difference is 0.02521. Guarded and common-canvas arms preserve initial sampled embeddings within $10^{-6}$, despite differences after the first attention block. In G5c, guarded BA is 71.88\% for Rotary, 95.31\% for 2F and 97.66\% for GRU. In G5r, it is 53.13\%, 97.66\% and 99.22\%. The eight-block bootstrap keeps all examples from one block together. These draws reuse archived programs.

\paragraph{One common canvas.}
We preserve the entire source program in place and add a distractor program in a parallel horizontal band outside the query footprints. The band is above or below with balanced assignment. It is filled with either the matched donor or a copy of the source. One forward pass provides all query features. With the G5c Rotary head the original scores 89.06\%, source duplication 65.63\% and donor addition 56.25\%. Donor addition versus source duplication is $-9.38$ pp ($[-17.97,-0.78]$). Both duplicate arms change image density and layout. The contrast compares two added rows at the same locations and is a different intervention from query-specific replacement. In the donor-added condition 96/128 pairs are jointly exact, with a $-32.67$ pp difference from original ($[-41.65,-22.65]$). The score drop under source duplication and changes for 2F and GRU indicate sensitivity to the two-row layout itself.

\paragraph{Disjoint program replay.}
We rerun the same arms on 384 previously generated 11-operation rows with 192 canonical program families. Each family has legal and illegal endpoint variants. Donor assignment deranges whole families and matches legality for this panel, unlike the archival assignment. Original and ordinary-donor replay reproduces all 2,304 stored binary predictions across the three heads, with maximum probability difference $1.34\times10^{-5}$. Rotary clean, ordinary donor, and guarded donor BA are 65.10\%, 56.51\%, and 56.51\%. Guarded and ordinary predictions differ on two of the 384 programs ($0.00$ pp BA difference, interval $[-0.78,0.78]$). Final feature RMS is 0.900 for each donor. The two-row donor and own-copy arms score 47.14\% and 48.96\%, a paired difference of $-1.82$ pp ($[-6.77,3.13]$). On the same two-row arms, 2F scores 60.16\% and 61.20\% from 90.36\% clean, and GRU scores 65.10\% and 65.63\% from 92.97\% clean. Both added rows give similar scores, so on these longer programs the drop is not specific to donor composition.

\subsection{Sensitivity of the renderer capability gate}
We prespecified 80\% original BA in every block to require a clear margin over binary chance and avoid phase interpretation for weak learned arms. This is a design criterion rather than a hypothesis-test threshold. On the frozen operation repair, lowering the gate to 60\% still leaves only 2/8, 3/8, 2/8 and 2/8 passing blocks across the four renderer-length panels. At 70, 80 and 90\% the corresponding counts are 0/8, 0/8, 1/8 and 0/8. Thus no renderer-length panel passes an all-block criterion at any of these thresholds. We also record start and candidate confusion matrices stratified by view, renderer and length.

\subsection{Validation-selected consistency follow-up}
\label{sec:transport-followup}
\begin{figure}[!htbp]\centering\includegraphics[width=\linewidth]{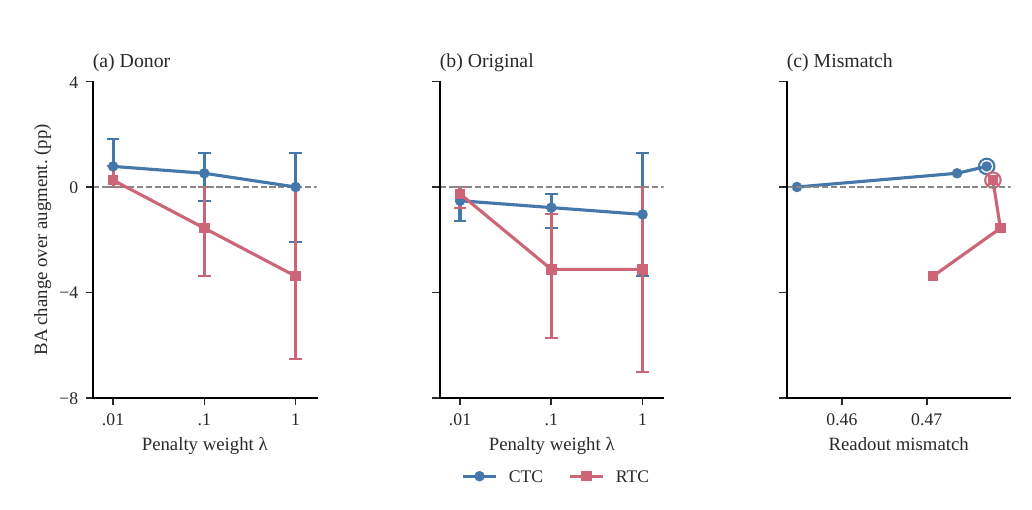}\caption{\textbf{Validation-locked weight sensitivity.} 384 paired programs in eight blocks. Panels (a,b) show donor and original BA changes relative to augmentation, with paired 95\% block-bootstrap intervals. Panel (c) relates mean donor BA change to readout mismatch; open rings mark the validation-selected $\lambda=0.01$. Lines connect the tested weights.}\label{fig:transport-sweep}\end{figure}

We test whether the local transport objective improves accuracy, and whether also constraining the query and residual paths helps. This follow-up uses a smaller data budget than the earlier 2,000-training-row, 480-validation-row, 256-update experiment (batch size 128). Eight frozen G5c Rotary heads are initialized from their archived checkpoints; each is adapted on 80 training and 80 validation rows using 64 updates, batch size 32, AdamW ($3\!\times\!10^{-4}$ initial learning rate, cosine decay, weight decay 0.01), and gradient clipping at 1. All arms share initialization and minibatch draws within a block. We compare clean continuation, ordinary two-view augmentation, probability consistency, binary-margin consistency, CTC, and readout transport consistency (RTC). The task loss is the legality cross-entropy plus 0.5 times the supervised state cross-entropy.
Let $\mathbf z^v$ be the binary logits, $m^v=z^v_1-z^v_0$, $r^v=\operatorname{vec}(S_T^v q_T^v)$, and $u^v=u_T^v$ for view $v\in\{F,D\}$. The added penalties are
\begin{align}
\mathcal L_{\mathrm{prob}}&=\mathbb E\|\operatorname{softmax}(\mathbf z^D)-\operatorname{softmax}(\mathbf z^F)\|_2^2,\\
\mathcal L_{\mathrm{margin}}&=\mathbb E(m^D-m^F)^2,\\
\mathcal L_{\mathrm{RTC}}&=\frac 12\mathbb E\left[\frac{\|r^D-r^F\|_2^2}{\operatorname{sg}(\|r^F\|_2^2)+\epsilon}+\frac{\|u^D-u^F\|_2^2}{\operatorname{sg}(\|u^F\|_2^2)+\epsilon}\right],
\end{align}
where $\epsilon=10^{-4}$ and $\operatorname{sg}$ stops the gradient. Gradients pass through both view trajectories in the numerators. Each penalty is added with weight $\lambda$ to the average task loss across the two views. RTC covers the actual pre-normalization readout and residual paths identified by the bound; it permits state-preserving transition/write cancellation. The comparison tests whether this broader objective improves accuracy.
We evaluate $\lambda\in\{0.01,0.1,1\}$ for every consistency objective. A checkpoint maximizes the mean of original/donor validation balanced accuracy, checked every 16 updates; one weight per objective maximizes mean validation performance across the eight blocks, with ties favoring the smaller weight. All four objectives select $\lambda=0.01$. Checkpoints and weights are locked before generating the test. The adaptation cache has no shared base keys or exact programs between training and validation; 12--21 broader reversal/renaming canonical families overlap per block.
The subsequent test contains 384 balanced rows from 192 globally distinct canonical families, excluded from all previously recorded panels. Each program has 10 nonidentity operations plus one identity. The same random derangement of donor families is evaluated with donor endpoint labels either preserved or reversed. These two conditions stratify a mixed-label donor rule under the per-query pixel-compositing protocol. Paired 95\% percentile intervals resample the eight blocks 40,000 times. Unselected weights are evaluated after selection, with checkpoints and selected weights fixed.
Table~\ref{tab:transport-followup} gives accuracy contrasts for the selected weights, with the remaining weight responses in Fig.~\ref{fig:transport-sweep}.

\begin{table}[!htbp]
\centering\small
\setlength{\tabcolsep}{4pt}
\caption{\textbf{Fresh-family consistency follow-up.} Accuracies are percentages and differences are percentage points; differences use 64-update augmentation as control. The CTC and RTC intervals touch zero. This panel reports the smaller-budget comparison.}
\label{tab:transport-followup}
\fitsuppcolumn{\begin{adjustbox}{max width=\linewidth}\begin{tabular}{lrrrrll}
\toprule
Method & $\lambda$ & Clean & Same donor & Opp. donor & $\Delta$ same [95\% CI] & $\Delta$ opp. [95\% CI] \\
\midrule
Frozen head & 0 & 78.12 & 69.01 & 69.79 & +1.82 [$-$4.69, +8.07] & +1.30 [$-$5.73, +8.07] \\
Clean continuation & 0 & 73.96 & 66.41 & 66.93 & $-$0.78 [$-$7.03, +4.69] & $-$1.56 [$-$7.55, +4.17] \\
Augmentation (64) & 0 & 72.14 & 67.19 & 68.49 & +0.00 [+0.00, +0.00] & +0.00 [+0.00, +0.00] \\
Augmentation (128) & 0 & 70.83 & 66.15 & 67.71 & $-$1.04 [$-$2.60, +0.26] & $-$0.78 [$-$3.39, +1.56] \\
Probability consistency & 0.01 & 72.14 & 67.19 & 68.49 & +0.00 [+0.00, +0.00] & +0.00 [+0.00, +0.00] \\
Margin consistency & 0.01 & 73.44 & 65.89 & 66.15 & $-$1.30 [$-$5.47, +2.34] & $-$2.34 [$-$6.77, +1.56] \\
CTC & 0.01 & 71.61 & 67.97 & 68.49 & +0.78 [+0.00, +1.82] & +0.00 [+0.00, +0.00] \\
Readout transport & 0.01 & 71.88 & 67.45 & 68.49 & +0.26 [+0.00, +0.78] & +0.00 [$-$0.78, +0.78] \\
\bottomrule
\end{tabular}\end{adjustbox}}

\end{table}
The fresh clean accuracy of the frozen head is 78.13\%, and every selected adaptation arm reduces it. CTC gains 0.78 pp on label-preserving donors (95\% CI $[0.00,1.82]$), and RTC gains 0.26 pp ($[0.00,0.78]$); exact two-sided sign-flip tests of the eight paired block effects give $p=0.50$ and $p=1.00$, respectively (all $2^8$ assignments, assuming sign-exchangeability under the null). On reversed-label donors, both mean differences are 0.00 pp; RTC has interval $[-0.78,0.78]$. Against margin consistency, RTC gives $+1.56$ pp $[-1.82,5.47]$. Neither objective gives a clear gain at the tested weights and budget.
\begin{table}[!htbp]
\centering\small
\caption{\textbf{Actual CPU adaptation costs and selected steps.} All candidates train to the fixed budget before validation selects a checkpoint.}
\label{tab:transport-cost}
\fitsuppcolumn{\begin{adjustbox}{max width=\linewidth}\begin{tabular}{lrrl}
\toprule
Selected arm & Updates & Mean seconds/block & Selected checkpoint step (blocks 1--8) \\
\midrule
Clean continuation & 64 & 4.07 & 16, 16, 32, 32, 32, 16, 16, 16 \\
Augmentation (64) & 64 & 3.92 & 48, 16, 32, 16, 64, 32, 64, 48 \\
Augmentation (128) & 128 & 9.46 & 16, 80, 48, 16, 16, 32, 48, 48 \\
Probability consistency & 64 & 4.37 & 48, 16, 32, 16, 64, 32, 64, 48 \\
Margin consistency & 64 & 3.87 & 16, 48, 32, 16, 16, 16, 32, 48 \\
CTC & 64 & 6.79 & 48, 16, 32, 16, 32, 32, 48, 48 \\
Readout transport & 64 & 8.30 & 48, 16, 32, 16, 48, 32, 48, 48 \\
\bottomrule
\end{tabular}\end{adjustbox}}

\end{table}
Table~\ref{tab:transport-cost} reports the measured training costs. RTC uses four recurrent trajectories per update versus two for ordinary augmentation. The 128-update augmentation control approximately matches the recurrent-trajectory count of 64-update RTC, and took 9.46 s/block versus 8.30 s/block for RTC. Task/state losses, readout computations and validation frequency differ, so FLOP counts can differ. All 120 adaptation fits took 701.1 s. Each 128-update control has eight checkpoint evaluations, while a 64-update fit has four. Against this control, CTC and RTC have donor differences of $+1.82$ pp $[0.00,3.65]$ and $+1.30$ pp $[0.00,2.60]$. Model selection, block predictions, discrepancies, costs and trained weights are saved.

\subsection{Full-budget consistency rerun}
\label{sec:full-budget-followup}
\begin{figure}[!htbp]
\centering\includegraphics[width=\linewidth]{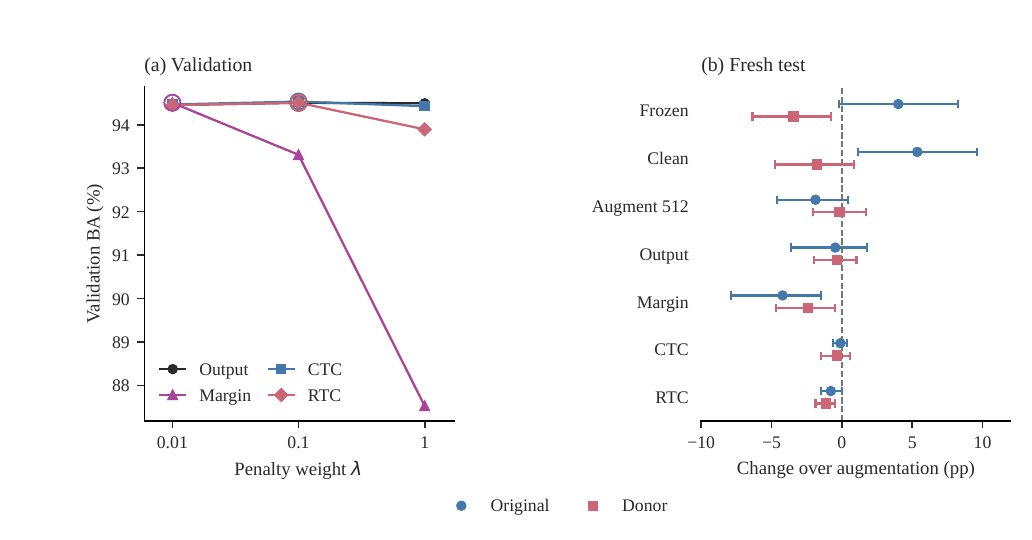}
\caption{\textbf{Locked weight choice and full-budget test effects.} (a) Mean paired-view validation BA over eight blocks; bold blue values mark the selected weight for each objective. (b) Original and donor BA changes relative to 256-update augmentation on the fresh 1,920-pair test (eight blocks), with paired 95\% block-bootstrap intervals. The 512-update augmentation arm is a trajectory-count control. No test score selects a weight.}
\label{fig:full-budget-transport}
\end{figure}

\begin{figure}[!htbp]
\centering\includegraphics[width=\linewidth]{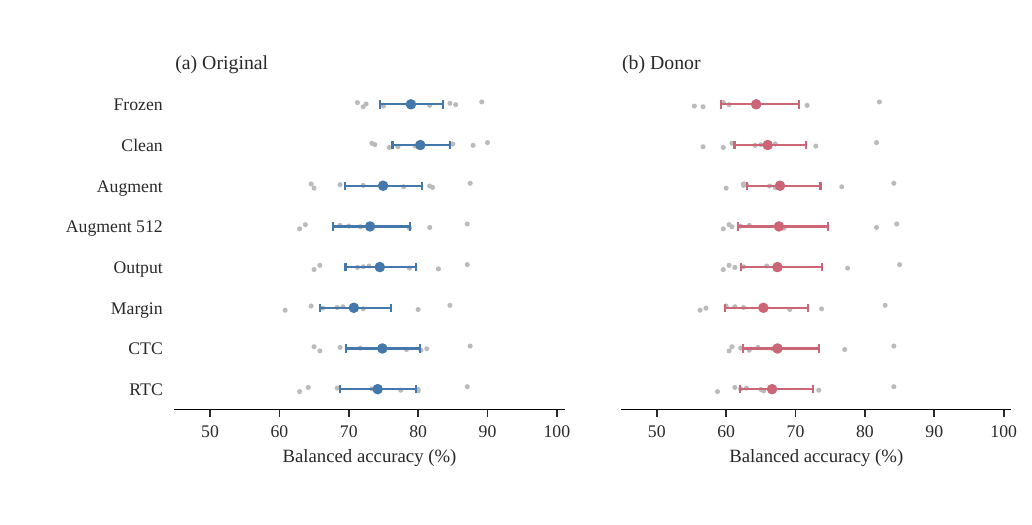}
\caption{\textbf{Absolute capability and donor robustness for every selected adaptation arm.} The existing fresh test contains 1,920 paired rows in eight blocks. Gray dots are blocks; colored marks are means and bars are 95\% block-bootstrap intervals. Weights use validation only. The 512-update augmentation arm is a compute control. Paired differences are shown in Fig.~\ref{fig:full-budget-transport}.}
\label{fig:absolute-adaptation}
\end{figure}

We repeat the weight comparison at the earlier adaptation budget: 2,000 training and 480 validation rows per block, 256 updates with batch size 128, and eight frozen G5c Rotary initializations. All methods use one freshly encoded data pool; the source selection, optimizer, minibatch seeds, cosine learning-rate schedule, and 32-update validation cadence match the archived recipe. We train probability, margin, CTC, and RTC penalties at $\lambda\in\{0.01,0.1,1\}$, alongside clean continuation, ordinary augmentation, and 512-update augmentation. Checkpoints maximize the mean of original/donor validation balanced accuracy; each objective selects one weight across all eight blocks. Because the earlier adaptation weights and full feature caches were unavailable, every comparison arm is retrained on the common new pool.
The final test is generated only after these choices and the capability-matched 2F selection are locked. It contains 1,920 balanced rows from 960 canonical 10-operation families, disjoint from all recorded prior panels and the separate 256-row capability-validation panel. All images have 13 nodes. The primary donor assignment is exactly balanced between preserving and reversing the source legality label at the family level; both strata are additionally reported on the same 1,920 source rows. This protocol defines a full-budget rerun under one CUDA environment with newly trained comparison arms.
Table~\ref{tab:full-transport} and Fig.~\ref{fig:full-budget-transport} report the locked-weight test; Fig.~\ref{fig:absolute-adaptation} shows absolute capability.

\begin{table}[!htbp]
\centering\small\setlength{\tabcolsep}{4pt}
\caption{\textbf{Full-budget, validation-selected rerun on fresh canonical families.} Accuracies are percentages; differences are percentage points relative to 256-update augmentation. Paired 95\% intervals resample the eight blocks 40,000 times.}
\label{tab:full-transport}
\fitsuppcolumn{\begin{adjustbox}{max width=\linewidth}\begin{tabular}{lrrrrrl}
\toprule
Arm & $\lambda$ & Clean & Donor & Same & Opp. & $\Delta$ donor [95\% CI] \\
\midrule
Frozen & 0 & 78.96 & 64.32 & 64.17 & 64.53 & $-$3.44 [$-$6.35, $-$0.78] \\
Clean & 0 & 80.31 & 65.99 & 65.42 & 65.89 & $-$1.77 [$-$4.74, +0.89] \\
Augmentation & 0 & 74.95 & 67.76 & 68.02 & 67.55 & +0.00 [+0.00, +0.00] \\
Augmentation ($2\times$) & 0 & 73.07 & 67.60 & 67.66 & 66.98 & $-$0.16 [$-$2.08, +1.72] \\
Probability & 0.1 & 74.48 & 67.40 & 67.50 & 67.55 & $-$0.36 [$-$1.98, +1.04] \\
Margin & 0.01 & 70.73 & 65.36 & 65.47 & 64.95 & $-$2.40 [$-$4.69, $-$0.47] \\
CTC & 0.1 & 74.84 & 67.40 & 67.86 & 67.40 & $-$0.36 [$-$1.51, +0.57] \\
RTC & 0.1 & 74.17 & 66.61 & 66.77 & 66.09 & $-$1.15 [$-$1.88, $-$0.52] \\
\bottomrule
\end{tabular}\end{adjustbox}}

\end{table}
CTC has a donor difference of $-$0.36 [$-$1.51, +0.57] pp and a clean difference of $-$0.10 [$-$0.62, +0.36] pp. Its exact two-sided sign-flip $p$-value is 0.7188 (Holm-adjusted 0.7188 for the two CTC/RTC comparisons). Against the 512-update augmentation control, its donor difference is $-$0.21 [$-$2.34, +1.98] pp.
RTC has a donor difference of $-$1.15 [$-$1.88, $-$0.52] pp and a clean difference of $-$0.78 [$-$1.46, +0.00] pp. Its exact two-sided sign-flip $p$-value is 0.0312 (Holm-adjusted 0.0625 for the two CTC/RTC comparisons). Against the 512-update augmentation control, its donor difference is $-$0.99 [$-$3.44, +1.15] pp.
Selected CTC reduces mean normalized local transport discrepancy from 0.13190 to 0.12488 (5.32\% reduction), without an accuracy gain over augmentation. At the unselected weight $\lambda=1$, discrepancy falls to 0.10825 (17.93\% reduction), while donor accuracy changes by $-$1.46 [$-$2.34, $-$0.68] pp. In this setting, lower transport discrepancy accompanies similar or lower accuracy.
Table~\ref{tab:full-transport-cost} gives the measured costs and selected training steps.

\begin{table}[!htbp]
\centering\small
\caption{\textbf{Actual adaptation costs.} The 512-update augmentation control approximately matches the RTC recurrent-trajectory count; validation checks and loss computations differ.}
\label{tab:full-transport-cost}
\fitsuppcolumn{\begin{adjustbox}{max width=\linewidth}\begin{tabular}{lrrl}
\toprule
Arm & Updates & Mean GPU seconds/block & Selected step (blocks 1--8) \\
\midrule
Clean & 256 & 22.93 & 96, 96, 64, 64, 64, 224, 96, 160 \\
Augmentation & 256 & 22.43 & 160, 192, 64, 160, 64, 128, 256, 192 \\
Augmentation ($2\times$) & 512 & 44.58 & 64, 416, 64, 288, 256, 448, 416, 256 \\
Probability & 256 & 23.01 & 160, 256, 192, 160, 160, 128, 256, 192 \\
Margin & 256 & 23.04 & 192, 192, 96, 160, 224, 64, 96, 160 \\
CTC & 256 & 26.12 & 160, 256, 160, 160, 64, 128, 160, 160 \\
RTC & 256 & 31.33 & 160, 192, 64, 128, 64, 64, 192, 192 \\
\bottomrule
\end{tabular}\end{adjustbox}}

\end{table}

\subsection{Capability-matched 2F control}
We trained a fresh two-factor head in each block on all 16,000 original training examples for 5,040 updates (80 epochs, batch 256). At every update, an independent clean length-10 validation panel of 32 examples selected the checkpoint closest in balanced accuracy to the frozen Rotary head; ties selected the earlier update. Neither donor features nor the final test entered selection. All eight choices were locked before generating the 1,920-example final test. A smaller 2,000-example/512-update pilot did not match capability; the full-budget version was fixed before the final test was accessed.
The mean validation absolute gap is 0.000 pp, while the mean test absolute gap is 13.12 pp. Clean test BA is 65.83\% for the selected 2F and 78.96\% for Rotary. The control failed to match capability on the independent test.
Table~\ref{tab:capmatch-full} evaluates the clean-capability matching control and common-context responses.

\begin{table}[!htbp]\centering\small
\caption{\textbf{Capability matching and common-context evaluation.} BA and flips are percentages. The primary donor uses a fixed 50/50 mixture of same-label and opposite-label donor families. Full per-block intervals and recovery denominators accompany the released results.}\label{tab:capmatch-full}
\fitsuppcolumn{\begin{adjustbox}{max width=\linewidth}\begin{tabular}{lrrr}\toprule Head & Clean BA & Donor BA & Flips \\\midrule
Val.-matched 2F & 65.83 & 63.44 & 27.19 \\
Final-update 2F & 93.54 & 88.80 & 13.80 \\
Archived 2F & 93.75 & 89.22 & 13.91 \\
Archived GRU & 94.69 & 94.17 & 7.92 \\
Frozen Rotary & 78.96 & 64.32 & 31.93 \\
\bottomrule\end{tabular}\end{adjustbox}}\end{table}
The matched-2F minus Rotary donor contrast is $-$0.89 [$-$7.92, 6.09] pp (paired block bootstrap). The difference between their clean--donor drops is $-$12.24 [$-$17.71, $-$5.62] pp. These estimates compare the fitted heads at their observed capabilities. The shared supervised decoder recovers both complete chains on 1,409/1,920 pairs; conditioning uses the same witness for every head.
Training uses the shared saved Linux/Pillow13 canvases; capability validation and final evaluation retain the prespecified common Pillow12.3 features. A separate balanced historical-IID audit fixes 16 examples per class and block before comparing the two renderers. Models and the evaluation environment remain fixed for this audit.
Across these 256 audit examples, the matched 2F changes 2 predictions and Rotary changes 0; mean pixel RMS is 0.4115 on the 0--255 scale. Producer metadata records cross-platform pixel-anchor differences. The 32-example capability panels have 3.125-pp resolution.

\subsection{Full-budget second-encoder comparison}
\label{sec:supp-second-encoder}
We use frozen DINO ViT-S/16~\cite{caron2021emerging} with 384-dimensional query features and the same 16,000 training rows, 6,000 validation rows, 5,040 updates, and checkpoint rule for Rotary, 2F, and GRU in eight blocks. Both encoders read identical saved Linux/Pillow13 RGB on the channel panel. DINOv2 replay produces the same original and donor decisions as the archived run in all 5,760 head--pair comparisons. Four of 48 source-Mac/Linux RGB anchors differ at a small number of rasterized pixels. The comparison shares test RGB, programs and training budget, while the pretrained encoder and archived training environment differ. DINOv1's initial bilinear patch support lies inside each preserved tile. The patch embeddings agree exactly at all valid queries. Relative feature displacement is nonzero at later stages.

\begin{figure}[!htbp]
\centering
\includegraphics[width=\linewidth]{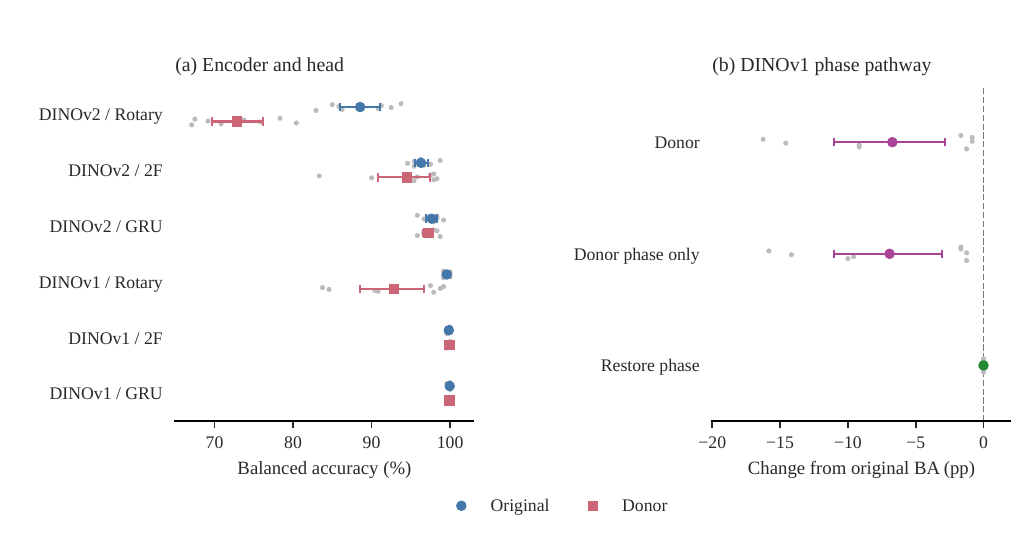}
\caption{\textbf{Execution sensitivity across frozen DINOv1 and DINOv2 encoders.} (a) Absolute original and donor BA for three learned heads under each encoder, using the full original training budget on 1,920 channel-panel pairs in eight blocks. (b) DINOv1 Rotary channel interventions; points and intervals are the mean and block-bootstrap 95\% CI, and gray dots show eight blocks. Restoring the original phase fixes all 134 new donor errors and exactly reproduces all original decisions. The restoration uses unavailable original information and is an oracle mechanism test.}
\label{fig:second-encoder}
\end{figure}

\begin{figure}[!htbp]
\centering
\includegraphics[width=\linewidth]{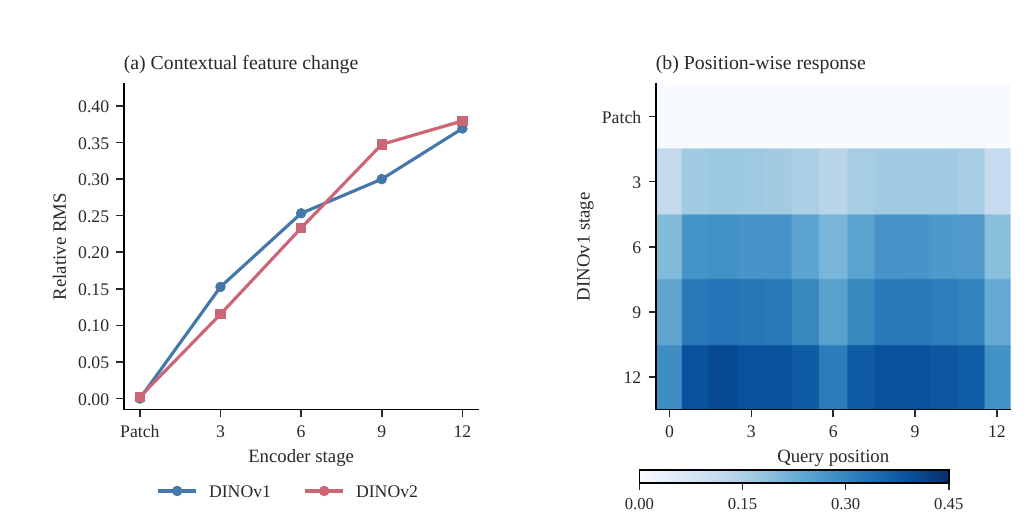}
\caption{\textbf{Initial support and later contextual changes.} 1,920 channel-panel pairs in eight blocks. Relative RMS divides the RMS feature change by the original feature RMS, pooling valid queries and feature dimensions within each block. (a) Eight-block means and 95\% block-bootstrap intervals at the patch embedding, Transformer blocks 3/6/9, and normalized final block 12. (b) DINOv1 per-position responses, averaging block-specific ratios over valid positions. Initial sampled DINOv1 patch features are exactly preserved, while later features change. The comparison jointly varies patch size, pretraining and positional representation.}
\label{fig:second-encoder-layers}
\end{figure}

On the same fresh 1,920 rows as the consistency and capability tests, all DINOv1 chains are exactly decoded in both views. Rotary changes from 99.01\% to 91.09\% BA ($-7.92$ pp, $[-12.14,-3.70]$). 2F changes from 99.79\% to 99.48\% and GRU from 99.79\% to 99.95\%. No test sample selects a head or decoder. Same-label and opposite-label donors use these same source rows. The separate phase replay on the channel panel was specified before computing its outputs, with the earlier DINOv2 results already known. Its channel interventions use the original phase and both fitted runs. All 24 reported DINOv1 fits use the same deterministic and TF32 settings. Table~\ref{tab:second-encoder} and Fig.~\ref{fig:second-encoder} compare execution, while Fig.~\ref{fig:second-encoder-layers} separates initial support from later contextual changes.

\begin{table}[!htbp]
\centering
\small
\setlength{\tabcolsep}{5pt}
\caption{\textbf{Full-budget second-encoder replication on the 1,920-pair G5c channel panel.} Both encoders consume the same saved test RGB and donor assignments; DINOv2 uses archived checkpoints, whereas DINOv1 uses the same training/validation programs, head recipe, and 5,040-update budget. Confidence intervals resample eight paired training/program blocks. All 24 DINOv1 block/head combinations exceed 90\% clean-long BA. Exact recovery is an information witness using symbol supervision and exact rules.}
\label{tab:second-encoder}
\fitsuppcolumn{\begin{adjustbox}{max width=\linewidth}\begin{tabular}{llrrrrr}
\toprule
Encoder & Head & Original BA & Donor BA & $\Delta$ BA [95\% CI], pp & Flip (\%) & Both exact (\%) \\
\midrule
\rowcolor{resultorange}DINOv2 & Rotary & 88.54 & 72.86 & $-$15.68 [$-$16.87, $-$14.01] & 25.05 & 91.77 \\
DINOv2 & 2F & 96.30 & 94.53 & $-$1.77 [$-$5.10, +0.68] & 7.19 & 91.77 \\
DINOv2 & GRU & 97.66 & 97.24 & $-$0.42 [$-$1.35, +0.52] & 3.85 & 91.77 \\
\rowcolor{resultorange}DINOv1 & Rotary & 99.58 & 92.86 & $-$6.72 [$-$11.04, $-$2.86] & 7.24 & 100.00 \\
DINOv1 & 2F & 99.84 & 99.95 & +0.10 [+0.00, +0.26] & 0.10 & 100.00 \\
DINOv1 & GRU & 99.95 & 99.95 & +0.00 [$-$0.16, +0.16] & 0.10 & 100.00 \\
\bottomrule
\end{tabular}\end{adjustbox}}

\end{table}

\subsection{Phase substitutions by program role}
\label{sec:supp-role-phase}
In each panel and encoder we evaluate eight fixed blocks of 240 G5c programs. We exchange phase between the two views at operation nodes, the true last candidate, or both. Operation nodes include identities, and the candidate index uses each program's length. Decay, write, query and residual use the base view. The start phase acts on zero initial memory, and exchanging it leaves the read and logits exactly unchanged. The factual replay agrees with the archived forward within $1.47\times10^{-5}$ in probability, and the expanded final read agrees within $2.15\times10^{-6}$. Neither check changes a factual decision.

\begin{table}[!htbp]\centering\small
\caption{\textbf{Operation and candidate phase in both replay directions.} Each panel and encoder has 1,920 pairs in eight blocks. F and D denote original and donor. Injection (restoration) uses F (D) as the base. BA change and paired 95\% intervals are relative to that base. Fixed counts repairs of original-correct, donor-wrong errors under restoration. Corrected and introduced count all base errors corrected and new errors introduced. Disagree counts predictions different from F. Held-out denotes the reanalysis of the previously evaluated panel.}
\label{tab:role-phase-supp}
\setlength{\tabcolsep}{3pt}
\fitsuppcolumn{\begin{adjustbox}{max width=\linewidth}\begin{tabular}{lllrrrrrr}
\toprule
Panel / encoder & Base & Phase swapped & BA (\%) & $\Delta$ BA [95\% CI], pp & Fixed & Corrected & Introduced & Disagree \\
\midrule
Channel / DINOv1 & F & Base & 99.58 & --- & --- & --- & --- & 0 \\
 &  & Operations & 93.80 & $-5.78$ [$-9.22,-2.29$] & --- & 5 & 116 & 121 \\
 &  & Candidate & 99.38 & $-0.21$ [$-0.57,+0.05$] & --- & 2 & 6 & 8 \\
 &  & Both & 92.66 & $-6.93$ [$-11.04,-3.07$] & --- & 5 & 138 & 143 \\
 & D & Base & 92.86 & --- & --- & --- & --- & 139 \\
 &  & Operations & 99.48 & $+6.61$ [$+2.81,+10.83$] & 132/134 & 132 & 5 & 6 \\
 &  & Candidate & 93.85 & $+0.99$ [$+0.16,+2.14$] & 26/134 & 26 & 7 & 120 \\
 &  & Both & 99.58 & $+6.72$ [$+2.86,+11.04$] & 134/134 & 134 & 5 & 0 \\
\addlinespace[2pt]
Channel / DINOv2 & F & Base & 88.54 & --- & --- & --- & --- & 0 \\
 &  & Operations & 74.27 & $-14.27$ [$-16.25,-12.03$] & --- & 74 & 348 & 422 \\
 &  & Candidate & 87.76 & $-0.78$ [$-1.61,+0.00$] & --- & 33 & 48 & 81 \\
 &  & Both & 72.19 & $-16.35$ [$-19.01,-13.65$] & --- & 78 & 392 & 470 \\
 & D & Base & 72.86 & --- & --- & --- & --- & 481 \\
 &  & Operations & 88.07 & $+15.21$ [$+13.28,+16.41$] & 343/391 & 373 & 81 & 159 \\
 &  & Candidate & 74.64 & $+1.77$ [$+0.68,+2.97$] & 61/391 & 70 & 36 & 443 \\
 &  & Both & 88.54 & $+15.68$ [$+13.70,+17.08$] & 351/391 & 386 & 85 & 142 \\
\addlinespace[2pt]
Held-out / DINOv1 & F & Base & 99.01 & --- & --- & --- & --- & 0 \\
 &  & Operations & 92.40 & $-6.61$ [$-10.26,-3.07$] & --- & 5 & 132 & 137 \\
 &  & Candidate & 98.65 & $-0.36$ [$-0.52,-0.21$] & --- & 2 & 9 & 11 \\
 &  & Both & 90.31 & $-8.70$ [$-13.02,-4.32$] & --- & 5 & 172 & 177 \\
 & D & Base & 91.09 & --- & --- & --- & --- & 164 \\
 &  & Operations & 98.59 & $+7.50$ [$+3.33,+11.67$] & 153/158 & 154 & 10 & 14 \\
 &  & Candidate & 92.60 & $+1.51$ [$+0.52,+2.60$] & 38/158 & 39 & 10 & 135 \\
 &  & Both & 98.80 & $+7.71$ [$+3.49,+12.03$] & 157/158 & 158 & 10 & 10 \\
\addlinespace[2pt]
Held-out / DINOv2 & F & Base & 78.96 & --- & --- & --- & --- & 0 \\
 &  & Operations & 65.94 & $-13.02$ [$-15.21,-11.04$] & --- & 138 & 388 & 526 \\
 &  & Candidate & 79.48 & $+0.52$ [$-0.52,+1.35$] & --- & 75 & 65 & 140 \\
 &  & Both & 64.95 & $-14.01$ [$-16.93,-11.35$] & --- & 143 & 412 & 555 \\
 & D & Base & 64.32 & --- & --- & --- & --- & 613 \\
 &  & Operations & 78.23 & $+13.91$ [$+10.83,+17.24$] & 351/447 & 401 & 134 & 292 \\
 &  & Candidate & 65.36 & $+1.04$ [$-0.31,+2.29$] & 56/447 & 70 & 50 & 593 \\
 &  & Both & 76.82 & $+12.50$ [$+8.85,+16.35$] & 350/447 & 403 & 163 & 291 \\
\addlinespace[2pt]
\bottomrule
\end{tabular}\end{adjustbox}}
\end{table}

\begin{figure}[!htbp]
\centering\includegraphics[width=\linewidth]{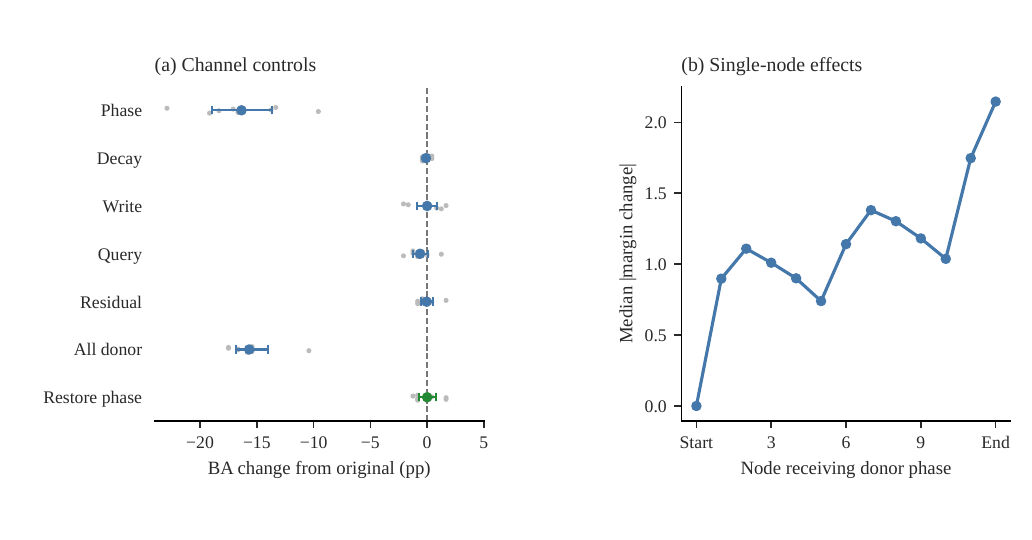}
\caption{\textbf{Channel and single-node checks.} (a) Donor-channel substitutions and original-phase restoration on the G5c Rotary channel panel (1,920 pairs, eight blocks), relative to original BA. Gray points are blocks, colored marks means and bars paired 95\% whole-block bootstrap intervals. (b) Each node separately receives donor phase in a 48-pair fixture from one block; the ordinate is median absolute final pre-softmax margin change. Full correctness transitions are displayed in the main mechanism figure; these panels add the single-node control.}
\label{fig:historical-channels}
\end{figure}

Table~\ref{tab:role-phase-supp} gives both substitution directions on the channel panel and on previously evaluated held-out programs. We use 40,000 paired whole-block bootstrap draws and report unadjusted intervals. On the previously evaluated DINOv2 panel, restoring both roles achieves 76.82\% BA and fixes 350 of 447 new errors, compared with 78.23\% and 351 repairs for operation-only restoration. The added candidate substitution can introduce errors as well as correct them. A role intervention over all operation nodes measures their joint effect, while the individual-node fixture in Fig.~\ref{fig:historical-channels} measures absolute margin displacement at one node.

For the candidate, we also exchange the query $q_n$, write $B_n$ or residual $u_n$ using a fixed base prefix. The residual substitution changes both its addition to the read and its separate classifier input. Each has a smaller average recovery effect than operation-phase restoration. The phase--write interaction at the candidate is zero at the raw read to numerical precision ($3.81\times10^{-6}$ maximum absolute difference in the four-cell contrast), as expected from the sum of the two terms in the read. Interactions can appear after normalization and classification. The fixed prefix can already contain information about the candidate through global encoding. The archived arm and factorial tables report these interventions for both directions and for the jointly exact subsets.

\subsection{Consistency objectives and margins on adaptation splits}
We reevaluate saved checkpoints on every adaptation training and validation row, with both original and donor views. These short programs contain three to seven nonidentity operations. The eight blocks total 16,000 adaptation training and 3,840 validation rows. The validation rows previously selected checkpoints and weights. We keep the locked weights and use 40,000 whole-block bootstrap draws for intervals. Table~\ref{tab:objective-margin-supp} reports donor cross-entropies and the paired objective computed over both views. The CTC mean weights valid nodes and the RTC mean weights rows.

\begin{table}[!htbp]\centering\small
\caption{\textbf{Complete short-program adaptation splits at the locked weights.} Donor BA and paired 95\% intervals relative to 256-update augmentation. CE and penalty values are means over eight blocks. The augmentation row gives both unweighted penalties in CTC / RTC order. Error changes use the fixed augmentation model's donor true-label margin $m$. Cells report corrections / introduced errors in the same rows. The $\lvert m\rvert<1$ group has 73 training and 103 validation rows.}
\label{tab:objective-margin-supp}
\setlength{\tabcolsep}{4.4pt}
\fitsuppcolumn{\begin{adjustbox}{max width=\linewidth}\begin{tabular}{llrrrrrrr}
\toprule
Split & Arm & BA (\%) & $\Delta$ BA [95\% CI], pp & Legality CE & State CE & Penalty & $\lvert m\rvert<1$ & $\lvert m\rvert\geq1$ \\
\midrule
Training & Augmentation & 99.538 & --- & 0.02046 & 0.04203 & 0.13969 / 0.26869 & --- & --- \\
 & CTC .1 & 99.713 & $+0.175$ [$-0.056,+0.581$] & 0.01348 & 0.03666 & 0.13179 & 10 / 1 & 22 / 3 \\
 & RTC .1 & 99.213 & $-0.325$ [$-0.875,-0.025$] & 0.02924 & 0.04638 & 0.26142 & 1 / 11 & 1 / 43 \\
\addlinespace[2pt]
Validation & Augmentation & 92.865 & --- & 0.37349 & 0.10814 & 0.13827 / 0.27788 & --- & --- \\
 & CTC .1 & 92.943 & $+0.078$ [$-0.130,+0.286$] & 0.36226 & 0.10257 & 0.13072 & 10 / 13 & 7 / 1 \\
 & RTC .1 & 92.995 & $+0.130$ [$-0.208,+0.469$] & 0.36965 & 0.10937 & 0.27215 & 11 / 11 & 9 / 4 \\
\bottomrule
\end{tabular}\end{adjustbox}}
\end{table}

CTC reduces its penalty by 5.66\% on training and 5.46\% on validation. In training, augmentation makes 74 donor errors. CTC corrects 32 and introduces four, with one block contributing 31 of the net 28 additional correct predictions. On validation, CTC corrects 17 and introduces 14. RTC lowers its penalty by 2.71\% and 2.06\% on training and validation, respectively. RTC training BA falls by $0.325$ pp ($[-0.875,-0.025]$). Of the 274 augmentation validation errors, 232 have absolute true-label margins of at least one. In the 103 validation rows with smaller absolute margins, CTC introduces 13 errors and corrects 10. This split-level audit distinguishes objective and cross-entropy gains from changes in classification. The recorded checkpoints were selected separately, so their comparison combines the effects of the penalty, optimization and checkpoint selection.

\section{Confirmation on New Program Families}
\label{sec:review-independent}
We generated this panel after the diagnostic results reported elsewhere in this document were known. It uses new canonical program families, while retaining the eight existing full-budget DINOv1 Rotary heads and their fixed role decoders. It is separate from the 1,920-pair channel panel and from the 11-operation and $10+1$ panels of Sec.~\ref{sec:supp-independent}. The protocol, model files, code, source inventory, exclusion union and eight input manifests were hashed and locked before encoding or scoring any new program. There is no new training, checkpoint selection or threshold tuning.

\paragraph{New families and the fixed donor law.}
Each of eight blocks contains 24 new canonical families, each with a legal and an illegal endpoint: 384 program rows from 192 globally distinct families. Every program has ten nonidentity operations and exactly one identity, hence 13 nodes. Canonicalization removes identities, normalizes operation names by first occurrence, and takes the minimum of the forward and reversed normalized sequences. Cyclic rotations retain distinct canonical keys. We exclude displayed and source/negative families from all earlier training, validation and test row containers, the eight full original G5c manifests and the earlier canonical exclusions. An independent pre-result audit found no overlap with the locked exclusion union and no duplicate new family across blocks. Length eight had no available family under these exclusions, and length nine had only 174, fewer than the 192 required; length ten was chosen on this availability check, without model outcomes.

Both carriers are balanced within block. Donors follow the original row-level cyclic-shuffle law within split, node count, carrier and vocabulary. No row is its own donor. Thirteen of the 384 donor assignments use the opposite-endpoint sibling from the same canonical family; we kept them as assigned (Sec.~\ref{sec:sibling-sensitivity}). The donor law operates at row level and permits these within-family assignments. At each query we preserve the source's full-height 32-pixel tile and position in the donor canvas, encode separately, then assemble features in the supplied order. No new row is selected based on recognition or model accuracy.

\paragraph{Anchors and capability.}
Before new encoding, 48 fixed source RGB anchors match byte-for-byte and 16 archived numerical anchors have no classification disagreement; their maximum probability discrepancy is $1.99\times10^{-9}$, below the locked $10^{-5}$ tolerance. MPS was used without invoking the predefined CPU fallback. All 4,992 new query-pixel checks pass, and the maximum sampled initial patch-embedding difference is zero. The factual phase replay satisfies its locked $2\times10^{-5}$ probability tolerance and preserves factual decisions. Original BA ranges from 95.83\% to 100\%, so all eight blocks exceed the 80\% interpretation gate.

\paragraph{Paired execution and restoration.}
All 384 complete chains decode exactly in both views, so the jointly exact analysis coincides with the all-row analysis. Original and donor BA are 99.22\% and 92.19\%, a paired change of $-7.03$ pp (95\% block-bootstrap interval $[-11.46,-2.60]$). There are 29 new donor errors, two donor corrections and 31 original/donor disagreements. Donor-phase injection lowers original BA to 92.97\%. Restoring original phase raises donor BA to 98.70\%, correcting 28 of the 29 new donor errors, while introducing three errors relative to donor predictions. Two of 384 predictions still differ from the original (99.48\% agreement). Thus this new-program result supports a phase-sensitive execution pathway in the capable frozen G5c model, while retaining the distinction between repairing new errors and recovering all original predictions.

Table~\ref{tab:review-independent} reports both phase-exchange directions; Table~\ref{tab:final-recovery-cost} compares repair benefits with introduced errors and residual disagreement.

\begin{table}[!htbp]
\centering\small
\caption{\textbf{Bidirectional phase interventions on all 384 new-program pairs.} Both complete chains are exact on every pair. Base identifies the recipient and the reference for BA change, corrected errors (Corr.) and introduced errors (New). Fixed / 29 counts repaired errors that were newly induced by the donor. Diff.\ F counts predictions different from original F. The start phase acts on zero initial memory; ``all'' exchanges operation and candidate phases. Intervals use 40,000 percentile resamples of the eight equally weighted paired training/program blocks (seed 2026111011).}
\label{tab:review-independent}
\begin{adjustbox}{max width=\linewidth}\begin{tabular}{llrrrrrr}
\toprule
Base & Phase source / site & BA (\%) & $\Delta$ [95\% CI], pp & Corr. & New & Fixed / 29 & Diff. F \\
\midrule
F & Original reference & 99.22 & -- & -- & -- & -- & 0 \\
F & Donor / all & 92.97 & $-6.25$ [$-9.90,-2.60$] & 2 & 26 & -- & 28 \\
F & Donor / operations & 93.75 & $-5.47$ [$-9.11,-1.82$] & 2 & 23 & -- & 25 \\
F & Donor / candidate & 98.96 & $-0.26$ [$-0.78,0.00$] & 0 & 1 & -- & 1 \\
\midrule
D & Donor reference & 92.19 & -- & -- & -- & 0/29 & 31 \\
\rowcolor{resultblue}D & Original / all & \textbf{98.70} & $+6.51$ [$+2.60,+10.42$] & 28 & 3 & \textbf{28/29} & \textbf{2} \\
D & Original / operations & 98.70 & $+6.51$ [$+2.60,+10.42$] & 28 & 3 & \textbf{28/29} & \textbf{2} \\
D & Original / candidate & 92.45 & $+0.26$ [$0.00,+0.78$] & 1 & 0 & 1/29 & 30 \\
\bottomrule
\end{tabular}\end{adjustbox}

\end{table}

Operation-only restoration has the same aggregate results as full phase restoration; candidate-only restoration fixes one new donor error. All substitutions keep the recipient's other channels and use both views as a diagnostic reference. This confirmation tests new program families with the fixed encoder, known vocabulary, supplied traversal and previously selected model family.

\begin{table}[!htbp]\centering\small
\caption{\textbf{Benefits and costs of full phase restoration.} Both views supply the fixed fitted-model intervention. Corr.\ D and New\ D count all donor errors corrected and donor-correct answers made wrong. Fixed counts corrected errors newly induced from an originally correct answer. Diff.\ F counts disagreements with the original predictions, independently of correctness.}
\label{tab:final-recovery-cost}
\begin{adjustbox}{max width=\linewidth}\begin{tabular}{lrrrrrr}\toprule
Panel & Pairs & Fixed & Corr.\ D & New\ D & Diff.\ F & BA (\%) \\\midrule
DINOv2 channel panel & 1,920 & \textbf{351/391} & 386 & \textbf{85} & \textbf{142} & 88.54 \\
DINOv1 new families & 384 & \textbf{28/29} & 28 & \textbf{3} & \textbf{2} & 98.70 \\
DINOv1 excluding siblings & 371 & 28/29 & 28 & 3 & 2 & 98.64 \\
\bottomrule\end{tabular}\end{adjustbox}\end{table}

\subsection{Sensitivity to within-family donors}
\label{sec:sibling-sensitivity}
Thirteen donor assignments in the 384-row confirmation use an opposite-endpoint sibling from the same canonical family. We retain the prespecified all-row estimate and additionally remove exactly these rows, with no re-encoding, refitting or donor reassignment. The sensitivity subset has 371 rows. Since deletion can unbalance labels, each block's BA is recomputed as the mean of its classwise recalls before averaging the eight blocks equally. Intervals use 40,000 paired whole-block percentile resamples with seed 2026111011. This post hoc check uses the same frozen outputs.
Table~\ref{tab:final-sibling} reports the recomputed classwise scores and paired contrasts.

\begin{table}[!htbp]\centering\small
\caption{\textbf{Confirmation after excluding 13 within-family donors.} All 371 retained pairs have complete recovery in both views. Intervals are 95\% paired block-bootstrap intervals.}
\label{tab:final-sibling}
\begin{adjustbox}{max width=\linewidth}\begin{tabular}{lrrr}\toprule
Computation & BA (\%) & Change from original [95\% CI], pp & Change from donor [95\% CI], pp \\\midrule
Original & 99.18 & $+0.00$ [$+0.00,+0.00$] & $+7.30$ [$+2.82,+11.81$] \\
Inject donor phase & 92.71 & $-6.47$ [$-10.39,-2.67$] & $+0.83$ [$+0.27,+1.66$] \\
Donor & 91.89 & $-7.30$ [$-11.81,-2.82$] & $+0.00$ [$+0.00,+0.00$] \\
Restore original phase & 98.64 & $-0.54$ [$-1.36,+0.00$] & $+6.76$ [$+2.70,+10.88$] \\
\bottomrule\end{tabular}\end{adjustbox}\end{table}

\FloatBarrier
\section{Operation Phase from Local Pixels}
\label{sec:supp-local}
In a two-block DINOv1 validation pilot we replace the operation-node phase projection with a small CNN. Both CNN variants use a $96\times96$ RGB window at the supplied query center, the same architecture, initialization and fixed channel normalization. The local arm sets the exterior of the central $32\times32$ tile to white. The context arm retains the surrounding pixels with white padding at image boundaries. Both CNNs supply all operation angles, including at identities. Start and candidate angles, decay, writes, queries and residuals still use the global features. The original Rotary head is retrained as a third arm. All fits use the same 16,000 training rows, common head initialization, batch sequence, 5,040 updates and task supervision. Checkpoint selection uses original-view BA on 6,000 validation rows. The balanced 480-row paired subset of that split is used for the donor diagnostic. The pilot ends at validation.

\begin{table}[!htbp]\centering\small
\caption{\textbf{Two-block operation-phase pilot on validation data.} All three arms use 5,040 updates per block. Original BA on the 6,000-row validation split selects checkpoints. Original and donor BA on its 480-row paired subset characterize the chosen fits. Lost / gained counts correct-to-wrong and wrong-to-correct predictions. The last columns report operation-phase RMS in radians and measured training plus validation and diagnostics time on cached features. All 480 pairs have exact recovery in each block.}
\label{tab:local-phase-pilot}
\setlength{\tabcolsep}{4.1pt}
\fitsuppcolumn{\begin{adjustbox}{max width=\linewidth}\begin{tabular}{llrrrrrrrr}
\toprule
Block & Phase input & Orig.\ 6,000 BA & Orig.\ 480 BA & Donor BA & Flips (\%) & Lost / gained & Phase RMS & Params & Time (s) \\
\midrule
1 & DINO features & 100.000 & 100.000 & 99.583 & 0.417 & 2 / 0 & 0.1594 & 123,090 & 274.9 \\
 & Context CNN & 99.800 & 99.792 & 50.833 & 49.375 & 236 / 1 & 0.6086 & 139,806 & 618.0 \\
 & Local CNN & 98.317 & 97.708 & 65.208 & 34.583 & 161 / 5 & 0.0000 & 139,806 & 596.7 \\
\addlinespace[2pt]
2 & DINO features & 99.950 & 100.000 & 99.583 & 0.417 & 2 / 0 & 0.1830 & 123,090 & 274.2 \\
 & Context CNN & 97.650 & 97.500 & 53.333 & 47.083 & 219 / 7 & 0.6963 & 139,806 & 614.0 \\
 & Local CNN & 99.867 & 100.000 & 98.125 & 1.875 & 9 / 0 & 0.0000 & 139,806 & 589.3 \\
\bottomrule
\end{tabular}\end{adjustbox}}
\end{table}

\begin{table}[!htbp]\centering\small
\caption{\textbf{Restoring one channel in the fitted local-phase heads.} Each block uses 480 jointly exact validation pairs. For each row between the reference rows, the specified channels use their original-view values and the others use donor values. Corrections and introduced errors are relative to donor predictions. Disagreement is measured against the original-view predictions.}
\label{tab:local-phase-remaining}
\setlength{\tabcolsep}{8pt}
\fitsuppcolumn{\begin{adjustbox}{max width=\linewidth}\begin{tabular}{lrrrrrr}
\toprule
& \multicolumn{3}{c}{Block 1} & \multicolumn{3}{c}{Block 2} \\
\cmidrule(lr){2-4}\cmidrule(lr){5-7}
Channel restored & BA (\%) & Corrected / introduced & Disagree & BA (\%) & Corrected / introduced & Disagree \\
\midrule
Donor reference & 65.208 & 0 / 0 & 166 & 98.125 & 0 / 0 & 9 \\
Operation phase & 65.208 & 0 / 0 & 166 & 98.125 & 0 / 0 & 9 \\
Candidate phase & 66.250 & 9 / 4 & 159 & 98.125 & 0 / 0 & 9 \\
All non-operation phase & 66.250 & 9 / 4 & 159 & 98.125 & 0 / 0 & 9 \\
All decay & 75.625 & 53 / 3 & 118 & 98.333 & 1 / 0 & 8 \\
All writes & 83.542 & 95 / 7 & 74 & 100.000 & 9 / 0 & 0 \\
All queries & 69.375 & 30 / 10 & 144 & 99.792 & 9 / 1 & 1 \\
All residuals & 65.417 & 6 / 5 & 163 & 98.333 & 1 / 0 & 8 \\
Original reference & 97.708 & 161 / 5 & 0 & 100.000 & 9 / 0 & 0 \\
\bottomrule
\end{tabular}\end{adjustbox}}
\end{table}

All 6,859 tested operation crops are byte-identical within pairs. The local angle differences are exactly zero in every recorded epoch in both blocks. On these queries the operation innovation becomes
\[
E_t=S_{t-1}^F(\rho_t^D-\rho_t^F)R(\theta_t)+(B_t^D-B_t^F).
\]
Equal angles remove one term in the transition difference, while decay and writes can still contribute to the state difference. To test reliance on input-varying operation angles, we replace each operation angle with the per-plane pooled circular mean of the original validation angles and apply it to both views. Original BA on the 480-row subset drops from 97.708\% to 48.125\% and from 100\% to 52.083\%. Donor BA changes from 65.208\% to 50.417\% and from 98.125\% to 51.667\%. The intervention uses no labels to compute the mean and makes no change to weights or checkpoint selection.

The local CNN retains higher donor BA than the context CNN on both blocks and lower donor BA than the original global head (Table~\ref{tab:local-phase-pilot}). All arms pass the initial 85\% learning screen. The later expansion criterion requires both CNNs to reach 99\% original validation BA in both blocks and the local deficit to stay within 0.5 pp of the global reference. The context CNN in block 2 and the local CNN in block 1 miss that criterion. The phase pilot therefore ends after two blocks.

In the selected local heads, donor-to-original operation-phase exchange changes no probabilities. Restoring all other phase angles produces the same probabilities as restoring only the candidate phase. Among single-channel restorations, replacing the composed writes gives the largest BA gain in both blocks. Restoring decay in the first block also increases BA by 10.42 pp. Table~\ref{tab:local-phase-remaining} gives all correction and introduced-error counts. These substitutions measure each channel conditional on the other donor channels within each independently fitted head.

\paragraph{All-node local visual reference.}
In a separate two-block engineering comparison after the phase pilot, the same CNN trunk projects each local crop to a 384-dimensional vector for the entire Rotary head. It uses no frozen encoder and retains the pilot's data, 5,040 updates, recurrent-head initialization and validation selection. All 151,026 parameters are in the optimizer. Both CNN and head have nonzero gradients and weight changes. With identical local inputs across the original and donor views, embeddings, phase angles and predictions agree exactly. The original validation BA is 52.53\% and 64.03\% in the two blocks.

A separate role-wise ridge decoder is fitted on the frozen CNN features with symbol labels and selected on validation. It recovers 5,948 and 5,817 of 6,000 complete validation chains. On the previously examined channel panel it recovers 238 and 234 of 240 programs, with rule BA of 99.58\% and 98.33\%. The learned head's low BA coexists with exact paired invariance and high recovery under this readout. Table~\ref{tab:all-node-engineering} also gives the four fitted heads' channel-panel results. The two blocks are drawn from the channel panel, with the exact ten-nonidentity-plus-one-identity subset reported separately. The analysis follows all model and checkpoint selection.

\begin{table}[!htbp]\centering\small
\caption{\textbf{All-node local CNN.} Top: model and separate symbol-supervised decoder measurements for the RGB-only CNN feeding all Rotary channels. The learned head uses the same 6,000-row validation panel for selection and its 480-row paired subset for the donor diagnostic. Rule BA executes the decoded program with the known grammar. Bottom: frozen heads on the two previously examined channel-panel blocks, with the exact $10+1$ composition subset shown separately. BA is class-balanced. Flips are counts, and transition counts are original-correct to donor-wrong / original-wrong to donor-correct.}
\label{tab:all-node-engineering}
\setlength{\tabcolsep}{6.1pt}
\fitsuppcolumn{\begin{adjustbox}{max width=\linewidth}\begin{tabular}{lrrrrrrr}
\toprule
\multicolumn{8}{l}{\textbf{All-node local CNN}} \\
\addlinespace[2pt]
Block & Val.\ BA & Paired F / D BA & Val.\ exact chains & Val.\ rule BA & Channel exact chains & Channel rule BA & Time (s) \\
\midrule
1 & 52.53 & 53.333 / 53.333 & 5,948/6,000 & 99.38 & 238/240 & 99.58 & 598.2 \\
2 & 64.03 & 61.875 / 61.875 & 5,817/6,000 & 98.17 & 234/240 & 98.33 & 598.2 \\
\bottomrule
\end{tabular}\end{adjustbox}}
\par\medskip
\fitsuppcolumn{\begin{adjustbox}{max width=\linewidth}\begin{tabular}{llrrrrrr}
\toprule
& & \multicolumn{3}{c}{All channel-panel pairs ($n=240$ per block)} & \multicolumn{3}{c}{$10+1$ subset ($n=80/77$)} \\
\cmidrule(lr){3-5}\cmidrule(lr){6-8}
Block & Arm & F / D BA & Flips & Transitions & F / D BA & Flips & Transitions \\
\midrule
1 & DINO features & 99.17 / 97.50 & 8 & 6 / 2 & 97.50 / 97.50 & 4 & 2 / 2 \\
 & Context phase CNN & 97.50 / 48.75 & 117 & 117 / 0 & 95.00 / 50.00 & 36 & 36 / 0 \\
 & Local phase CNN & 87.50 / 55.83 & 96 & 86 / 10 & 81.25 / 56.25 & 30 & 25 / 5 \\
 & All-node local CNN & 49.58 / 49.58 & 0 & 0 / 0 & 51.25 / 51.25 & 0 & 0 / 0 \\
\addlinespace[2pt]
2 & DINO features & 100.00 / 99.17 & 2 & 2 / 0 & 100.00 / 98.75 & 1 & 1 / 0 \\
 & Context phase CNN & 83.33 / 49.58 & 123 & 102 / 21 & 76.89 / 43.99 & 40 & 33 / 7 \\
 & Local phase CNN & 97.08 / 90.83 & 17 & 16 / 1 & 93.65 / 85.95 & 8 & 7 / 1 \\
 & All-node local CNN & 55.42 / 55.42 & 0 & 0 / 0 & 50.88 / 50.88 & 0 & 0 / 0 \\
\bottomrule
\end{tabular}\end{adjustbox}}
\end{table}

\FloatBarrier
\section{Fixed-Layout Whole-Canvas Context}
\label{sec:supp-canvas}
We render a ten-node G5c distractor in rows $[0,32)$ of the $112\times448$ canvas. The target occupies rows $[32,112)$ with its original query centers at $y=56$. Distractor draws D0 and D1 follow the same law, use different families, and fully cross distractor legality across the two views and target legality. Candidate siblings share their distractor pair. All target features in a view come from a single encoder pass. The target pixels, query coordinates and sampled patch embeddings are identical across views. The respective vertical patch footprints are $[48,64)$ and $[42,70)$ for DINOv1 and DINOv2. Sixteen duplicate inputs per block and panel agree exactly at all measured layers. All 24 block-by-panel paired RGB hashes agree across encoders. Across blocks on L7, final-layer feature RMS differences range from 0.662 to 0.714 for DINOv1 and from 0.517 to 0.545 for DINOv2.

For each encoder and head we fit eight models in the same two-row layout using 16,000 training rows, 6,000 validation rows and 5,040 updates per block. Global canonical target-family exclusions separate training, validation and test. Replacement draws preserve the archived training and validation counts by split, nonidentity length, node count and label. The frozen test manifests are released only after checkpoint selection for each encoder. Selected validation BA ranges for Rotary, 2F and GRU are 99.95--100\%, 99.98--100\% and 99.97--100\% with DINOv1 and 89.40--94.98\%, 93.98--95.63\% and 94.37--95.60\% with DINOv2. Role-specific ridge decoders use training and validation only.

\begin{table}[!htbp]\centering\small
\caption{\textbf{Paired distractors in the fixed two-row layout.} Pooled BA and disagreement rates in percent. L7, L3--7 and L10 contain 2,432, 7,680 and 3,840 pairs per encoder. Intervals cluster on blocks and families. Degenerate denotes zero disagreement in every observed-data bootstrap draw. Exact chains is shared across heads within encoder and panel. Exact flips gives the count in the jointly exact subset. The last column counts correct-to-wrong and wrong-to-correct transitions from D0 to D1.}
\label{tab:coherent-canvas-supp}
\setlength{\tabcolsep}{3.1pt}
\fitsuppcolumn{\begin{adjustbox}{max width=\linewidth}\begin{tabular}{lllrrrrrrr}
\toprule
Encoder & Panel & Head & D0 / D1 BA & Flips (\%) & 95\% interval & Both correct & Exact chains & Exact flips & Transitions \\
\midrule
DINOv1 & L7 & Rotary & 100.000 / 100.000 & 0 (0.000) & Degenerate & 2,432 & 2,432/2,432 & 0 & 0 / 0 \\
 &  & 2F & 99.959 / 100.000 & 1 (0.041) & [0.000, 0.247] & 2,431 &  & 1 & 0 / 1 \\
 &  & GRU & 99.959 / 99.959 & 0 (0.000) & Degenerate & 2,431 &  & 0 & 0 / 0 \\
\addlinespace[2pt]
 & L3--7 & Rotary & 99.987 / 99.974 & 1 (0.013) & [0.000, 0.077] & 7,678 & 7,679/7,680 & 1 & 1 / 0 \\
 &  & 2F & 100.000 / 99.974 & 2 (0.026) & [0.000, 0.117] & 7,678 &  & 2 & 2 / 0 \\
 &  & GRU & 100.000 / 99.987 & 1 (0.013) & [0.000, 0.078] & 7,679 &  & 1 & 1 / 0 \\
\addlinespace[2pt]
 & L10 & Rotary & 98.047 / 97.995 & 24 (0.625) & [0.208, 1.224] & 3,752 & 3,840/3,840 & 24 & 13 / 11 \\
 &  & 2F & 99.948 / 99.896 & 4 (0.104) & [0.000, 0.260] & 3,835 &  & 4 & 3 / 1 \\
 &  & GRU & 99.974 / 99.922 & 4 (0.104) & [0.000, 0.286] & 3,836 &  & 4 & 3 / 1 \\
\addlinespace[2pt]
DINOv2 & L7 & Rotary & 91.036 / 91.365 & 214 (8.799) & [6.949, 10.691] & 2,111 & 2,053/2,432 & 153 & 103 / 111 \\
 &  & 2F & 94.490 / 94.778 & 131 (5.387) & [4.030, 6.867] & 2,236 &  & 96 & 62 / 69 \\
 &  & GRU & 94.243 / 94.408 & 156 (6.414) & [4.770, 8.306] & 2,216 &  & 105 & 76 / 80 \\
\addlinespace[2pt]
 & L3--7 & Rotary & 92.760 / 93.047 & 612 (7.969) & [6.763, 9.301] & 6,829 & 6,457/7,680 & 435 & 295 / 317 \\
 &  & 2F & 94.935 / 94.909 & 464 (6.042) & [5.221, 6.937] & 7,058 &  & 315 & 233 / 231 \\
 &  & GRU & 94.453 / 94.154 & 471 (6.133) & [5.130, 7.266] & 7,007 &  & 313 & 247 / 224 \\
\addlinespace[2pt]
 & L10 & Rotary & 71.328 / 70.781 & 717 (18.672) & [15.990, 21.406] & 2,370 & 3,313/3,840 & 610 & 369 / 348 \\
 &  & 2F & 93.516 / 93.958 & 309 (8.047) & [6.979, 9.167] & 3,445 &  & 230 & 146 / 163 \\
 &  & GRU & 93.620 / 94.323 & 293 (7.630) & [6.432, 8.906] & 3,462 &  & 200 & 133 / 160 \\
\addlinespace[2pt]
\bottomrule
\end{tabular}\end{adjustbox}}
\end{table}

Tables~\ref{tab:coherent-canvas-supp} and \ref{tab:coherent-phase-supp} report paired capability, disagreement and phase exchanges. The fixed L7 panel crosses 76 canonical families with eight blocks, two instances per family and both candidate labels. The IID-length L3--7 panel uses 106 reserved short families and approximates the earlier joint length and node-count distribution. The 3-operation row represents one family. The L10 panel has 1,920 disjoint canonical families, two candidate labels per family and no identity operations. The L7 and L10 panels therefore differ in program composition as well as length. In the short panels, 10,000 bootstrap draws resample training blocks and canonical families in crossed fashion and hold the number of families within each length fixed. L10 resamples families nested within blocks. Zero-event resamples describe the observed panel. We report prediction disagreement because both views follow the same law, and the expected directional BA difference is zero. On L7, DINOv2 Rotary's jointly exact subset has 7.45\% disagreement ($[5.65,9.38]$). The all-pair D1-minus-D0 BA difference is $+0.33$ pp ($[-1.60,+2.26]$). On the 153 jointly exact L7 DINOv2 disagreements, exchanging all phase reproduces the source decision on 60 and 70 pairs in the two directions. Exchanging operation phase alone gives 57 and 50, and exchanging candidate phase alone gives 26 and 15. A semantic control edits only the target candidate, reverses the label in both backgrounds, and keeps the other target pixels fixed. The archive reports capability and correction counts under this edit.

\begin{table}[!htbp]\centering\small
\caption{\textbf{Rotary phase exchange on whole-canvas pairs.} Each replay uses the other view's phase at the indicated nodes and the base view's remaining channels. Agreement to source is conditional on the original pair disagreeing. Corrected and introduced count true-label error transitions over all pairs in the corresponding panel.}
\label{tab:coherent-phase-supp}
\setlength{\tabcolsep}{6pt}
\fitsuppcolumn{\begin{adjustbox}{max width=\linewidth}\begin{tabular}{lllrrrr}
\toprule
& & & \multicolumn{2}{c}{Agree with source on disagreements} & \multicolumn{2}{c}{Corrected / introduced over all pairs} \\
\cmidrule(lr){4-5}\cmidrule(lr){6-7}
Encoder & Panel & Phase exchanged & D0 from D1 & D1 from D0 & D0 from D1 & D1 from D0 \\
\midrule
DINOv1 & L7 & All valid nodes & --- & --- & 0 / 0 & 0 / 0 \\
 &  & Operations & --- & --- & 0 / 0 & 0 / 0 \\
 &  & Candidate & --- & --- & 0 / 0 & 0 / 0 \\
 &  & Start & --- & --- & 0 / 0 & 0 / 0 \\
\addlinespace[2pt]
 & L3--7 & All valid nodes & 0/1 & 0/1 & 0 / 0 & 0 / 0 \\
 &  & Operations & 0/1 & 0/1 & 0 / 0 & 0 / 0 \\
 &  & Candidate & 0/1 & 0/1 & 0 / 0 & 0 / 0 \\
 &  & Start & 0/1 & 0/1 & 0 / 0 & 0 / 0 \\
\addlinespace[2pt]
 & L10 & All valid nodes & 23/24 & 20/24 & 13 / 19 & 12 / 12 \\
 &  & Operations & 21/24 & 11/24 & 12 / 17 & 6 / 8 \\
 &  & Candidate & 9/24 & 3/24 & 4 / 6 & 4 / 4 \\
 &  & Start & 0/24 & 0/24 & 0 / 0 & 0 / 0 \\
\addlinespace[2pt]
DINOv2 & L7 & All valid nodes & 84/214 & 85/214 & 61 / 60 & 62 / 80 \\
 &  & Operations & 80/214 & 60/214 & 59 / 57 & 48 / 58 \\
 &  & Candidate & 41/214 & 21/214 & 39 / 25 & 23 / 19 \\
 &  & Start & 0/214 & 0/214 & 0 / 0 & 0 / 0 \\
\addlinespace[2pt]
 & L3--7 & All valid nodes & 228/612 & 229/612 & 175 / 167 & 169 / 170 \\
 &  & Operations & 173/612 & 184/612 & 148 / 133 & 126 / 154 \\
 &  & Candidate & 89/612 & 89/612 & 66 / 84 & 80 / 71 \\
 &  & Start & 0/612 & 0/612 & 0 / 0 & 0 / 0 \\
\addlinespace[2pt]
 & L10 & All valid nodes & 420/717 & 434/717 & 278 / 289 & 290 / 299 \\
 &  & Operations & 394/717 & 383/717 & 279 / 261 & 263 / 265 \\
 &  & Candidate & 121/717 & 128/717 & 94 / 101 & 108 / 103 \\
 &  & Start & 0/717 & 0/717 & 0 / 0 & 0 / 0 \\
\addlinespace[2pt]
\bottomrule
\end{tabular}\end{adjustbox}}
\end{table}

\paragraph{Complementary feature recombination.}
A further replay reuses the same two feature sequences and the same fitted heads. We draw eight operation masks per target and mix features from the two scenes with each mask and its complement. The two mixed sequences together use exactly the same feature multiset at each node as the original pair. The start and candidate features remain together in their original view. Masks and their complements contribute to the within-target average before resampling. For DINOv1 Rotary on L10, mixed minus coherent mean BA is $-0.0081$ pp ($[-0.1367,+0.1335]$). With DINOv2 Rotary on L7, the change is $+0.2596$ pp ($[-0.3316,+0.9020]$). DINOv2 2F shows small average declines on the IID-length and L10 panels. The replay tests feature mixing over these two backgrounds and shows little BA change for Rotary. Its mixed sequences combine features from the two RGB images.

\FloatBarrier
\raggedbottom
\section{Reproducibility Package}
\label{sec:supp-repro}
The accompanying \texttt{REPRODUCIBILITY.md} maps every display to its inputs and commands for cached replay, encoding, refits and builds. Manifests retain programs, labels, render seeds and donors. Cached replay needs no download; re-encoding needs the documented encoder weights, and earlier VLM runs need their external weights.

RGB encoding uses float32, channel mean $(.485,.456,.406)$ and standard deviation $(.229,.224,.225)$; ordered 384-dimensional query features are stored in float16. Training standardization, threshold 0.5, classwise BA and complete-chain recovery are defined above. Eight-block intervals weight blocks equally; two-block glyph panels report individual observations. The package records software versions, checkpoints, splits and encoder portability checks.

\end{document}